\pdfoutput=1
\documentclass{article}

\PassOptionsToPackage{numbers}{natbib}

\usepackage[eandd, final]{neurips_2026}

\usepackage[utf8]{inputenc}
\usepackage[T1]{fontenc}
\usepackage{hyperref}
\usepackage{orcidlink}
\usepackage{url}
\usepackage{booktabs}
\usepackage{array}
\usepackage{multirow}
\usepackage{amsfonts}
\usepackage{nicefrac}
\usepackage{microtype}
\usepackage{xcolor}

\usepackage{amsmath,amssymb}
\usepackage{graphicx}

\usepackage{algorithm}
\usepackage{algorithmic}
\usepackage{makecell}
\usepackage{tabularx}
\usepackage{subcaption}
\usepackage{wrapfig}

\usepackage{tikz}
\usetikzlibrary{positioning, arrows.meta, calc, fit, backgrounds, shapes.geometric}

\title{AbsoluteDegradation: A Physics-Inspired Synthetic Film-Degradation Pipeline and Archival Film Restoration Benchmark}

\author{%
  \textbf{Mikołaj Jastrzębski}\,\orcidlink{0009-0003-3526-935X} \quad
  \textbf{Dawid Glinkowski}\,\orcidlink{0009-0008-4342-9663} \quad
  \textbf{Dawid Zieliński}\,\orcidlink{0009-0009-6181-7114} \\
  \textbf{Daniel Borkowski}\,\orcidlink{0009-0003-0757-4103} \quad
  \textbf{Wojciech Kozłowski}\,\orcidlink{0009-0009-1532-4415} \quad
  \textbf{Kamil Adamczewski}\,\orcidlink{0000-0002-2917-4392} \\[6pt]
  \mdseries
  Wrocław University of Science and Technology \\[4pt]
  \texttt{contact@mikjas.com}, \texttt{dawid.g.dg@gmail.com}, \\
  \texttt{\{dawid.zielinski.ml,d.a.borkowski\}@outlook.com}\\
  \texttt{\{wojciech.kozlowski,kamil.adamczewski\}@pwr.edu.pl}
}

\begin{document}

\maketitle

\begin{abstract}
Restoring archival film remains a fundamentally challenging problem due to the absence of paired training data and the lack of standardized evaluation benchmarks. Pristine versions of deteriorated footage are physically unrecoverable, requiring supervised methods to rely on synthetic data that often fail to capture the complex, temporally coherent nature of real film degradation. At the same time, existing real-world datasets are limited in scale, quality, and accessibility, hindering reliable evaluation and fair comparison across methods. We address both limitations with \textbf{AbsoluteDegradation}, a physics-inspired, modular pipeline for synthesizing realistic film degradations, and a new large-scale archival benchmark. The proposed pipeline models the analog-to-digital process as a structured composition of artifact families, incorporating signal-dependent grain, parametric scratches, and temporally coherent camera motion, enabling controlled generation of diverse degradation regimes. In parallel, we introduce a curated dataset of 81,576 high-resolution frames sourced from real archival footage, designed for consistent evaluation under real-world conditions. Together, these contributions provide a unified framework for training and benchmarking restoration models. Experiments across six restoration architectures, evaluated with no-reference image quality metrics, clip-level temporal quality assessment, and a blinded human preference study (93 raters, 1,395 trials), show that models trained with AbsoluteDegradation generalize better to real-world footage: human raters preferred them in 64.7\% of comparisons and for every evaluated architecture. The proposed benchmark also reveals systematic failure modes of current methods and of no-reference metrics. We hope this work establishes a foundation for reproducible and domain-authentic evaluation in archival film restoration.
\end{abstract}

\section{Introduction}
\label{sec:intro}

Millions of hours of culturally significant footage exist exclusively on deteriorating analog film. This medium suffers from a range of artifacts, including chemical decay and mechanical scratches. As manual frame-by-frame restoration is prohibitively labor-intensive at this scale, deep learning has emerged as a necessity. Using high-capacity neural architectures, it is now possible to automate the recovering underlying visual content from physical decay, offering a scalable path towards preserving audiovisual heritage in the digital age. However, restoring archival film remains fundamentally challenging for two key reasons.

First, the effectiveness of modern neural restoration models is inherently limited by the absence of paired training data: pristine, ground-truth versions of deteriorated archival footage are impossible to recover. As a result, supervised learning must rely on synthetic approximations, which often fail to faithfully capture the complex, physically grounded nature of real film degradation. Existing degradation models, whether high-order pipelines~\citep{wang2021realesrgan} or film-specific~\citep{wan2022oldfilm}, largely treat artifacts as independent image-domain perturbations. Furthermore, they neglect analog capture physics, such as signal-dependent grain or exposure fluctuations. They also fail to preserve temporal coherence across frames. Consequently, a significant domain gap persists, limiting effective training.

Second, there is a lack of high-quality standardized evaluation datasets for this domain. Copyright restrictions often block the release of authentic archival footage. Although recent public benchmarks such as the Synthetic/Real-world Old Video (SRWOV) dataset~\citep{mambaofr} exist, it remains limited in scale and mixes the domain of animation with archival film. This heterogeneity introduces artifacts unrelated to analog decay. Ultimately, the scarcity of comprehensive and domain-specific datasets makes it difficult to reliably assess restoration quality and compare methods in a consistent manner.

To bridge this gap, we present \textbf{AbsoluteDegradation}, a modular, physics-inspired synthesis pipeline, and a corresponding real-world benchmark. Our contributions are: \textbf{(i) AbsoluteDegradation Synthesis Engine:} A modular framework that models the analog-to-digital chain. By utilizing signal-dependent blue-noise grain and mean-reverting Ornstein--Uhlenbeck processes for gate-weave, our pipeline captures the complex temporal and spatial dependencies of authentic film decay; and  \textbf{(ii) Large-Scale Real-World Archival Benchmark:} We introduce a curated dataset of 81,576 frames sourced directly from the real-world archival domain. To our knowledge, this represents the largest high-resolution benchmark of \textit{in-the-wild} early analog film degradations, providing a rigorous domain-focused benchmark for zero-shot evaluation and cross-domain validation.

Together, these contributions form a unified framework for training and evaluation of archival film restoration. Experiments on six architectures demonstrate that models trained with AbsoluteDegradation generalize better to real archival footage: human raters prefer them for every evaluated architecture, with the size of the benefit varying across models. Critically, our benchmark exposes a systematic flaw in the current evaluation paradigm: standard no-reference quality metrics actively reward historically inauthentic hallucinations and over-smoothing. These findings suggest that misaligned evaluation proxies have partially obscured real progress in this domain. This reinforces the need for our physics-inspired data synthesis and domain-representative benchmark.

\section{Related Work}
\label{sec:related}

\paragraph{Synthetic Degradation Pipelines.}
Supervised restoration requires paired data, since real
sources almost never preserve originals, Real-ESRGAN~\citep{wang2021realesrgan}
and BSRGAN~\citep{zhang2021bsrgan} established the high-order degradation
composition paradigm, randomising sequences of blur, downsampling, noise,
and JPEG compression to approximate real-world image corruption.
These pipelines are effective for general photographic degradation but contain
no film-specific artifacts: they model no grain heteroscedasticity, no scratch
geometry or temporal drift, no gate-weave jitter, and no texture blending.

\citet{ivanova2023simulating} introduced physically motivated per-artifact models for high-resolution photographic scans. They showed that degradation fidelity, not just coverage, determines downstream restoration quality. However, their simulator is restricted to still images and does not address the temporal coherence required for video restoration.

Previous works \citep{wan2022oldfilm,iizuka2019deepremaster} explored the synthesis of video specific degradation using curated scanned film textures composed onto clean footage.
Their approach produces realistic artifacts, but it has several important limitations that we address.
(i) Temporal dynamics are simplified to a single texture with moving lines. The method does not use a parametric model for scratch creation, lifetime, polarity, or independent motion paths.
(ii) Noise changes randomly between signal-dependent heteroscedastic noise and additive Gaussian noise within the same clip. This does not reflect the stable grain pattern typically found in a single film strip. The method also uses Gaussian statistics instead of the blue noise structure of silver halide film grain.
(iii) The approach does not model gate weave or camera motion jitter.
(iv) Capture stage degradations, such as optical blur and grain added on the negative, and digitization stage degradations, such as JPEG compression, downsampling, and photometric shifts, are applied together in a single pass. This ignores the layered processing structure that appears in real analog to digital film pipelines. AbsoluteDegradation addresses all four limitations through six ordered stages that introduce the missing artifact families, a permuted core spanning  $7!{=}5{,}040$ orderings (vs.\ $4!{=}24$ in~\citep{wan2022oldfilm}),  and a three-tier severity curriculum.

\paragraph{Data Resources for Film Restoration}

Evaluation and supervised training of video restoration models rely on paired datasets
synthesized from modern and advanced cameras.
REDS~\citep{nah2019reds} is the standard benchmark for video super-resolution
and deblurring, providing clean 720p clips alongside synthetically degraded
counterparts at controlled severity levels.
Such resources have driven significant progress, but they represent natural
scene degradation with known, simple statistics; none contains the compound
artifacts characteristic of analog film deterioration.

\begin{wraptable}{r}{0.52\textwidth}
\vspace{-10pt}

\centering
\scriptsize

\begin{minipage}{0.52\textwidth}

\caption{Comparison of synthetic degradation pipelines for image and video
restoration. \textbf{TC}~=~temporal coherence across frames;
\textbf{FA}~=~film-specific artifacts (textures, gate-weave, flicker);
\textbf{AFC}~=~artifact-family coverage count.
Symbols: $\times$~=~absent, $\sim$~=~partial, $\checkmark$~=~present.}

\label{tab:pipeline-comparison}

\centering

\begin{tabular}{@{}lcccc@{}}
\toprule
\textbf{Pipeline} & \textbf{Domain} & \textbf{TC} & \textbf{FA} & \textbf{AFC} \\
\midrule
Real-ESRGAN~\citep{wang2021realesrgan}
& Image & $\times$ & $\times$ & 3 \\

BSRGAN~\citep{zhang2021bsrgan}
& Image & $\times$ & $\times$ & 3 \\

Bringing Old Films~\citep{wan2022oldfilm}
& Video & $\sim$ & $\sim$ & 4 \\

Ivanova et al.~\citep{ivanova2023simulating}
& Image & $\times$ & $\sim$ & 2 \\

\textbf{AbsoluteDegradation (ours)}
& Video & $\checkmark$ & $\checkmark$ & \textbf{7} \\
\bottomrule
\end{tabular}

\vspace{1.5mm}

\parbox{0.98\linewidth}{\scriptsize\emph{Taxonomy for AFC}:
\emph{(1)} optical blur,
\emph{(2)} film grain,
\emph{(3)} parametric scratches,
\emph{(4)} dust and texture overlays,
\emph{(5)} negative-film distortions,
\emph{(6)} camera gate-weave,
\emph{(7)} compression and photometric shifts.}

\end{minipage}

\vspace{-10pt}
\end{wraptable}

MambaOFR~\citep{mambaofr} introduced the first old-film benchmark (SRWOV),
but its low-resolution frames, prevalence of animated content
(over 26\%), and scarcity of authentic film degradations limit its
domain representativeness. We address these shortcomings with the benchmark introduced in Section~\ref{sec:old_video_dataset}.

Table~\ref{tab:pipeline-comparison} positions \textbf{AbsoluteDegradation}
against prior pipelines across the most crucial attributes for archival film restoration.
No existing pipeline simultaneously covers all seven analog artifact
families, maintains temporal coherence, and provides a configurable severity curriculum.
\textbf{AbsoluteDegradation} is the first synthesis pipeline to jointly satisfy all three properties.

The literature overview of works on old image and video restoration is postponed to Appendix~\ref{sec:appendix-related-works}.

\section{AbsoluteDegradation Pipeline}
\label{sec:method}

The degradation of archival film is the result of a complex interplay of
physical, chemical, and mechanical processes that act at different stages of
the analog-to-digital pipeline. Rather than modeling degradation as a single
composite corruption, we treat it as a structured process composed of multiple
artifact families, each governed by distinct physical mechanisms and temporal
dynamics.

To this end, we introduce \textbf{AbsoluteDegradation}, a modular pipeline that
transforms clean video clips into realistically degraded counterparts through a
sequence of parameterized operators. The pipeline is illustrated in Figure \ref{fig:pipeline} and is designed around three key
principles:

\textbf{(i) Physics-inspired, interpretable modeling.} Each degradation family is modeled using empirically chosen
parameters with direct physical interpretation, such as exposure-dependent
grain, diffraction-based scratch profiles, and mean-reverting camera jitter.
This ensures that the generated artifacts reflect real-world film behavior
rather than purely heuristic perturbations. Each stage is derived from a physical
process of capture, storage, or scanning, as discussed with institutions
conserving such footage.
\textbf{(ii) Temporal coherence.} Degradations are applied at the clip level,
with parameters sampled once per sequence, and evolved through structured
stochastic processes where appropriate. This enforces consistency across frames
and avoids unrealistic frame-to-frame variation.
\textbf{(iii) Compositional diversity.} The degradation process is constructed
as a stochastic composition of operators, enabling a large and diverse space of
artifact combinations.

Formally,
given a clean video clip $\mathbf{x} = (\mathbf{x}_1, \dots, \mathbf{x}_T)$, the pipeline
generates a degraded counterpart $\tilde{\mathbf{x}}$ by applying a sequence of
stochastic transformations conditioned on a sampled severity level $d \in \{\textsc{light}, \textsc{medium}, \textsc{heavy}\}$. The
resulting process defines a distribution over degraded video data,
which can be used both for training restoration models and for systematic
evaluation under varying degradation conditions.

In the following sections, we describe the overall pipeline structure and detail
the individual degradation stages displayed in Figure \ref{fig:distortion_operator_atlas}. We also present the resulting dataset of
synthetically degraded videos paired with their corresponding clean ground-truth
counterparts.

\subsection{Degradation Pipeline Stages}

\providecommand{\fighint}[1]{{\tiny\itshape\textcolor{black!60}{#1}}}

\begin{figure}[t]
    \centering
    \definecolor{stCropBg}{HTML}{F5E8D0}\definecolor{stCropDk}{HTML}{B28936}
    \definecolor{stArtBg}{HTML}{F9D5C8}\definecolor{stArtDk}{HTML}{B35B41}
    \definecolor{stNegBg}{HTML}{D7E5EE}\definecolor{stNegDk}{HTML}{4E7EA4}
    \definecolor{stCoreBg}{HTML}{FCE4A3}\definecolor{stCoreDk}{HTML}{A87C1D}
    \definecolor{stFinBg}{HTML}{E2D3E7}\definecolor{stFinDk}{HTML}{7D5A94}
    \definecolor{stScrBg}{HTML}{D9E4C8}\definecolor{stScrDk}{HTML}{6E8C48}
    \definecolor{stIOBg}{HTML}{EEE9DE}
    \definecolor{stBanBg}{HTML}{FBEAE6}\definecolor{stBanDk}{HTML}{8A382A}
    \begin{tikzpicture}[
        >={Stealth[length=1.8mm, width=1.5mm]},
        font=\scriptsize,
        stage/.style={
            draw, rounded corners=2pt, line width=0.5pt,
            minimum height=17mm, minimum width=31mm,
            text width=28mm, align=center, inner sep=1.1mm,
        },
        io/.style   ={stage, fill=stIOBg,   draw=black!45},
        sCrop/.style={stage, fill=stCropBg, draw=stCropDk},
        sArt/.style ={stage, fill=stArtBg,  draw=stArtDk},
        sNeg/.style ={stage, fill=stNegBg,  draw=stNegDk},
        sCore/.style={stage, fill=stCoreBg, draw=stCoreDk},
        sFin/.style ={stage, fill=stFinBg,  draw=stFinDk},
        sScr/.style ={stage, fill=stScrBg,  draw=stScrDk},
        arr/.style  ={->, line width=0.75pt, draw=black!70},
        gtarr/.style={->, line width=0.55pt, dashed, draw=black!55},
        banner/.style={
            rectangle, rounded corners=2pt, fill=stBanBg, draw=stBanDk,
            line width=0.4pt, inner sep=1.4mm, align=center,
        },
    ]

    \node[banner, text width=0.913\linewidth] (banner) {%
        \textbf{\textcolor{stBanDk}{Per-clip parameter lock.}}
        Degradation degree $d\in\{\textsc{l},\textsc{m},\textsc{h}\}$
        , all degradation parameters
        and the random-order permutation $\pi(\cdot)$ are drawn
        \emph{once per clip} and held fixed across all $T$ frames.
    };

    \node[io, below=2.5mm of banner.south west, anchor=north west]
        (input) {%
            \textbf{INPUT}\\[0.4mm]
            Clean clip\\[-0.2mm]$x_{1:T}$%
        };
    \node[sCrop, right=2mm of input] (crop) {%
        {\bfseries\textcolor{stCropDk}{\footnotesize STAGE 1}}\\[-0.3mm]
        \textbf{Frame Crop}\\[0.3mm]
        {\scriptsize window of size $H\!\times\!W$ with motion padding }%
    };
    \node[sArt, right=2mm of crop] (artif) {%
        {\bfseries\textcolor{stArtDk}{\footnotesize STAGE 2}}\\[-0.3mm]
        \textbf{Dust \&\ Texture Overlay}\\[0.3mm]
        {\scriptsize $1\!-\!3$ scanned emulsion textures per frame}%
    };
    \node[sNeg, right=2mm of artif] (neg) {%
        {\bfseries\textcolor{stNegDk}{\footnotesize STAGE 3}}\\[-0.3mm]
        \textbf{Negative-Film Emulation}\\[0.3mm]
        {\scriptsize invert $\to\{$blur, grain, down., $\gamma\}_{\pi}$ $\to$ invert}%
    };

    \draw[arr] (input) -- (crop);
    \draw[arr] (crop)  -- (artif);
    \draw[arr] (artif) -- (neg);

    \node[sCore, below=10mm of input.south, anchor=north] (core) {%
        {\bfseries\textcolor{stCoreDk}{\footnotesize STAGE 4}}\\[-0.3mm]
        \textbf{Stochastic Core $\pi(\cdot)$}\\[0.3mm]
        {\scriptsize random-order per clip:\\
        blur $\cdot$ grain $\cdot$ JPEG $\cdot$ down. $\cdot$ $\gamma$ $\cdot$ $x$/$y$-motion}%
    };
    \node[sFin, right=2mm of core] (fin) {%
        {\bfseries\textcolor{stFinDk}{\footnotesize STAGE 5}}\\[-0.3mm]
        \textbf{Finalize}\\[0.3mm]
        {\scriptsize size match $\to$ color jitter $\to$ 8-bit quantize}%
    };
    \node[sScr, right=2mm of fin] (scr) {%
        {\bfseries\textcolor{stScrDk}{\footnotesize STAGE 6}}\\[-0.3mm]
        \textbf{Vertical Scratches}\\[0.3mm]
        {\scriptsize clip-level overlay: Brownian drift, flicker, multi-frame lifespan}%
    };
    \node[io, right=2mm of scr] (out) {%
        \textbf{OUTPUT}\\[0.4mm]
        $(\,\tilde{x}_{1:T}\,,\; x_{1:T}\,)$\\[0.3mm]
        \fighint{degraded / ground truth pair}%
    };

    \draw[arr] (core) -- (fin);
    \draw[arr] (fin)  -- (scr);
    \draw[arr] (scr)  -- (out);

    \coordinate (cnRight) at ($(neg.east)   + (4mm, 0)$);
    \coordinate (cnLeft)  at ($(core.west)  + (-4mm, 0)$);
    \coordinate (cnMidY)  at ($(neg.south)!0.5!(core.north)$);
    \draw[arr, rounded corners=3pt]
        (neg.east) -- (cnRight)
                   -- (cnRight |- cnMidY)
                   -- (cnLeft  |- cnMidY)
                   -- (cnLeft)
                   -- (core.west);

    \coordinate (gtBotY) at ($(core.south) + (0,-5mm)$);
    \coordinate (gtLx)   at ($(input.west) + (-8mm, 0)$);
    \coordinate (gtRx)   at ($(out.east)   + (8mm, 0)$);
    \draw[gtarr, rounded corners=3pt]
        (input.west) -- (gtLx)
                     -- (gtLx |- gtBotY)
                     -- (gtRx |- gtBotY)
                     -- (gtRx)
                     -- (out.east);
    \node[fill=white, draw=black!55, rounded corners=1.5pt, line width=0.4pt,
          inner sep=0.8mm, align=center]
        at ($(gtLx |- gtBotY)!0.5!(gtRx |- gtBotY)$)
        {\scriptsize\textbf{GT sharpening:} unsharp mask (weight $0.5$, radius $51$, threshold $10$)
         \fighint{\,applied to ground-truth $x_{1:T}$}};
    \end{tikzpicture}
    \caption{Pipeline for dataset creation.
    Each clean clip $x_{1:T}$ traverses six stages to yield a degraded counterpart $\tilde{x}_{1:T}$; the ground-truth clip is cropped identically and sharpened with the unsharp mask of~\citep{wang2021realesrgan,wan2022oldfilm}.
    }
    \label{fig:pipeline}
\end{figure}
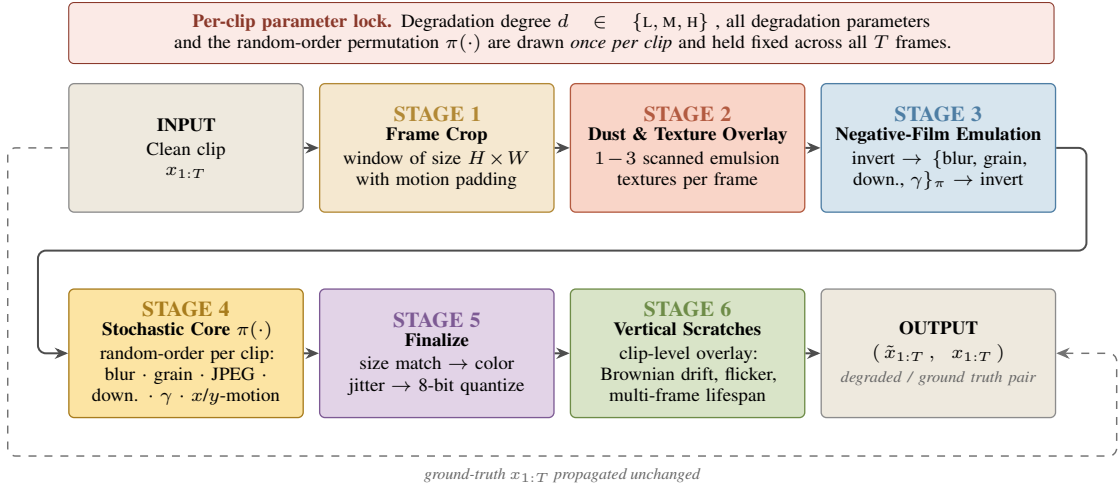

\paragraph{Stage 1: Frame Crop with Motion Padding.}
\label{sec:method:crop}
Each input frame is first cropped to a slightly larger working region that
includes additional margins around the original image. These margins act as a
buffer for subsequent geometric transformations introduced later in the pipeline.
The crop location is sampled once per clip and kept fixed across all frames,
ensuring that the degraded sequence remains spatially aligned with the
ground-truth video.
After all geometric perturbations are applied, the padded region is cropped
back to the target resolution.

\paragraph{Stage 2: Dust and Texture Overlays.}
\label{sec:method:dust}

Stage~2 simulates dust, emulsion staining, and projection marks accumulated over
decades of storage and exhibition.
Texture assets are sourced from the scanned emulsion library introduced in Bringing Old Films Back to Life~\citep{wan2022oldfilm} and converted to transparency-keyed
RGBA PNGs via adaptive Gaussian thresholding and morphological erosion.
Unlike \citep{wan2022oldfilm}, which models temporal coherence with a single moving line element, we introduce a multi slot persistence model. Independent overlay slots are assigned random lifetimes, allowing the same contaminant to remain over several consecutive frames. This encourages restoration networks to use temporal information instead of processing each frame independently.
Clips are converted to grayscale before entering the pipeline and temporarily expanded to three channels only to blend the RGBA textures.

\paragraph{Stage 3: Negative-Film Emulation.}
\label{sec:method:negative}

Analog film recording is a two-pass chemical process: capture-stage corruptions
occur on the raw negative, not the positive print.
With some probability
Stage~3 models this by applying a mild,
independently permuted sub-pipeline $\mathcal{F}_\pi$ in the inverted domain:
$
  \tilde{\mathbf{x}}_t = \operatorname{clip} \bigl(1 - \mathcal{F}_\pi(1 - \mathbf{x}_t),\;0,\,1\bigr).
  \label{eq:negative}
$
$\mathcal{F}_\pi$ comprises Gaussian blur, film grain, downsampling, and gamma
correction at reduced amplitude.
This two-pass inversion structure is inspired by the high-order degradation
composition of Real-ESRGAN~\citep{wang2021realesrgan}, but unlike its generic use
in the image domain, here it is physically grounded in the chemical reality.

\paragraph{Stage 4: Permuted Degradation Core.}
\label{sec:method:core}

Seven operators: blur ($B$), grain ($G$), JPEG ($J$), resolution change ($D$),
gamma ($\Gamma$), and horizontal/vertical gate-weave $(M_x, M_y)$ are composed
in a uniformly random order $\pi\!\sim\!\mathcal{U}(S_7)$, drawn once per clip.
With $|S_7|{=}5{,}040$ orderings, the pipeline significantly expands the space of degradation combinations~\citep{zhang2021bsrgan,wang2021realesrgan}.
The ordering carries physical meaning: $\Gamma$ before $G$ injects grain into an
exposure-corrected signal (positive-print chemistry); $G$ before $\Gamma$ places
grain in the raw-capture domain (enlarger printing from negative).
The permutation spans both regimes without separate code paths.

\subparagraph{Signal-Dependent Film Grain.}
\label{sec:method:core:grain}

Existing film restoration pipelines such as~\citep{wan2022oldfilm} model grain as white Gaussian or speckle noise ($\tilde{\mathbf{x}}_t = \mathbf{x}_t + \sigma \mathbf{x}_t \epsilon$). This is a simplification: prior analyses show that film grain has structured, signal-dependent spectral behavior and substantial high-frequency content rather than spatially independent noise~\citep{stephenson2007,oh2009}. Motivated by this, we use a Signal-Dependent Noise (SDN) formulation with a blue-noise prior~\citep{ulichney1988} as a practical high-frequency basis:
\begin{equation}
    \tilde{\mathbf{x}}_t = \mathbf{x}_t + k \cdot \mathbf{x}_t^{\gamma} \odot \mathbf{u} + \mathbf{w},
    \label{eq:sdn}
\end{equation}
Crucially, we map these variables directly to real-world attributes. The parameter $k$ controls ISO-dependent grain intensity, $\gamma$ models signal-dependent amplification, and white noise $\mathbf{w} \sim \mathcal{N}(\mathbf{0}, \sigma_w \mathbb{I})$ simulates scanner-characteristic sensor noise. The blue noise $\mathbf{u}$ (drawn from a $2048{\times}2048{\times}5$ basis) is further modulated by physical grain size. By sampling $k$, $\gamma$, grain size, and $\sigma_w$ across continuous ranges, we systematically simulate a diverse array of real film stocks and scanning equipment.
\paragraph{Camera Gate Weave. }
\label{sec:method:core:gateweave}

Gate weave is a film artifact that appears as small horizontal and vertical shifts between consecutive frames. It is caused by mechanical inaccuracies in the film transport system during scanning or projection.
The motion is smooth and gradually returns toward a central position instead of drifting without limit or changing abruptly between frames. Because of this behavior, we model the gate weave using the Ornstein-Uhlenbeck (OU) process, which naturally captures temporally correlated motion with mean reversion.
\textit{Horizontal gate weave} is modeled using the first-order OU process discretized using the Euler-Maruyama method
\begin{equation}
    X_{t+1} = X_t - \theta^{(x)} X_t + \sqrt{2\theta^{(x)}\sigma^{(x)}}\,\epsilon_t^{(x)},
    \quad \epsilon_t^{(x)} \sim\mathcal{N}(0,1)
    \label{eq:ou_x}
\end{equation}
where $\theta^{(x)}$ denotes the rate of change and $\sigma^{(x)}$ defines the variance of the stationary distribution. The generated shift $X_t$ is clipped between the padding boundary $(-p_x,p_x)$.
To better model the \textit{vertical gate weave}, which can become stronger or weaker over time, we use a second order OU process in which the parameter $\theta^{(y)}$ follows its own OU process.
\begin{align}
    \theta_{t+1}^{(y)} &= \theta_t^{(y)} + \kappa(\theta_{\mathrm{avg}}-\theta_t^{(y)})
                    + \sqrt{2 s \kappa}\,\epsilon_t^{(\theta)}, \quad \epsilon_t^{(\theta)} \sim \mathcal{N}(0, 1)
    \label{eq:ou_theta} \\
    Y_{t+1} &= Y_t - \theta_t^{(y)} Y_t + \sqrt{2\theta_t^{(y)}\sigma^{(y)}}\,\epsilon_t^{(y)},
    \qquad\;\; \epsilon_t^{(y)} \sim \mathcal{N}(0, 1)
    \label{eq:ou_y}
\end{align}
with parameters of the first process $\kappa$, $\theta_{\mathrm{avg}}$, $s$ that represent, respectively, the rate of change, mean and variance of the stationary distribution of $\theta^{(y)}$; $\theta_t^{(y)}$ is clipped to $[0.25\,\theta_{\mathrm{avg}}, 2.5\,\theta_{\mathrm{avg}}]$.
This auxiliary OU process governs the time-varying mean-reversion strength of the vertical displacement process $Y_t$, while $\sigma^{(y)}$ denotes its temperature, i.e., the variance of its stationary distribution. Similarly to the horizontal shift, the values are clipped between $(-p_y, p_y)$. All parameters were set empirically and are listed in Table~\ref{tab:severity}.
Both displacements are applied as integer crop offsets to the padding from Stage~1, producing camera jitter; both axes are active in every clip.

\subparagraph{Digitization Operators. }
\label{sec:method:core:standard}
We propose the following operators, \textbf{Optical blur} ($B$) models lens defocus, film-plane misalignment, and projection vibration. It uses an isotropic or anisotropic Gaussian kernel with parameters sampled per clip.
\textbf{JPEG compression} ($J$) mimics a digitalization pass.
\textbf{Resolution change} ($D$) downscales $s_\downarrow$ (with p=0.7), upscales $s_\uparrow$ (p=0.2) or identity (p=0.1), with the interpolation kernel drawn from \{cubic, bilinear, Lanczos-4\}.
\textbf{Gamma correction} ($\Gamma$) models frame-to-frame exposure variability with $\tilde{\mathbf{x}}_t = \mathbf{x}_t^{\gamma_t}$,
where $\gamma_t{=}\text{clip}(\exp(\mathcal{N}(0,\,0.125)), 0.2, 4)$, resampled
\emph{per frame} (the sole exception to the per-clip lock). For more details, see Table~\ref{tab:severity}.

\paragraph{Stage 5: Finalization.}
The per-frame pipeline closes with four fixed steps: (i) spatial resolution
is restored to $H{\times}W$; (ii) mild photometric jitter - brightness
$\mathcal{U}[0.8,1.2]$, contrast $\mathcal{U}[0.9,1.0]$ - is applied with
probability $0.5$, modelling analog transfer variability without assuming it is
always present; (iii) following~\citep{wan2022oldfilm}, the ground truth is sharpened with the soft-masked unsharp mask of Real-ESRGAN~\citep{wang2021realesrgan} (weight $0.50$, radius $51$, threshold $10$), so that target preprocessing is identical for both pipelines compared in Section~\ref{sec:experiments}; (iv) both the degraded frame and aligned ground truth are
quantized to 8-bit integer, matching real digitization output precision.

\paragraph{Stage 6: Mechanical scratches.}
\label{sec:method:mechanical_scratches}
Mechanical scratches, arising from abrasive contact with projector components or debris, manifest as persistent, thin vertical streaks. This reflects their nature as physical emulsion damage captured directly during scanning.

\paragraph{Geometry and rendering.}
Base scratch width is physically grounded ($[5,40]\,\mu\mathrm{m}$), yielding $1$--$3$\,px at 2000\,DPI. To approximate the Fraunhofer diffraction PSF of a narrow slit, the soft-edged scratch profile is rendered via a $\mathrm{sinc}^2$--Gaussian hybrid kernel:
$    k(x) \propto \mathrm{sinc}^2\!\!\left(\tfrac{x}{w/2}\right) \cdot\exp\!\!\left(-\tfrac{x^2}{2(1.5w)^2}\right).
\label{eq:scratch_kernel}
$
Width varies longitudinally via a cumulative random walk, while per-frame intensity flickers according to a CubicSpline envelope ($\pm20\%$).

\subsection{Temporal coherence}
\label{method:temporal_coherence}
We sample all stochastic parameters $\boldsymbol{\Theta}_d$ (e.g., noise type, grain amplitude, permutation order, JPEG quality) once per clip and fix them across all $T$ frames. To accurately reflect the consistent appearance of real film processed in a single pass, intra-clip variation is strictly limited to two physically motivated processes: the per-frame update of the Ornstein-Uhlenbeck gate weave (Section~\ref{sec:method:core:gateweave}) and the toroidal shift of the blue noise grain tile (Section~\ref{sec:method:core:grain}).

In case of mechanical scratches, a clip is scratched with probability $0.3$ and then receives $1$--$5$ scratches (lifespan $\mathcal{U}\{2,9\}$ frames, $0.3$ respawn probability). Unlike gate weave's mean-reverting motion, a scratch's lateral position follows a slow, periodic CubicSpline interpolation to simulate a fixed projector defect. Scratches are either \emph{bright} (emulsion abrasion, $p=0.6$) or \emph{dark} (surface contamination, $p=0.4$).

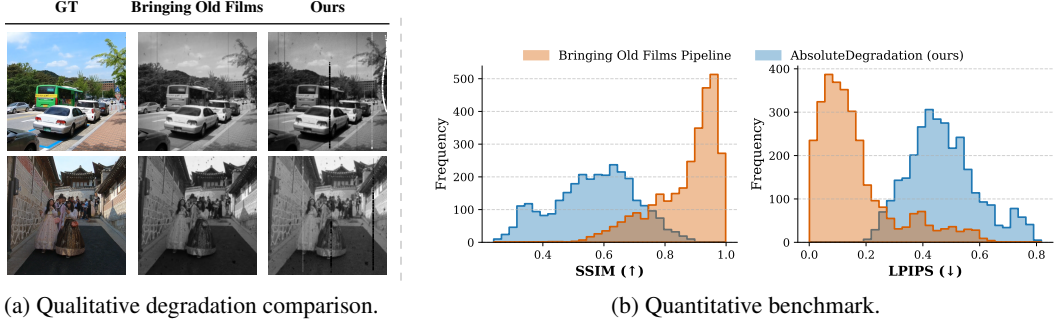
\begin{figure}[t]
  \centering
  \small

  \begin{minipage}[b]{0.35\textwidth}
    \centering
    \setlength{\tabcolsep}{1pt}
    \renewcommand{\arraystretch}{1.1}
    \def\gridpanel{0.32\linewidth}
     \tiny
    \begin{tabular}{ccc}
      \textbf{GT} & \textbf{Bringing Old Films} & \textbf{Ours} \\
      \midrule
      \includegraphics[width=\gridpanel]{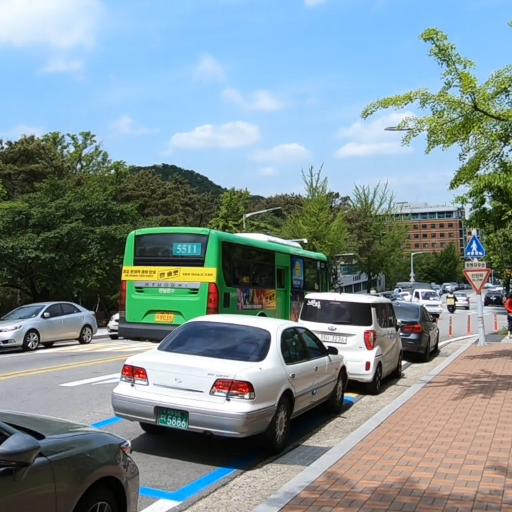} &
      \includegraphics[width=\gridpanel]{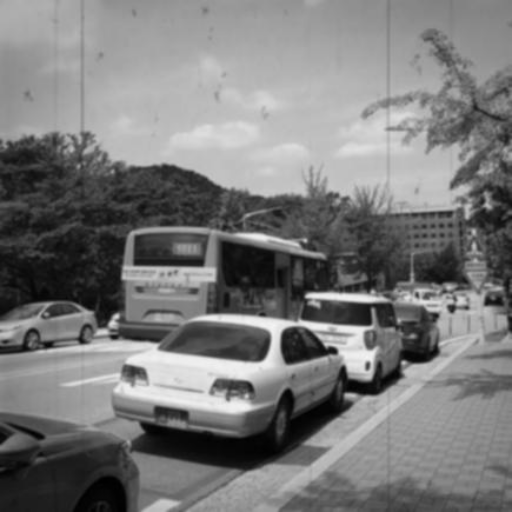} &
      \includegraphics[width=\gridpanel]{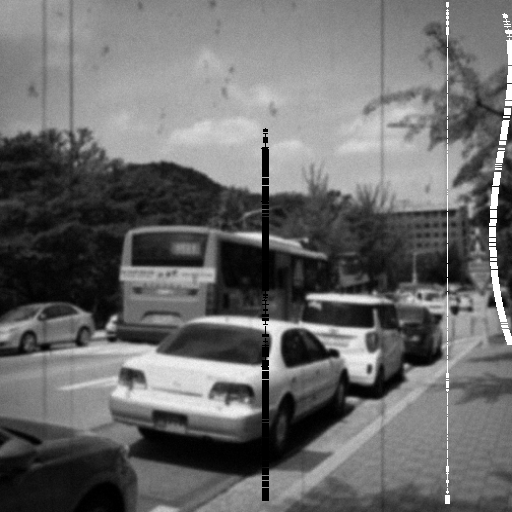} \\
      \includegraphics[width=\gridpanel]{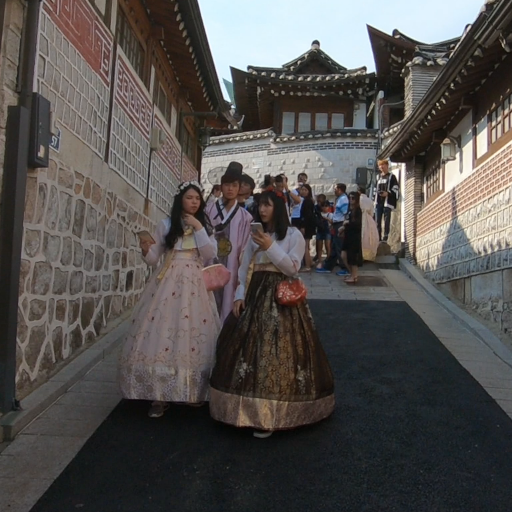} &
      \includegraphics[width=\gridpanel]{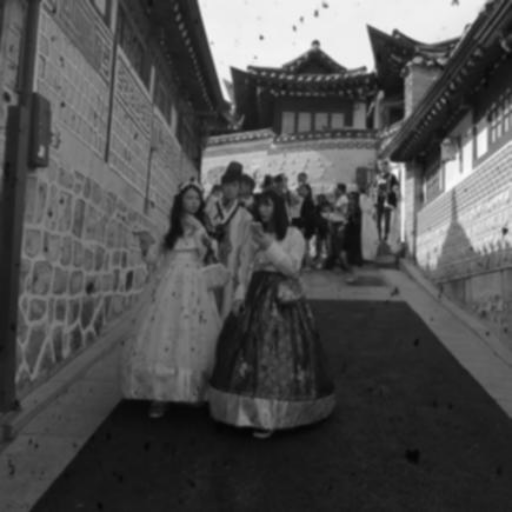} &
      \includegraphics[width=\gridpanel]{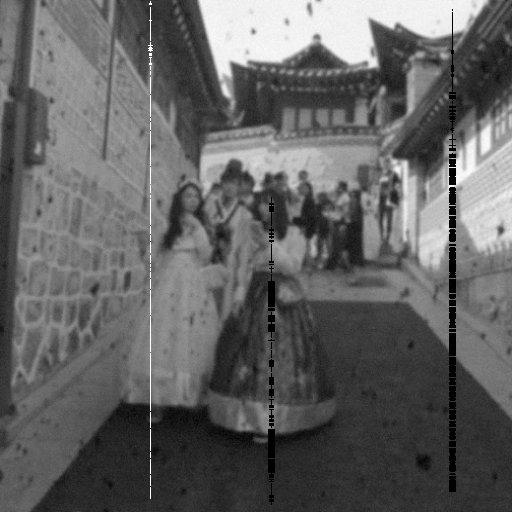} \\
    \end{tabular}
    \par\vspace{5pt} \small (a) Qualitative degradation comparison.
  \end{minipage}%
  \hfill
  \raisebox{12pt}{
    \begin{tikzpicture}
      \draw[dashed, gray!60, thin] (0,0) -- (0, 3.5cm);
    \end{tikzpicture}
  }%
  \hfill
  \begin{minipage}[b]{0.6\textwidth}
    \centering
    \includegraphics[width=1\linewidth]{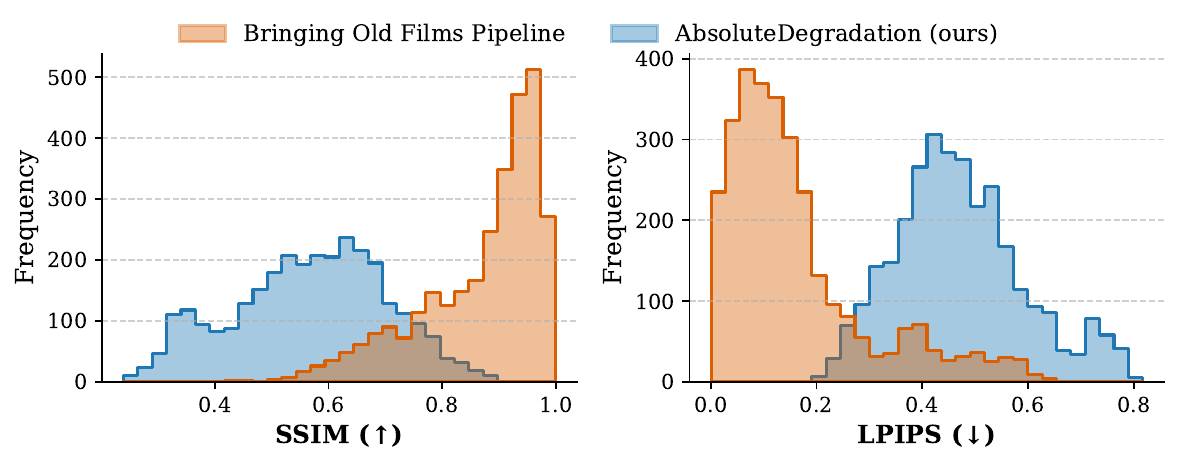}
    \par (b) Quantitative benchmark.
  \end{minipage}

  \caption{Qualitative and quantitative comparison of the proposed AbsoluteDegradation and Bringing Old Films~\citep{wan2022oldfilm} against REDS ground truth. Lower SSIM and higher LPIPS confirm our method applies more severe distortions, increasing restoration difficulty.}

  \label{fig:combined_comparison}
\end{figure}

\subsection{AbsoluteRestoration Synthetic Paired Dataset}
\label{sec:method:stats}

We construct a synthetic validation dataset by degrading the \emph{complete} REDS sharp validation split \citep{nah2019reds}. This provides 30 clips ($1270{\times}690$, 100 frames each), evenly partitioned into low, medium, and high severities. Furthermore, while we strongly advocate for online training, we additionally release a degraded version of the entire 240-clip REDS sharp training split for offline training. Figure~\ref{fig:combined_comparison}a depicts $512{\times}512$ crops from different REDS frames degraded using fixed medium-tier parameters. Figure~\ref{fig:combined_comparison}b reports aggregate quantitative differences on the full validation benchmark. Additional comparisons displayed in Figure \ref{fig:tier_comparison_fixed_frame}.

\section{Old Video Dataset}
\label{sec:old_video_dataset}

Evaluating restoration models on real archival footage requires a benchmark
that faithfully represents the target domain. Existing resources, such as the
real-world partition of SRWOV~\citep{mambaofr}, are limited by low spatial
resolution, lossy compression, heterogeneous content (including cartoons),
and the presence of artifacts such as watermarks that are not part of the
original analog degradation process. These issues reduce their suitability as
a domain-representative benchmark.

To address these limitations, we construct a new archival dataset from
30 public-domain films (1896--1918) sourced from the Library of Congress.
All footage is stored in lossless PNG format at resolutions between
$1440{\times}1080$ and $1920{\times}1080$, preserving the fidelity of the
original scans and ensuring that compression artifacts do not confound
evaluation.

\begin{wraptable}{r}{0.45\textwidth}

\centering
\scriptsize

\begin{minipage}{0.45\textwidth}
\caption{Quantitative comparison of degradation pipelines. Lower FID scores indicate that the synthetic data generated by our pipeline aligns better with the real-world distribution.}
\label{tab:fid-comparison}

\centering
\begin{tabular}{l cc}
\toprule
& \multicolumn{2}{c}{\textbf{FID Score $\downarrow$}} \\
\cmidrule(r){2-3}
\textbf{Method} & SRWOV \citep{mambaofr} & Our Dataset \\
\midrule
Bringing Old Films \citep{wan2022oldfilm}
& 193.84 & 166.04 \\

AbsoluteDegradation (ours)
& \textbf{185.78} & \textbf{154.27} \\
\bottomrule
\end{tabular}
\end{minipage}

\end{wraptable}

\paragraph{Data Processing}
\label{sec:old_video_dataset:acquisition}

We extract temporally coherent clips through automatic scene splitting
followed by manual curation. Scene boundaries are detected using feature
matching between frames built on SuperPoint~\citep{DeTone_2018_CVPR_Workshops} and LightGlue~\citep{Lindenberger_2023_ICCV}.
All clips are manually verified to remove title cards, blank frames, and
non-informative segments, ensuring consistent visual content within each clip.
Clips shorter than 100 frames are discarded.
Full details on the acquisition pipeline, curation protocol, and dataset
construction are provided in Appendix~\ref{sec:appendix-old-video-dataset}.

\paragraph{Dataset Variants}

From the curated footage, we construct three datasets for different evaluation
settings:
\textbf{(i) Full scene-split dataset.}
Contains \emph{81,576 frames} across single-scene clips, suitable for
evaluating temporally consistent restoration methods.
\textbf{(ii) Compact subset.}
A reduced evaluation set of \emph{13,252 frames}, obtained by selecting a
subset of clips and frames, enabling efficient benchmarking.
\textbf{(iii) Multi-scene dataset.}
A dataset of \emph{121,200 frames} formed by splitting videos into fixed-length
clips without scene segmentation, supporting evaluation under real-world scene transitions.

\label{sec:old_video_dataset:datasets}

\begin{figure}[t]
\centering

\begin{minipage}[b]{0.25\textwidth}
  \centering

  \begin{subfigure}{0.48\linewidth}
    \centering
    \includegraphics[width=\linewidth, height=2.5cm, keepaspectratio]{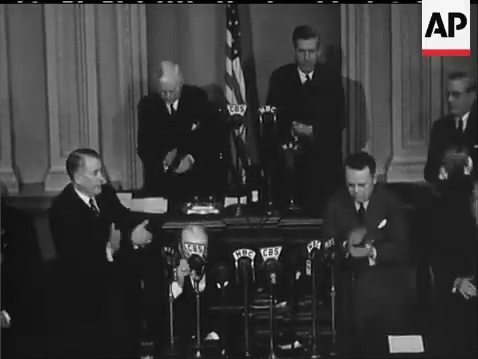}
  \end{subfigure}
  \begin{subfigure}{0.48\linewidth}
    \centering
    \includegraphics[width=\linewidth, height=2.5cm, keepaspectratio]{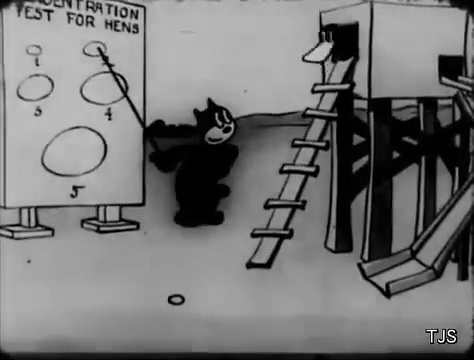}
  \end{subfigure}
  \begin{subfigure}{0.48\linewidth}
    \centering
    \includegraphics[width=\linewidth, height=2.5cm, keepaspectratio]{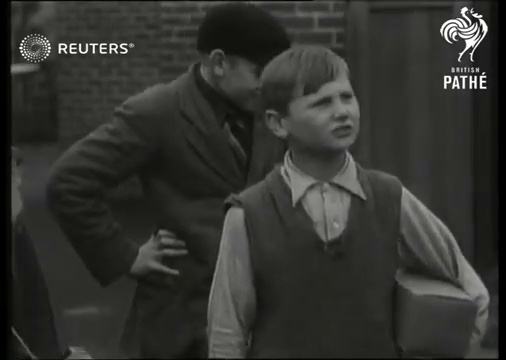}
  \end{subfigure}
  \begin{subfigure}{0.48\linewidth}
    \centering
    \includegraphics[width=\linewidth, height=2.5cm, keepaspectratio]{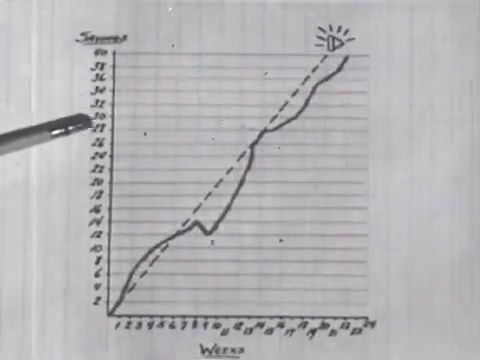}
  \end{subfigure}

  \vspace{0.2em}
  {\small (a) SRWOV dataset \citep{mambaofr}}
\end{minipage}%
\raisebox{12pt}{
\begin{tikzpicture}
  \draw[dashed, gray!60, thin] (0, 0) -- (0, 2.6cm);
\end{tikzpicture}
}%
\begin{minipage}[b]{0.25\textwidth}
  \centering

  \begin{subfigure}{0.48\linewidth}
    \centering
    \includegraphics[width=\linewidth, height=2.5cm, keepaspectratio]{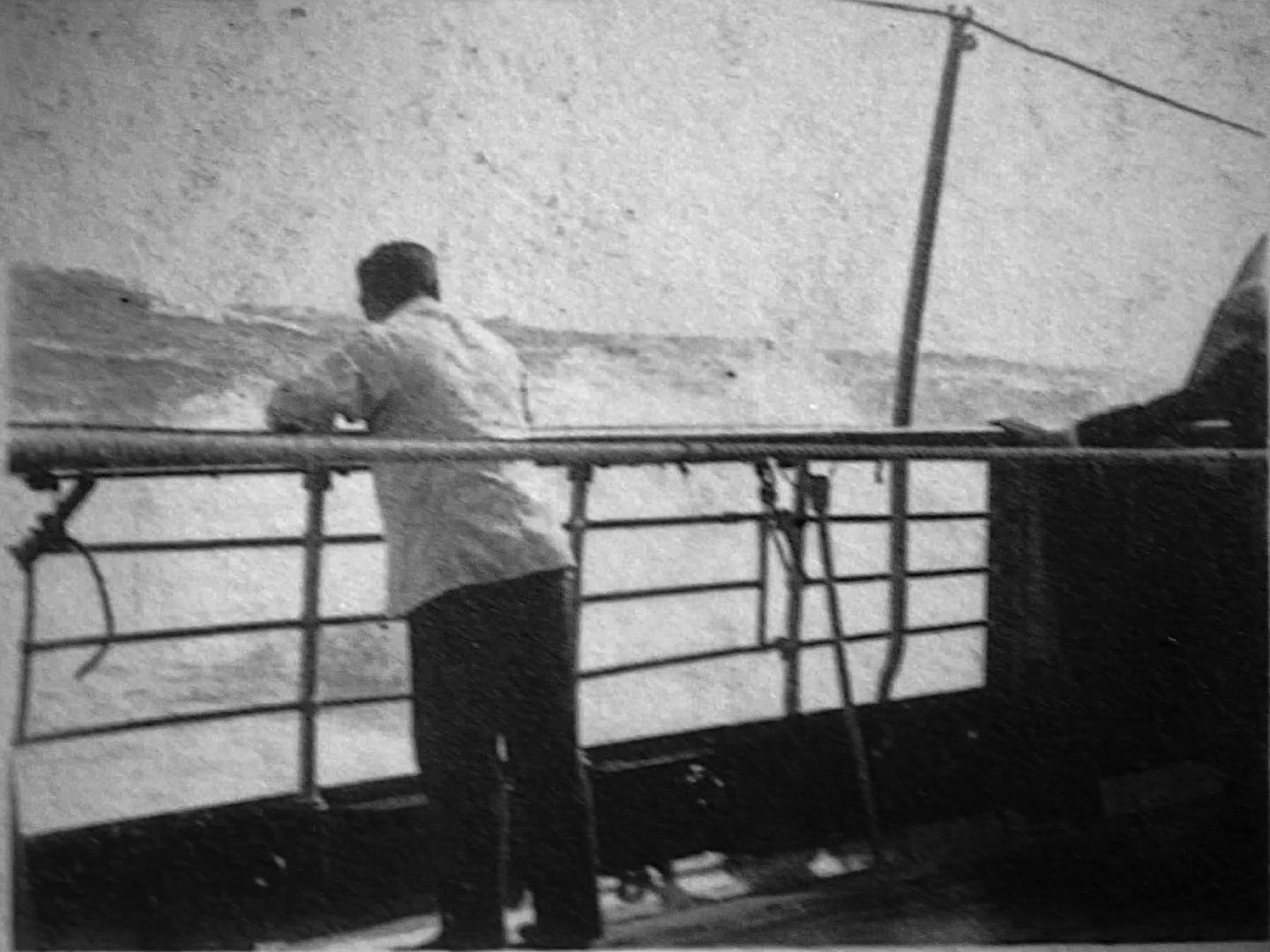}
  \end{subfigure}
  \begin{subfigure}{0.48\linewidth}
    \centering
    \includegraphics[width=\linewidth, height=2.5cm, keepaspectratio]{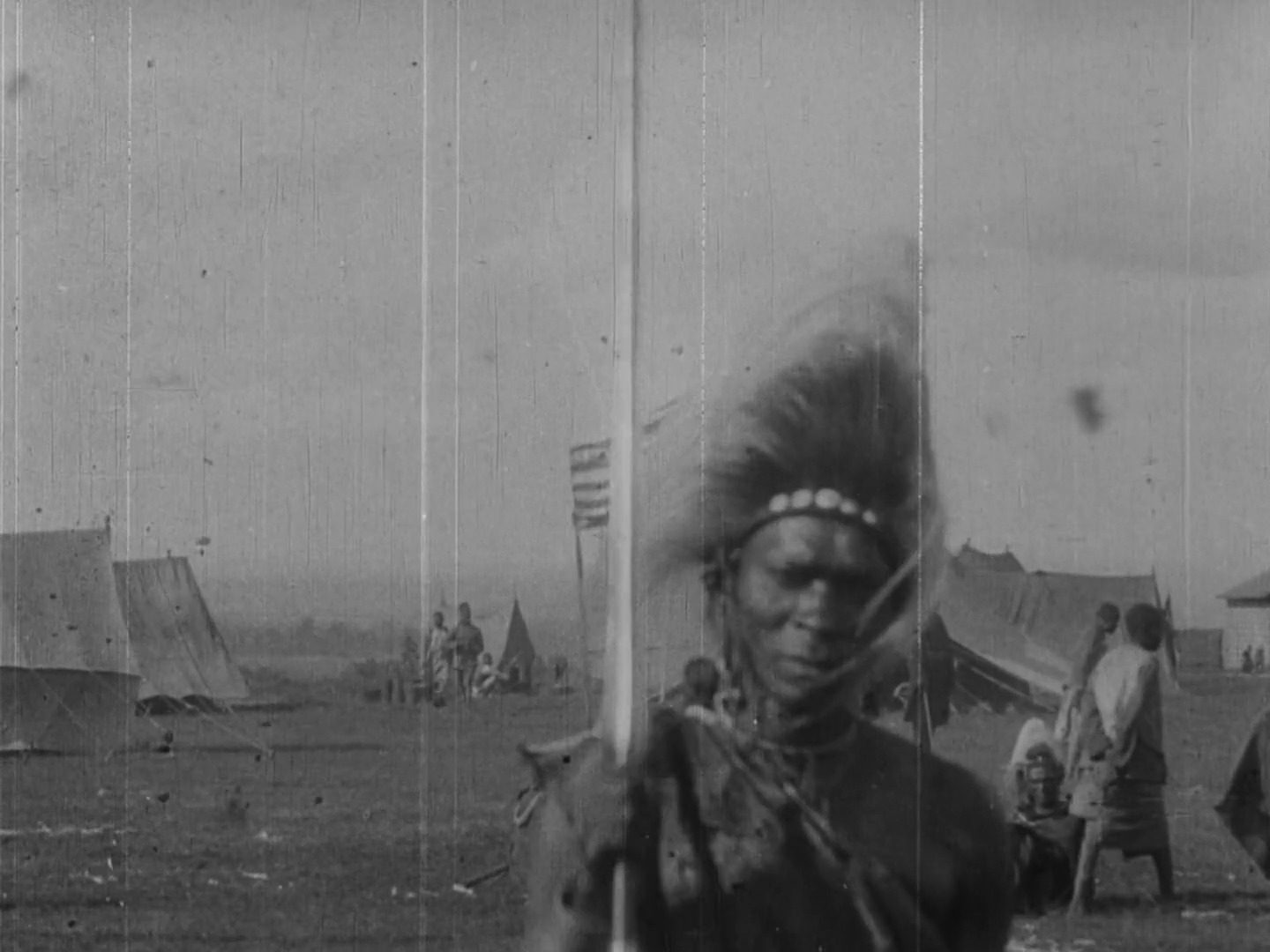}
  \end{subfigure}
  \begin{subfigure}{0.48\linewidth}
    \centering
    \includegraphics[width=\linewidth, height=2.5cm, keepaspectratio]{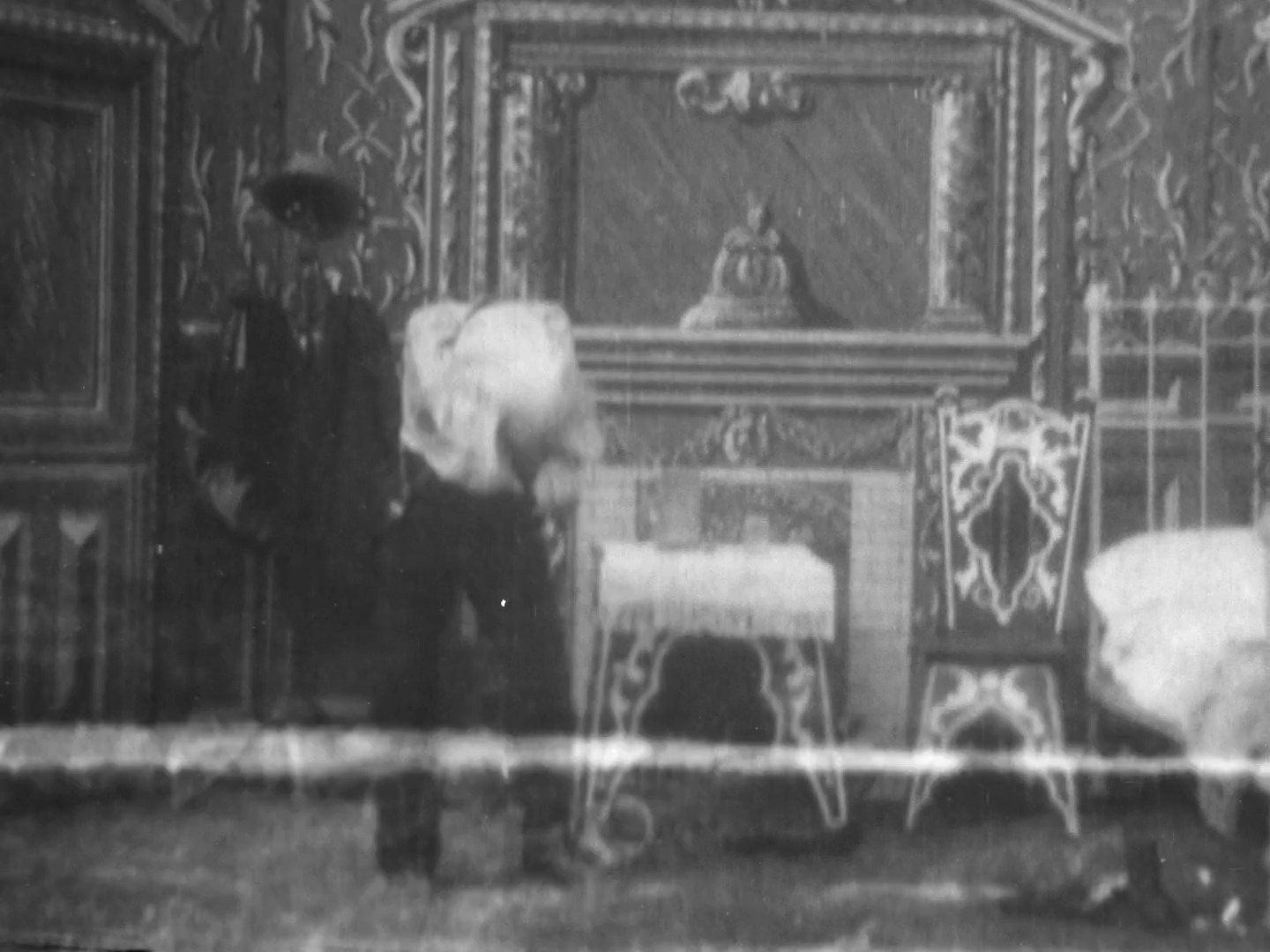}
  \end{subfigure}
  \begin{subfigure}{0.48\linewidth}
    \centering
    \includegraphics[width=\linewidth, height=2.5cm, keepaspectratio]{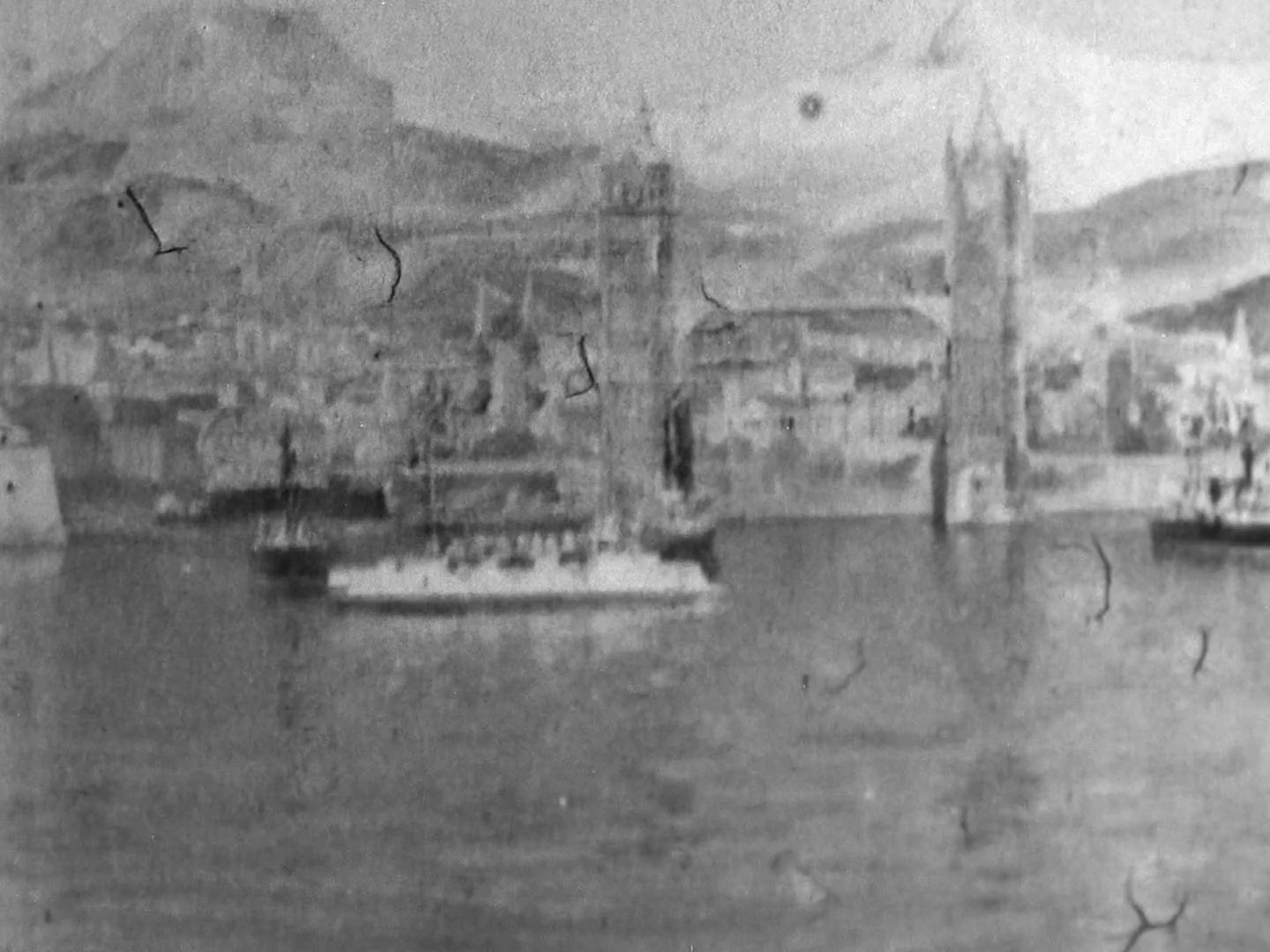}
  \end{subfigure}

  \vspace{0.2em}
  {\small (b) Our dataset}
\end{minipage}%
\raisebox{12pt}{
\begin{tikzpicture}
  \draw[dashed, gray!60, thin] (0, 0) -- (0, 2.6cm);
\end{tikzpicture}
}%
\begin{minipage}[b]{0.48\textwidth}
  \centering
  \includegraphics[width=\linewidth]{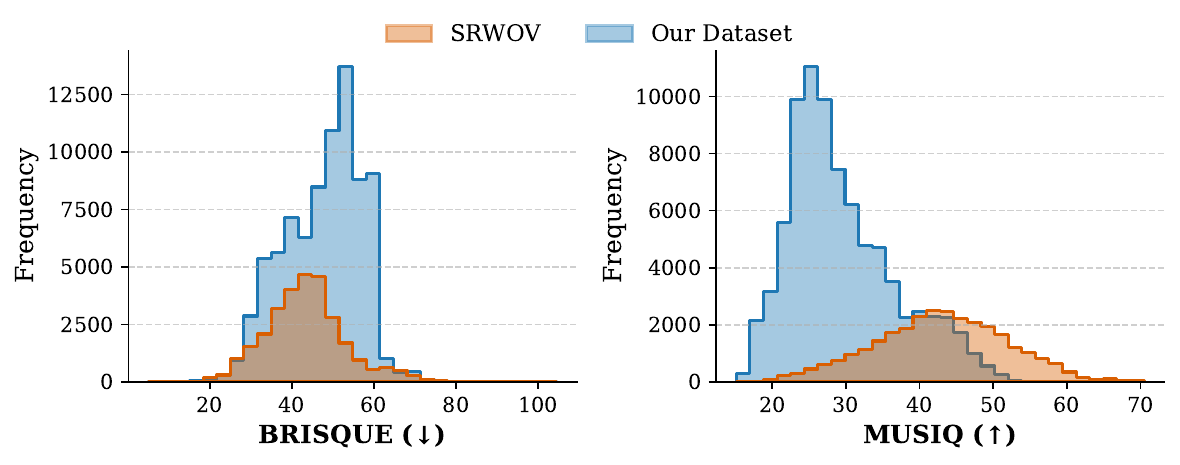}
  {\small (c) Quantitative comparison}
\end{minipage}%

\caption{Representative frames from \textbf{(a)} SRWOV \citep{mambaofr} and \textbf{(b)} our archival benchmark. \textbf{(c)} Quantitative comparison  of datasets. Unlike SRWOV, our dataset consists of high-resolution, lossless frames of genuine archival footage (rather than mixed content such as cartoons), with no overlay watermarks and substantially richer physical analog degradations, making it more domain-representative. Our benchmark exhibits higher BRISQUE and lower MUSIQ scores on average. These metrics indicate a lower overall frame quality compared to SRWOV, resulting in a more challenging restoration task.
}
\label{fig:samples}
\end{figure}
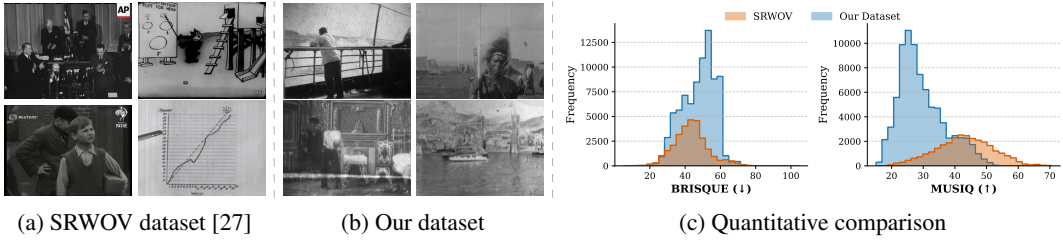

\paragraph{Comparison to Existing Datasets}

Figure~\ref{fig:samples} compares our dataset to SRWOV~\citep{mambaofr}. Our benchmark provides
higher resolution, lossless storage, and a consistent focus on real-world
archival footage. In contrast, SRWOV contains lower-resolution JPEG-compressed
frames, domain inconsistencies, and fewer instances of severe analog
degradations.

Quantitative comparisons using no-reference image quality metrics
(Figure~\ref{fig:samples}c) show that our dataset exhibits lower
perceptual quality scores, indicating more challenging degradation conditions.
Additionally, FID analysis (Table~\ref{tab:fid-comparison}) demonstrates that
synthetic data generated by our pipeline aligns more closely with the
distribution of this dataset than prior methods, supporting its use as a
realistic evaluation benchmark. No frames of the benchmark were examined while choosing pipeline parameters, and the FID values were computed only after the pipeline was fixed. For more details regarding FID evaluation see Appendix~\ref{sec:appendix-fid}.

\section{Experiments and Results}
\label{sec:experiments}

\subsection{Experimental Setup}
For the task of old video restoration, we evaluate our AbsoluteDegradation pipeline against the Bringing Old Films~\citep{wan2022oldfilm} degradation generation pipeline baseline. We span our data pipeline evaluation across six models tested on two real-world datasets of archival film footage, the proposed dataset and SRWOV dataset~\citep{mambaofr}. For every architecture, the two training conditions differ only in the degradation pipeline: clean sources, target preprocessing (Section~\ref{sec:method}, Stage~5), and training protocol are identical.

\paragraph{Datasets and metrics}
We utilize our newly collected benchmark of 81,576 lossless-PNG frames curated from the Library of Congress, alongside the real-world partition of the SRWOV benchmark~\citep{mambaofr}. Due to the lack of pristine, clean images we evaluate using non-reference metrics averaged over frames: BRISQUE~\citep{mittal2012brisque}, CLIPIQA+~\citep{wang2023clipiqa}, MUSIQ~\citep{ke2021musiq}, and MANIQA~\citep{yang2022maniqa}. Because per-frame scores cannot detect flicker or inconsistent artifact removal, we additionally assess each restored sequence with the clip-level DOVER~\citep{wu2023dover} video quality model, whose technical branch is sensitive to temporal distortions. Since learned quality scores can themselves be misaligned with restoration authenticity, we finally conduct a blinded human preference study (Section~\ref{sec:human}).

\paragraph{Comparison methods}
To systematically compare the two degradation pipelines, we train and evaluate six architectures on both pipelines and datasets: three general-purpose video restoration models, BasicVSR++~\citep{chan2022basicvsrplusplus}, RVRT~\citep{liang2022rvrt}, and Shift-Net~\citep{li2023shiftnet}, and three models designed for old-film restoration, MambaOFR~\citep{mambaofr}, RTN~\citep{wan2022oldfilm}, and DeepRemaster~\citep{iizuka2019deepremaster}. All models follow their original default training configurations, so that no architecture is tuned in favour of either pipeline. For restoration adaptation, RVRT uses its video deblurring setup, and BasicVSR++ employs the REDS4-style setting from its official repository. Furthermore, we modify BasicVSR++ by removing its upsampling layers and replacing the original SpyNet~\citep{ranjan2017optical} optical flow backend with RAFT~\citep{teed2020raft}. Implementation details are given in Appendix~\ref{sec:appendix-implementation}.

\subsection{Results}

\begin{table}[t]
\centering
\caption{Quantitative comparison of restorers trained with the Bringing Old Films (BOF)~\citep{wan2022oldfilm} versus AbsoluteDegradation (Ours) pipeline, evaluated on \textbf{our dataset} and \textbf{SRWOV dataset}~\citep{mambaofr}. Bold indicates the better pipeline for each model; underlined indicates the best result in each metric column. $^{\dagger}$RVRT is not fully trained (4 days on $4{\times}$A100 vs.\ 9.7 days on $8{\times}$A100 reported by its authors; see Appendix~\ref{sec:appendix-implementation}).}
\label{tab:results}
{\scriptsize\setlength{\tabcolsep}{3pt}%
\renewcommand{\arraystretch}{1.1}
\resizebox{\linewidth}{!}{%
\begin{tabular}{@{}ll|cccc|cccc@{}}
\Xhline{2.5\arrayrulewidth}
& & \multicolumn{4}{c|}{\textbf{Our dataset}} & \multicolumn{4}{c}{\textbf{SRWOV dataset~\citep{mambaofr}}} \\
\Xhline{1\arrayrulewidth}
\textbf{Model} & \textbf{Pipeline} & \textbf{BRISQUE}~$\downarrow$ &  \textbf{CLIPIQA+}~$\uparrow$ & \textbf{MUSIQ}~$\uparrow$ & \textbf{MANIQA}~$\uparrow$ & \textbf{BRISQUE}~$\downarrow$ &  \textbf{CLIPIQA+}~$\uparrow$ & \textbf{MUSIQ}~$\uparrow$ & \textbf{MANIQA}~$\uparrow$ \\
\Xhline{1\arrayrulewidth}
\multirow{2}{*}{\textbf{BasicVSR++}~\citep{chan2022basicvsrplusplus}} & BOF~\citep{wan2022oldfilm} & 30.62 & 0.313 & 32.55 & 0.170 & 26.20 & 0.397 & 49.17 & 0.245 \\
 & Ours & \textbf{20.68} & \textbf{0.369} & \textbf{42.91} & \textbf{0.238} & \textbf{24.99} & \textbf{0.443} & \textbf{58.04} & \textbf{0.323} \\
\Xhline{1\arrayrulewidth}
\multirow{2}{*}{\textbf{Shift-Net}~\citep{li2023shiftnet}} & BOF~\citep{wan2022oldfilm} & 26.74 & 0.370 & 36.74 & 0.213 & 28.00 & 0.419 & 50.64 & 0.284 \\
 & Ours & \textbf{16.59} & \underline{\textbf{0.431}} & \textbf{46.77} & \textbf{0.276} & \textbf{25.59} & \underline{\textbf{0.458}} & \textbf{57.93} & \underline{\textbf{0.334}} \\
\Xhline{1\arrayrulewidth}
\multirow{2}{*}{\textbf{MambaOFR}~\citep{mambaofr}} & BOF~\citep{wan2022oldfilm} & 32.79 & 0.347 & 38.81 & 0.221 & 27.39 & 0.398 & 50.89 & 0.271 \\
 & Ours & \textbf{20.28} & \textbf{0.419} & \underline{\textbf{47.24}} & \underline{\textbf{0.278}} & \textbf{20.73} & \textbf{0.440} & \textbf{57.36} & \textbf{0.321} \\
\Xhline{1\arrayrulewidth}
\multirow{2}{*}{\textbf{RTN}~\citep{wan2022oldfilm}} & BOF~\citep{wan2022oldfilm} & 19.89 & 0.366 & 41.46 & 0.220 & 19.86 & 0.436 & 56.42 & 0.300 \\
 & Ours & \underline{\textbf{14.39}} & \textbf{0.422} & \textbf{46.87} & \textbf{0.266} & \underline{\textbf{19.24}} & \textbf{0.444} & \underline{\textbf{58.46}} & \textbf{0.332} \\
\Xhline{1\arrayrulewidth}
\multirow{2}{*}{\textbf{DeepRemaster}~\citep{iizuka2019deepremaster}} & BOF~\citep{wan2022oldfilm} & 35.52 & 0.320 & 39.48 & 0.225 & 37.04 & \textbf{0.399} & \textbf{57.04} & \textbf{0.312} \\
 & Ours & \textbf{23.33} & \textbf{0.357} & \textbf{44.75} & \textbf{0.254} & \textbf{20.90} & 0.372 & 53.55 & 0.264 \\
\Xhline{2\arrayrulewidth}
\multirow{2}{*}{\textbf{RVRT}$^{\dagger}$~\citep{liang2022rvrt}} & BOF~\citep{wan2022oldfilm} & \textbf{38.29} & \textbf{0.315} & \textbf{40.28} & \textbf{0.237} & \textbf{32.98} & 0.402 & 48.83 & \textbf{0.272} \\
 & Ours & 38.55 & 0.300 & 35.55 & 0.200 & 34.23 & \textbf{0.410} & \textbf{49.66} & 0.270 \\
\Xhline{1\arrayrulewidth}
\multicolumn{2}{@{}l|}{Real Degraded} & 47.86 & 0.248 & 26.84 & 0.162 & 43.34 & 0.389 & 42.86 & 0.231 \\
\Xhline{2.5\arrayrulewidth}
\end{tabular}%
}
}
\end{table}

\paragraph{Quantitative comparisons.}
Table~\ref{tab:results} reports results on SRWOV~\citep{mambaofr} and our dataset. For the five fully trained architectures, models trained with AbsoluteDegradation improve 37 of the 40 metric entries, on both benchmarks; the three exceptions are DeepRemaster on SRWOV, which we report rather than omit. The gains hold across very different temporal mechanisms. RVRT, which could not be fully trained within our compute budget, is the only model whose no-reference scores on our dataset favour the baseline; as shown below, human raters nevertheless prefer its AbsoluteDegradation-trained outputs. The improvement is robust to the random seed: over four BasicVSR++ runs per pipeline, AbsoluteDegradation wins on every metric for every seed (Appendix~\ref{sec:appendix-seeds}).

Crucially, quantitative metrics are not entirely reliable for evaluating this specific task. BRISQUE fails to reliably track perceptual quality; baseline pipelines sometimes achieve "better" BRISQUE scores while simultaneously dropping in CLIPIQA+ and MUSIQ, highlighting an inverse relationship between statistical sharpness and genuine quality. In Figure~\ref{fig:qualitative_comparisons}, the baseline obtains higher full-frame CLIPIQA+/MUSIQ in both examples, while visibly preserving or amplifying scratches and edge artifacts, making the failure mode directly falsifiable. This is precisely why we do not rely on learned quality scores alone and complement them with clip-level temporal assessment and a blinded human preference study.

\paragraph{Human evaluation.}
\label{sec:human}
\begin{wraptable}{r}{0.47\textwidth}
\centering
\scriptsize
\begin{minipage}{0.47\textwidth}
\caption{Blinded 2AFC human preference study on our dataset: share of trials in which the AbsoluteDegradation-trained restoration was preferred over the Bringing Old Films-trained one (Wilson 95\% CI, exact binomial test).}
\label{tab:human}
\centering
\setlength{\tabcolsep}{3pt}
\begin{tabular}{@{}lcccc@{}}
\toprule
\textbf{Architecture} & \textbf{Trials} & \textbf{Win rate} & \textbf{95\% CI} & $p$ \\
\midrule
BasicVSR++   & 237 & 89.9\% & [85.4, 93.1] & $<10^{-4}$ \\
Shift-Net    & 228 & 71.5\% & [65.3, 77.0] & $<10^{-4}$ \\
RVRT         & 229 & 59.8\% & [53.4, 66.0] & 0.0036 \\
DeepRemaster & 234 & 56.8\% & [50.4, 63.0] & 0.0425 \\
MambaOFR     & 231 & 55.8\% & [49.4, 62.1] & 0.0869 \\
RTN          & 236 & 54.2\% & [47.9, 60.5] & 0.2161 \\
\midrule
\textbf{All} & 1395 & \textbf{64.7\%} & [62.2, 67.2] & $<10^{-4}$ \\
\bottomrule
\end{tabular}
\end{minipage}
\vspace{-8pt}
\end{wraptable}
In a blinded two-alternative forced-choice (2AFC) study, for each of the six architectures and 29 clips from our benchmark, raters saw the degraded input as a reference above the baseline-trained and AbsoluteDegradation-trained restorations of the same architecture, playing in sync in a loop so that flicker was directly visible; sides were randomised and no method was identifiable. Two independently recruited rater groups (Prolific and direct invitation) completed 93 surveys with 1,395 trials. AbsoluteDegradation was preferred in 64.7\% of all trials, and \emph{for every one of the six architectures} (Table~\ref{tab:human}; sign test across architectures, $p=0.031$). The size of the benefit varies: it is largest for the general-purpose models (BasicVSR++, Shift-Net, RVRT; pooled 73.9\%), whose priors about film degradation must be learned entirely from training data, and smaller but still significant for the three models designed for old-film restoration (pooled 55.6\%, $p=0.0032$); individually, MambaOFR and RTN are not significant at this sample size. Notably, RVRT, the only model worse on every no-reference metric on our dataset, is significantly preferred by raters (59.8\%). Full protocol, instructions, and the study interface are given in Appendix~\ref{sec:appendix-human}.

\paragraph{Temporal consistency.}
On restored sequences, AbsoluteDegradation improves the clip-level DOVER-technical score, the branch sensitive to flicker and temporal distortion, in 16 of 18 model--dataset comparisons, including all six architectures on our dataset, and DOVER-overall in 14 of 18. Inter-frame self-similarity measures (tLPIPS~\citep{chu2020tecogan}, CLIP-F), which are maximised by temporal over-smoothing, mostly favour the baseline on real footage; on the synthetic set, AbsoluteDegradation wins tLPIPS for five of six models. Full tables are in Appendix~\ref{sec:appendix-temporal}.

\paragraph{Qualitative comparisons.}

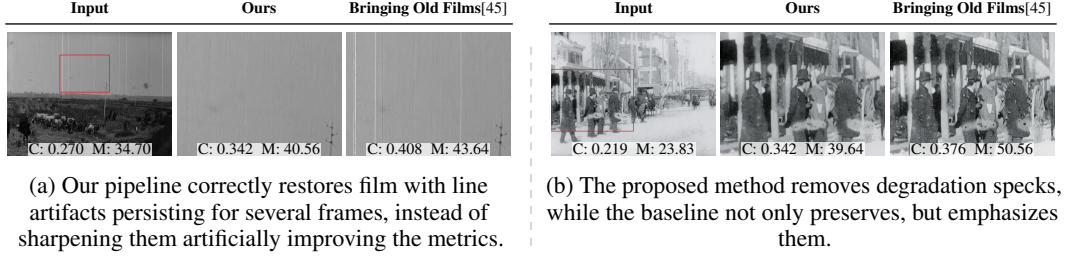
\begin{figure}[t]
  \centering
  \small

  \newcommand{\metricimg}[2]{%
    \begin{tikzpicture}
      \node[inner sep=0pt] (img) {\includegraphics[width=\gridpanel, height=2.15cm, keepaspectratio]{#1}};
      \node[anchor=south, yshift=-0.05ex, fill=white, fill opacity=0.75, text opacity=1,
            font=\tiny, inner sep=0.35pt, align=center] at (img.south) {#2};
    \end{tikzpicture}%
  }

  \begin{minipage}[b]{0.485\linewidth}
    \centering
    \setlength{\tabcolsep}{1pt}
    \def\gridpanel{0.32\linewidth}
    \tiny
    \begin{tabular}{ccc}
      \textbf{Input} & \textbf{Ours} & \textbf{Bringing Old Films}\citep{wan2022oldfilm} \\
      \midrule
      \metricimg{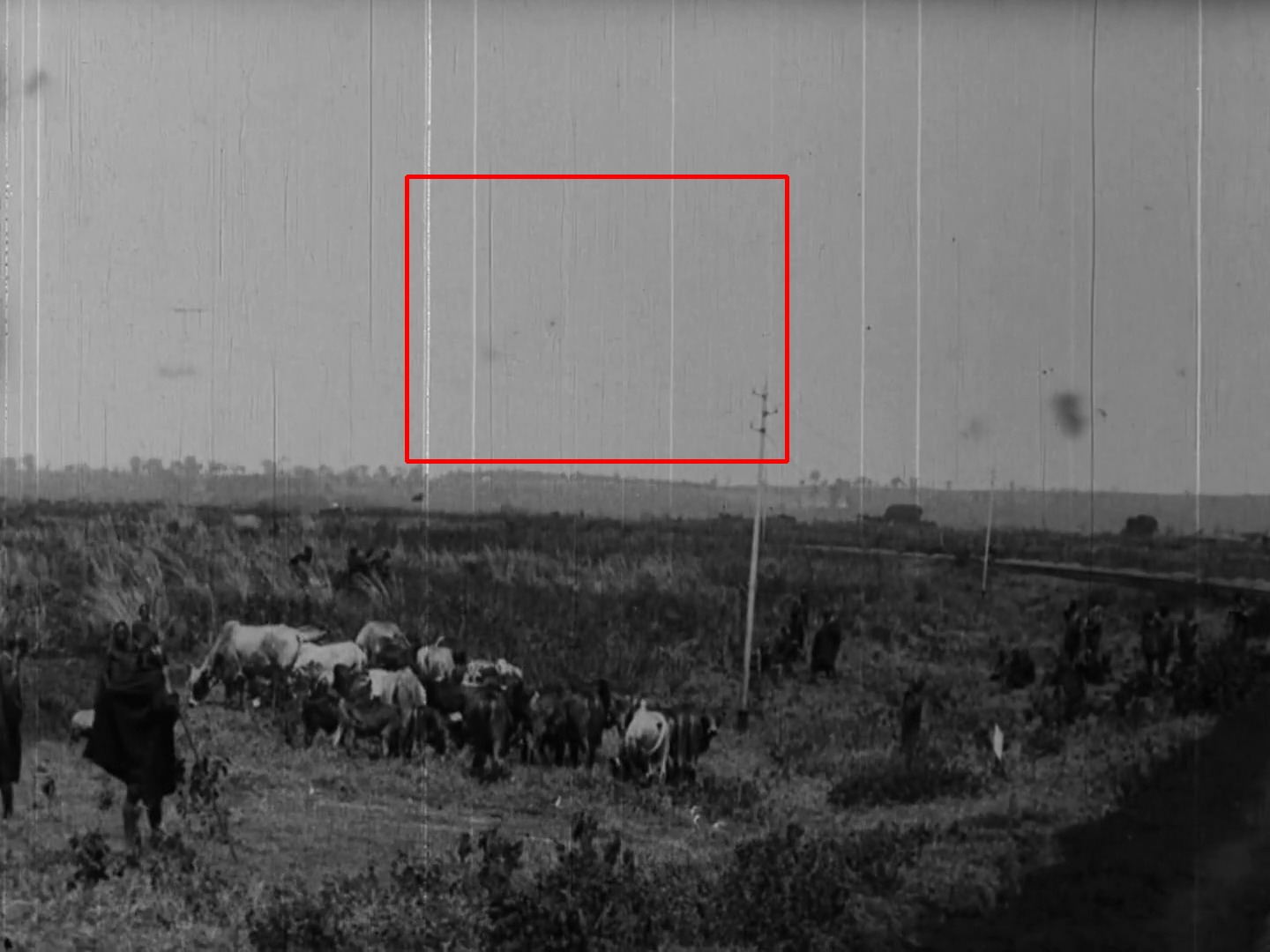}{C: 0.270\; M: 34.70} &
      \metricimg{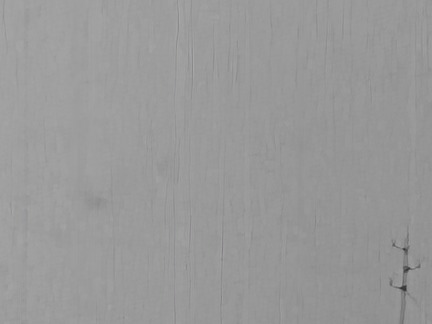}{C: 0.342\; M: 40.56} &
      \metricimg{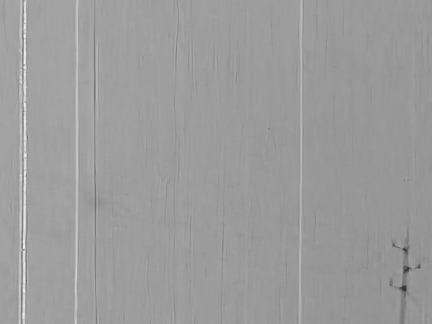}{C: 0.408\; M: 43.64} \\
    \end{tabular}
    \par\vspace{5pt} \small (a) Our pipeline correctly restores film with line artifacts persisting for several frames, instead of sharpening them artificially improving the metrics.
  \end{minipage}%
  \hfill
  \begin{minipage}[b]{0.01\linewidth}
    \centering
    \begin{tikzpicture}
      \draw[dashed, gray!60, thin] (0,0) -- (0, 2.65cm);
    \end{tikzpicture}
  \end{minipage}%
  \hfill
  \begin{minipage}[b]{0.485\linewidth}
    \centering
    \setlength{\tabcolsep}{1pt}
    \def\gridpanel{0.32\linewidth}
    \tiny
    \begin{tabular}{ccc}
      \textbf{Input} & \textbf{Ours} & \textbf{Bringing Old Films}\citep{wan2022oldfilm} \\
      \midrule
      \metricimg{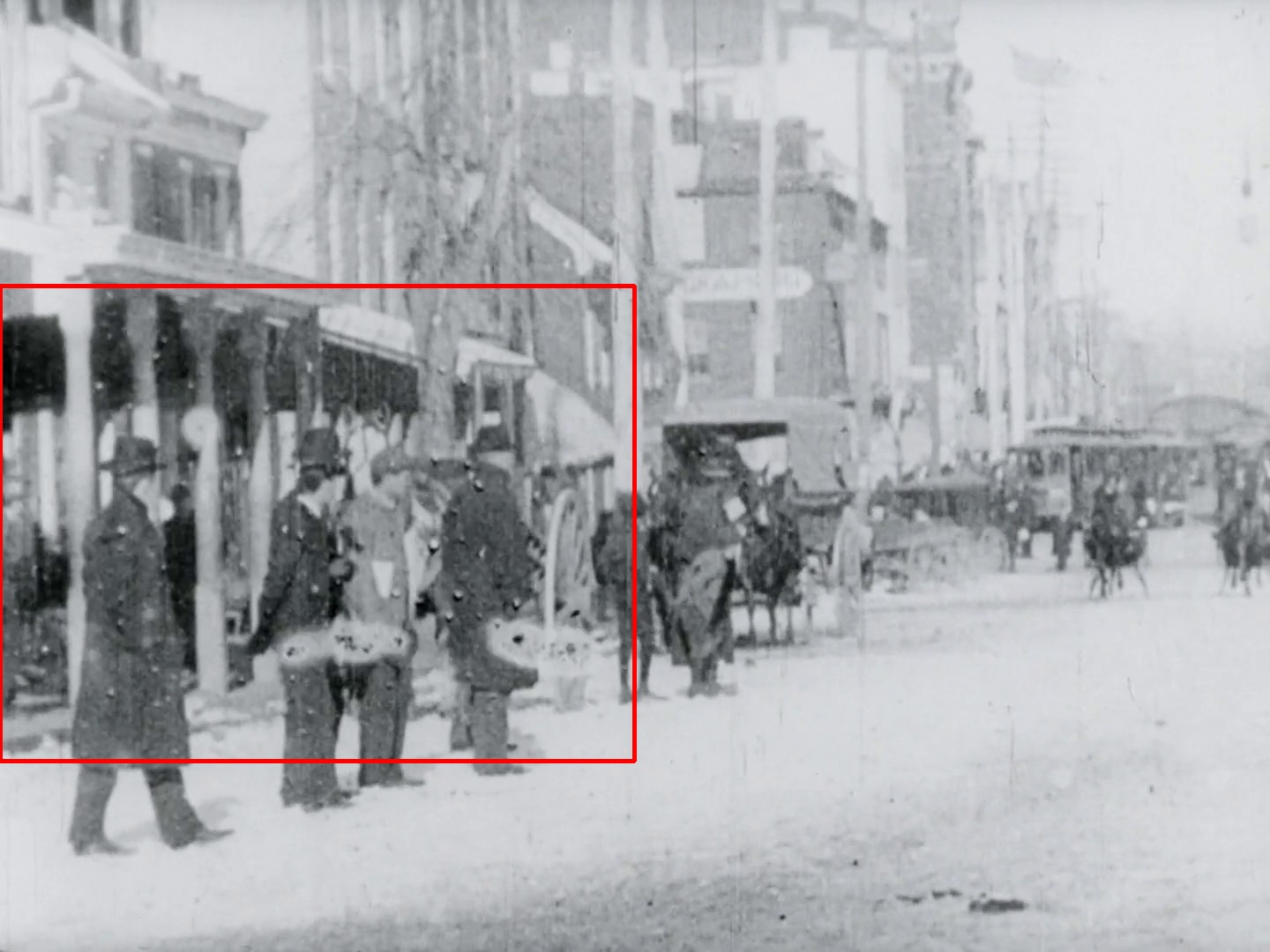}{C: 0.219\; M: 23.83} &
      \metricimg{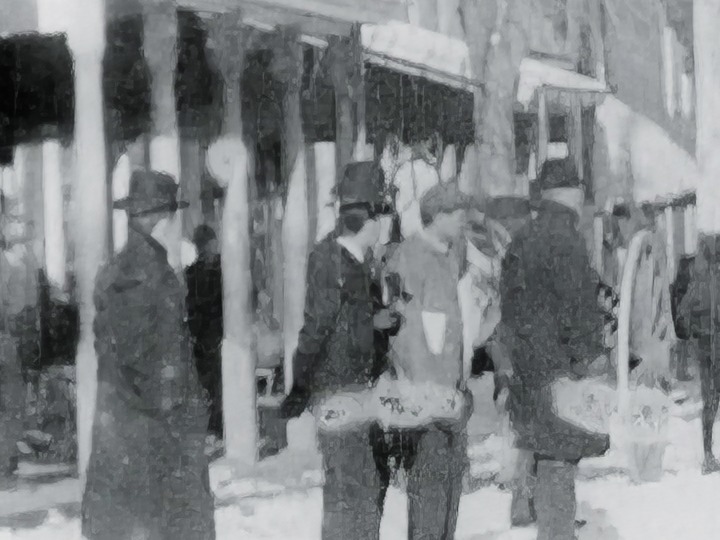}{C: 0.342\; M: 39.64} &
      \metricimg{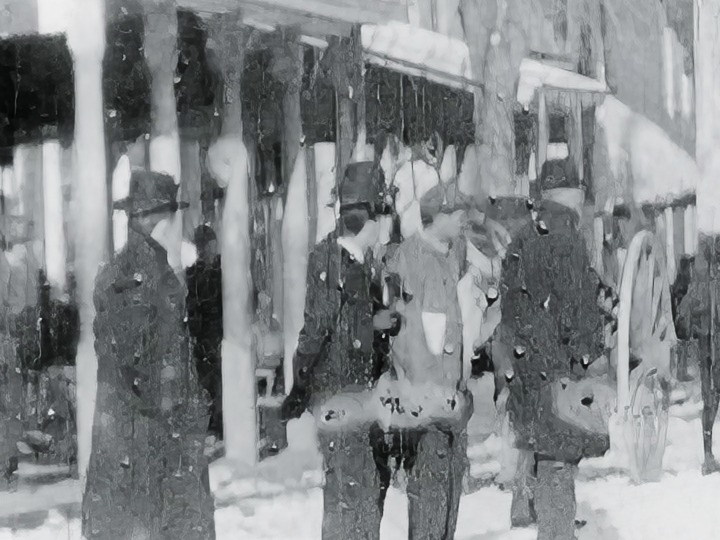}{C: 0.376\; M: 50.56} \\
    \end{tabular}
    \par\vspace{5pt} \small (b) The proposed method removes degradation specks, while the baseline not only preserves, but emphasizes them.

  \end{minipage}

  \vspace{-0.3em}
  \caption{Qualitative analysis of restored images using our pipeline vs. Bringing Old Films~\citep{wan2022oldfilm} on MambaOFR\citep{mambaofr} model. Metrics calculated on the entire restored frame (C: CLIP-IQA+ $\uparrow$, M: MUSIQ $\uparrow$), zoom in for better comparison.}
  \label{fig:qualitative_comparisons}
  \vspace{-0.8em}
\end{figure}

Our pipeline guides networks towards authentic, temporally consistent restorations without the over-sharpened edges typical of the baseline (Figure~\ref{fig:qualitative_comparisons}).

Signal-dependent blue-noise grain preserves the fine texture of the film, whereas the baseline's purely Gaussian noise causes networks to over-smooth real footage (see Ablation studies below). Temporally coherent training also makes networks cautious with unrecognized dust, hair, or complex textures: they do not sharpen and enhance them, unlike the baseline, which inflates metric scores through local sharpness while degrading visual quality (Figure~\ref{fig:qualitative_comparisons}). Our models further keep global exposure spatially and temporally consistent, avoiding the baseline's randomly darkened or overexposed patches. The goal of archival restoration is preserving authentic appearance, not unnatural enhancement. For more examples, see Figures~\ref{fig:restoration_comparison_srwov}, \ref{fig:restoration_comparison_duel}, and \ref{fig:restoration_comparison_roosevelt}.

\paragraph{Ablation studies.}
\label{ablation_studies}
To measure individual pipeline contributions on the architecture that benefits most from AbsoluteDegradation, we ablate BasicVSR++ by replacing our signal-dependent blue noise with \textit{Gaussian grain only} and by dropping \textit{moving lines} (Table~\ref{tab:ablations_basicvsr} in Appendix~\ref{sec:appendix-ablations}). The full pipeline achieves the best CLIPIQA+, MUSIQ, and MANIQA. Gaussian-only grain obtains the best BRISQUE while performing worse on all three learned metrics, which again indicates that BRISQUE favours over-smoothing: Gaussian noise alone makes the network flatten fine film texture, whereas signal-dependent blue noise preserves minor structures. A broader RTN ablation suite (gate-weave, negative-film distortions, texture persistence, severity curriculum) with qualitative examples is reported in Appendix~\ref{sec:appendix-ablations}.

\paragraph{Limitations.}
\label{sec:limitations}

Although AbsoluteDegradation demonstrates consistent improvements, it has several
limitations. \textit{Scope:} the benchmark and the released models target early black-and-white film (1896--1918) and grayscale restoration; we make no claim about colour-film chemistry, VHS, or other legacy media, and our evidence beyond this window (1950s--60s archival footage) is qualitative only. \textit{Physical fidelity:} operators are physically motivated, but their parameters were tuned by qualitative comparison with archival material; we do not yet compare per-operator statistics between synthetic and real footage, and the finite selection of textures can result in out-of-distribution failures on unique real-world artifacts. \textit{Evaluation:} because archival restoration is unpaired, no-reference metrics remain an imperfect proxy; we therefore base our conclusions on the agreement of NR-IQA, clip-level temporal assessment, and human preference, but the human study is limited to 29 clips of our benchmark and cannot individually resolve small effects (MambaOFR, RTN). \textit{Training budget:} all architectures use their default configurations, and the size of the benefit varies across models; RVRT could not be fully trained within our compute budget, so its results should not be read as reflecting the architecture's full capacity. \textit{Attribution and compute:} because video-scale training with online degradation is expensive, component ablations were run on two architectures only and cannot fully separate the contribution of each design choice.
Societal impact is positive through heritage preservation, public access, and reproducible evaluation, but the same tools could be used to over-restore archival material and distort its authentic appearance.
\section{Conclusion and Future Works}
\label{sec:conclusion}

In this work, we introduced \textbf{AbsoluteDegradation}, a physics-inspired synthetic degradation pipeline that models the analog-to-digital process through structured, temporally coherent artifact families. To resolve the scarcity of standardized evaluation data, we also presented a curated benchmark of 81,576 high-resolution, lossless frames sourced from real archival footage. Across six architectures, no-reference metrics, clip-level temporal assessment, and a blinded human preference study consistently show that our pipeline improves generalization to real archival footage for every evaluated architecture, to a varying degree, while exposing no-reference metrics that reward outputs human raters reject.
Future work can focus on: (1) calibrating operators against measured statistics of real degradations; (2) extending the modular pipeline to other eras and media, such as color film, biological decay, or magnetic tape; (3) domain-specific reference-free metrics; and (4) unpaired video-to-video domain translation.

\begin{ack}
\noindent
\begin{minipage}[c]{0.09\linewidth}
  \includegraphics[width=\linewidth]{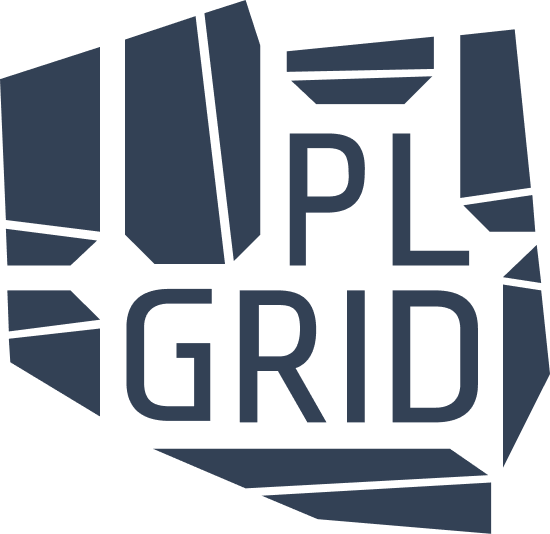}
\end{minipage}\hfill
\begin{minipage}[c]{0.89\linewidth}
We gratefully acknowledge Polish high-performance computing infrastructure PLGrid (HPC Center: ACK Cyfronet AGH) for providing computer facilities and support within computational grant no. PLG/2026/019902.
\end{minipage}

The film grain and dust texture assets used in this pipeline are adopted from
\emph{Bringing Old Films Back to Life}~\citep{wan2022oldfilm}. Ground-truth sequences follow
the REDS dataset format~\citep{nah2019reds}. We thank the authors of both works
for making their data publicly available.

The authors declare no competing interests.
\end{ack}

\clearpage
\bibliographystyle{abbrvnat}
\bibliography{references}

\clearpage
\appendix

\section{Related Work}
\label{sec:appendix-related-works}

\subsection{Restoration Methods: From Images to Video}

Image restoration has evolved from model-based priors over natural image
statistics~\citep{elad2006image,weiss2007makes} to task-specific discriminative
networks for denoising~\citep{zhang2017dncnn,zhang2018ffdnet,lefkimmiatis2018universal},
deblurring~\citep{xu2014deep,sun2015learning},
super-resolution~\citep{johnson2016perceptual,wang2021realesrgan},
and inpainting~\citep{suvorov2022lama,yu2019free}.
Unified architectures subsequently subsumed individual tasks under a single
model, progressing through multi-degradation routing~\citep{yu2018crafting,suganuma2019attention}
and latent-space translation~\citep{wan2020bringing} to large-capacity transformers
and state-space models~\citep{liang2021swinir,zamir2022restormer,guo2024mambair,guo2025mambairv2}
that handle arbitrary combinations of corruptions.

Video restoration introduces the additional requirement of temporal consistency:
a model must reconcile information across misaligned frames without introducing
flicker, ghosting, or motion discontinuities.
Task-specific methods address inpainting~\citep{xu2019flowguided,kim2019deepvideo},
denoising~\citep{claus2019videnn,tassano2020fastdvdnet},
deblurring~\citep{su2017deepvideo}, and
super-resolution~\citep{haris2019rbpn,chan2021basicvsr,youk2024fmanet} in
isolation, while unified frameworks - EDVR~\citep{wang2019edvr},
BasicVSR++~\citep{chan2022basicvsrplusplus}, VRT~\citep{liang2024vrt},
RVRT~\citep{liang2022rvrt}, and AverNet~\citep{zhao2024avernet} - tackle diverse
corruptions under a single architecture through optical-flow
estimation~\citep{teed2020raft,huang2022flowformer} and deformable
alignment~\citep{dai2017deformable,zhou2022revisiting}.

\subsection{Old film and photo restoration}

Classical signal-processing methods addressed individual artifact  classes
algorithmically~\citep{zhang2009scratches,stanco2003towards,bruni2004scratch,chang2005photo,giakoumis2006digital}.
DeepRemaster~\citep{iizuka2019deepremaster} pioneered end-to-end neural film
restoration, and~\citep{wan2020bringing,wan2022oldfilm} extended
neural restoration to both photographs and film via synthesis-driven training.
More recent architectures~\citep{lin2024restoring,mambaofr,liu2026restoration}
continue to advance the state of the art, and on the photographic side
~\citep{ivanova2023simulating} and DeOldify~\citep{salmona2022deoldify}
demonstrated the respective roles of degradation fidelity and colorization in perceptual recovery.

\section{Experimental and Implementation Details}
\label{sec:appendix-implementation}

\subsection{Implementation details}
We train all restoration models from scratch under identical conditions so
that any performance difference reflects the choice of synthesis pipeline
rather than training setup.
All models use clips of length 7 and $256{\times}256$ patches, trained for
20 epochs with mixed-precision (fp16) DDP on 2 NVIDIA A100 GPUs, following
the default configuration files of each architecture without additional
hyperparameter tuning.
Training and online degradation were run on the same node with 16 CPU cores
and 128GB RAM.
The only exception is RVRT~\citep{liang2022rvrt}, which has by far the longest
training schedule of the evaluated architectures: due to time and hardware
constraints it was trained for 4 days on $4{\times}$ A100 GPUs, compared with
9.7 days on $8{\times}$ A100 GPUs reported by its authors, i.e.\ with nearly
five times fewer GPU-hours. RVRT is therefore not fully trained, and its
results (Table~\ref{tab:results}) should not be read as a reflection of the
architecture's full capacity.

Both pipelines run fully online inside the DataLoader, but they differ in
\emph{when} the crop is extracted.
The prior pipeline of Bringing Old Films~\citep{wan2022oldfilm} applies degradation to the full
source clip first and then extracts the $256{\times}256$ training crop,
meaning all operators process a full-resolution frame at every step.
AbsoluteDegradation extracts the $256{\times}256$ crop from the REDS
source~\citep{nah2019reds} \emph{first} and degrades only that small patch,
reducing the per-step compute of the degradation pass significantly.

\subsection{Target preprocessing and colour space}
\label{sec:appendix-targets}

\paragraph{Ground-truth sharpening.}
Following Bringing Old Films~\citep{wan2022oldfilm}, the clean training targets
are sharpened with the soft-masked unsharp-masking procedure of
Real-ESRGAN~\citep{wang2021realesrgan} (weight $0.50$, radius $51$,
threshold $10$). The target processing is \emph{identical} for both
degradation pipelines compared in Table~\ref{tab:results}, so the
comparison isolates the synthetic degradation itself. In the released code the
operator is an explicit, configurable pipeline step
(\texttt{GroundTruthSharpenStep}), applied after photometric jitter and before
8-bit quantisation (Figure~\ref{fig:pipeline}).

\paragraph{Restoration only, no colourisation.}
For every trained model, both the network input and the training target are
grayscale frames stored in a 3-channel RGB representation; the learned task is
restoration only. Clean frames are converted to grayscale before entering the
pipeline. Inside Stage~2, the single-channel frame is temporarily expanded to a
3-channel grayscale representation so that it can be blended with the RGBA
texture overlays, and at the end of Stage~2 it is converted back to a single
channel, on which all subsequent stages operate.

\subsection{Experimental settings for reproducibility}
Training uses online degradation from clean REDS training clips, with degraded
samples generated on-the-fly. We train on the full REDS train split and
validate on our generated paired validation split; final testing is reported on
real archival datasets (our benchmark and SRWOV~\citep{mambaofr}), while synthetic data is used
as training proxy only. In our training codebase, runs use clip length $7$,
$256{\times}256$ crops, fp16 DDP, and fixed global seeds.
Optimization follows the released defaults: Adam for generator and
discriminator (\texttt{lr=2e-4}, $\beta_1=0.9$, $\beta_2=0.99$) with linear
decay over the last 10 of 20 epochs. BasicVSR++ uses the REDS4-style
restoration setting (with upsampling removed in our adaptation), and RVRT uses
its deblurring configuration.

\subsection{Statistical significance and repeated seeds}
\label{sec:appendix-seeds}
Repeating the full multi-model training and ablation suite with several seeds
would require a prohibitive additional compute budget: beyond long
training runs, test-time inference and frame export alone require substantial
resources (1$\times$ A100 GPU, 8 CPU cores, 32\,GB RAM), taking about 7--14\,h
on SRWOV and 12--20\,h on our benchmark depending on model, while evaluation on
the same node takes about 2\,h (SRWOV) and 4\,h (our benchmark).
We therefore repeat the configuration carrying the largest reported gain,
BasicVSR++ evaluated on our benchmark, with four seeds per pipeline
(Table~\ref{tab:seeds}). AbsoluteDegradation outperforms the baseline pipeline
on every metric for every seed, and the separation between the two pipelines is
several times larger than the seed-to-seed spread. Statistical significance of
the human preference study is reported in Section~\ref{sec:appendix-human}.

\begin{table}[ht]
\centering
\caption{Repeated-seed runs of BasicVSR++~\citep{chan2022basicvsrplusplus} on \textbf{our dataset}. ``Table~\ref{tab:results}'' denotes the run reported in the main text. Mean $\pm$ sample standard deviation over the four runs.}
\label{tab:seeds}
\scriptsize
\setlength{\tabcolsep}{5pt}
\begin{tabular}{@{}llcccc@{}}
\toprule
\textbf{Pipeline} & \textbf{Seed} & \textbf{BRISQUE}~$\downarrow$ & \textbf{CLIPIQA+}~$\uparrow$ & \textbf{MUSIQ}~$\uparrow$ & \textbf{MANIQA}~$\uparrow$ \\
\midrule
\multirow{5}{*}{Bringing Old Films~\citep{wan2022oldfilm}}
 & 42    & 30.86 & 0.303 & 32.36 & 0.163 \\
 & 123   & 27.58 & 0.319 & 35.18 & 0.178 \\
 & 9999  & 28.22 & 0.318 & 33.05 & 0.177 \\
 & Table~\ref{tab:results} & 30.62 & 0.313 & 32.55 & 0.170 \\
 & mean $\pm$ std & $29.32 \pm 1.66$ & $0.313 \pm 0.007$ & $33.29 \pm 1.30$ & $0.172 \pm 0.007$ \\
\midrule
\multirow{5}{*}{AbsoluteDegradation (ours)}
 & 42    & 22.71 & 0.363 & 41.58 & 0.250 \\
 & 123   & 22.73 & 0.359 & 42.49 & 0.247 \\
 & 9999  & 20.47 & 0.364 & 42.74 & 0.253 \\
 & Table~\ref{tab:results} & 20.68 & 0.369 & 42.91 & 0.238 \\
 & mean $\pm$ std & $\mathbf{21.65 \pm 1.24}$ & $\mathbf{0.364 \pm 0.004}$ & $\mathbf{42.43 \pm 0.59}$ & $\mathbf{0.247 \pm 0.006}$ \\
\bottomrule
\end{tabular}
\end{table}

\subsection{Compute notes}
Both restoration and online degradation run on the same node (2$\times$ NVIDIA
A100 GPUs, 16 CPU cores, 128\,GB RAM). For RTN, training takes
\textbf{1\,d\,13\,h\,46\,m} with our crop-first degradation order versus
\textbf{3\,d\,14\,h\,19\,m} with the prior full-frame-first order, indicating a
substantial avoidable compute and cost overhead in the latter due to degrading
full-resolution frames before crop extraction. RVRT was trained on
$4{\times}$ A100 GPUs for 4 days. In practice, debugging and verification
required additional runs beyond those reported in the paper.

\subsection{Test and evaluation details}
\label{sec:appendix-evaluation}
To ensure rigorous assessment of the video restoration performance, we employ a sliding temporal window strategy for all video evaluations. This approach allows the models to process long video clips while maintaining constant memory overhead.

We distinguish between two primary evaluation regimes based on the dataset. For SRWOV dataset~\citep{mambaofr} we utilize a temporal chunk size of $11$ frames with a stride of $3$ frames. For our own benchmark, we use smaller temporal window, using a chunk size of $7$ frames, due to "out of memory" GPU bottleneck, while maintaining the same stride of $3$ frames.

To avoid duplicate saves and biased metrics from overlapping segments, we implement a strict frame commitment policy. We save the entire first temporal window. For all subsequent windows, we retain and evaluate only the new frames introduced by the stride.

\paragraph{Per-frame metrics.}
CLIPIQA+, MUSIQ, MANIQA, and BRISQUE are aggregated over the entire test set by taking the arithmetic mean over all restored frames. Each frame contributes equally to the final score, regardless of which clip it belongs to. This matches the standard convention used in evaluation settings for image quality metrics and benchmark scoring.
\begin{equation}
    M_{final} = \frac{1}{T_{total}}\sum_{i=1}^N\sum_{t=1}^{T_i} M\left( \hat{x}_i^t \right)
\end{equation}
where $M_{final}$ is the final aggregated metric, $N$ is the total number of clips, $T_i$ the number of frames in $i$-th clip, $ \hat{x}_i^t$ denotes restored $t$-th frame from $i$-th clip, $M(\cdot)$ is a specific evaluation function applied to a single restored frame, $T_{total}$ denotes total number of frames (i.e. $T_{total}=\sum_{i=1}^{N}T_{i}$).

\paragraph{Clip-level and temporal metrics.}
Per-frame averages cannot detect flicker or inconsistent artifact removal
across frames. We therefore additionally evaluate each restored
\emph{sequence}. DOVER~\citep{wu2023dover} is a clip-level video quality
assessor with an aesthetic and a technical branch; the technical branch is the
one sensitive to temporal distortions such as flicker and jitter, and the
overall score fuses both. We further report two inter-frame stability
measures: tLPIPS~\citep{chu2020tecogan}, the LPIPS~\citep{zhang2018lpips}
distance between consecutive output frames, and CLIP-F, the cosine similarity
between CLIP~\citep{radford2021clip} embeddings of consecutive frames, computed
with the \texttt{ViT-B-32-quickgelu} model (OpenAI weights) through the
CLIP wrapper of MMagic~\citep{mmagic2023}.

\subsection{FID computation details}
\label{sec:appendix-fid}

We compute the Fréchet Inception Distance (FID) \citep{heusel2017gans} using a patched Inception-v3\citep{Szegedy_2016_CVPR} network, utilizing pre-trained ImageNet (ver. from 2015-12-05) \citep{Seitzer2020FID}. Our implementation relies on the standard \emph{mseitzer/pytorch-fid} library \citep{Seitzer2020FID}. We pre-process frames by:
(i) bit-depth normalization (scaling [0-255] integers to [0-1] floating numbers),
(ii) bilinear resizing of frame to 299x299,
(iii) scaling to [-1, 1].
We opted for the standard ImageNet-trained Inception-v3 over a domain-specific extractor to maintain alignment with industry standards.

\section{AbsoluteDegradation Pipeline Details}
\label{sec:appendix-pipeline}

\begin{table}[ht]
\centering
\caption{Complete degradation configuration (config-first, code-default fallback) for the active Light/Medium/Heavy pipelines. Single values indicate shared parameter across all three severity tiers. Blur uses $k=2r+3$; JPEG uses $Q=\operatorname{clip}(Q_c+\delta,40,100)$ with $\delta\sim\mathcal{U}\{-15,\dots,15\}$ sampled per frame. Noise types (per clip, $p=0.25$ each): heteroscedastic SDN grain, homoscedastic grain, additive Gaussian, multiplicative speckle.}
\label{tab:severity}
\tiny
\setlength{\tabcolsep}{3pt}
\begin{tabular}{@{}lllccc@{}}
\toprule
\textbf{Stage} & \textbf{Degradation} & \textbf{Parameter} &
  \textsc{Light} & \textsc{Medium} & \textsc{Heavy} \\
\midrule
Global & & Severity probabilities & $0.30$ & $0.35$ & $0.35$ \\
\midrule
Stage 1 & Random Crop & Height & \multicolumn{3}{c}{$286$} \\
Stage 1 & Random Crop & Width & \multicolumn{3}{c}{$266$} \\
\midrule
Stage 2 & Textures & Textures per frame range & \multicolumn{3}{c}{$[1, 3]$} \\
Stage 2 & Textures & Moving line probability & \multicolumn{3}{c}{$0.3$} \\
Stage 2 & Textures & Texture durations with probabilities & \multicolumn{3}{c}{$\{1,3,5\}$ with $p=(0.95, 0.04, 0.01)$} \\
\midrule
Stage 3 & & Permutation probability & \multicolumn{3}{c}{$0.5$} \\
Stage 3 & Blur & Kernel size range ($r$) & \multicolumn{3}{c}{$[0,2]$} \\
Stage 3 & Blur & Weight & \multicolumn{3}{c}{$0.5$} \\
Stage 3 & Noise & Grain ISO range ($k$)  & \multicolumn{3}{c}{$[0.01,0.05]$} \\
Stage 3 & Noise & Grain size range & \multicolumn{3}{c}{$[1.0,1.3]$} \\
Stage 3 & Noise & signal-dependent amplification range ($\gamma$) & \multicolumn{3}{c}{$[0.3,0.7]$} \\
Stage 3 & Noise & White noise std. range ($\sigma_w$) & \multicolumn{3}{c}{$[0.001,0.005]$} \\
Stage 3 & Noise & Gaussian noise std. range & \multicolumn{3}{c}{$[5,6]/255$} \\
Stage 3 & Noise & Noise type probabilities & \multicolumn{3}{c}{$0.25$ each (see caption)} \\
Stage 3 & Resize & Rescale bounds & \multicolumn{3}{c}{$[0.9,1.1]$} \\
Stage 3 & Gamma corr. & Gamma distribution ($\gamma$) & \multicolumn{3}{c}{$\text{clip}(\exp(\mathcal{N}(0.0, 1/12), 0.2, 4))$} \\
\midrule
Stage 4 & Noise & Grain ISO range ($k$) & $[0.02,0.075]$ & $[0.03,0.08]$ & $[0.04,0.10]$ \\
Stage 4 & Noise & Grain size range & $[1.0,1.3]$ & $[1.0,1.3]$ & $[1.0,1.4]$ \\
Stage 4 & Noise & signal-dependent amplification range ($\gamma$) & \multicolumn{3}{c}{$[0.3,0.7]$} \\
Stage 4 & Noise & White noise std. range ($\sigma_w$) & $[0.005,0.01]$ & $[0.005,0.015]$ & $[0.01,0.02]$ \\
Stage 4 & Noise & Gaussian noise std. range & $[5,6]/255$ & $[6,8]/255$ & $[8,10]/255$ \\
Stage 4 & Noise & Noise type probabilities & \multicolumn{3}{c}{$0.25$ each (see caption)} \\
Stage 4 & Camera Gate Weave & $X$-motion padding ($p_x$) & \multicolumn{3}{c}{$5$} \\
Stage 4 & Camera Gate Weave & $X$-motion pace ($\theta^{(x)}$) & \multicolumn{3}{c}{$1$} \\
Stage 4 & Camera Gate Weave & $X$-motion temperature range ($\sigma^{(x)}$) & $[0,0.7]$ & $[0,1.0]$ & $[0,1.5]$ \\
Stage 4 & Camera Gate Weave & $Y$-motion padding ($p_y$) & \multicolumn{3}{c}{$15$} \\
Stage 4 & Camera Gate Weave & $\theta$-motion average value ($\theta_{\text{avg}}$) & \multicolumn{3}{c}{$0.25$} \\
Stage 4 & Camera Gate Weave & $\theta$-motion pace ($\kappa$) & \multicolumn{3}{c}{$0.02$} \\
Stage 4 & Camera Gate Weave & $\theta$-motion temperature ($s$) & \multicolumn{3}{c}{$1$} \\
Stage 4 & Camera Gate Weave & $Y$-motion temperature range ($\sigma^{(y)}$) & $[0, 12]$ & $[0, 16]$ & $[0, 20]$ \\
Stage 4 & Blur & Kernel size range ($r$) & $[1,3]$ & $[1,5]$ & $[1,7]$ \\
Stage 4 & Blur & Weight & \multicolumn{3}{c}{$2.0$} \\
Stage 4 & JPEG compression & Average quality range & $[70,100]$ & $[50,80]$ & $[40,70]$ \\
Stage 4 & JPEG compression & Quality std. & \multicolumn{3}{c}{$15$} \\
Stage 4 & Resize & Downscale factor range & $[0.5, 0.9]$ & $[0.4, 0.9]$ & $[0.3, 0.8]$ \\
Stage 4 & Resize & Upscale factor range & $[1.0, 1.5]$ & $[1.0, 2.0]$ & $[1.0, 2.0]$ \\
Stage 4 & Gamma corr. & Gamma distribution ($\gamma$) & \multicolumn{3}{c}{$\text{clip}(\exp(\mathcal{N}(0.0, 0.125), 0.2, 4))$} \\
\midrule
Stage 5 & Color jitter & Probability & \multicolumn{3}{c}{$0.5$} \\
Stage 5 & Color jitter & Brightness range & \multicolumn{3}{c}{$[0.8,1.2]$} \\
Stage 5 & Color jitter & Contrast range & \multicolumn{3}{c}{$[0.9,1.0]$} \\
Stage 5 & GT sharpening & Unsharp mask weight / radius / threshold & \multicolumn{3}{c}{$0.50$ / $51$ / $10$} \\
\midrule
Stage 6 & Mechanical scratches & Clip probability & \multicolumn{3}{c}{$0.3$} \\
Stage 6 & Mechanical scratches & FPS & \multicolumn{3}{c}{24} \\
Stage 6 & Mechanical scratches & Count range & \multicolumn{3}{c}{$[1,5]$} \\
Stage 6 & Mechanical scratches & Spawn prob. & \multicolumn{3}{c}{$0.3$} \\
Stage 6 & Mechanical scratches & Microns width range & \multicolumn{3}{c}{$[5,40]$} \\
Stage 6 & Mechanical scratches & Scan resolution DPI & \multicolumn{3}{c}{$2000$} \\
Stage 6 & Mechanical scratches & Intensity range & \multicolumn{3}{c}{$[0.2,1.0]$} \\
Stage 6 & Mechanical scratches & Weave freq. [Hz] & \multicolumn{3}{c}{$2.0$} \\
Stage 6 & Mechanical scratches & Weave amplitude [pixels] & \multicolumn{3}{c}{$2.0$} \\
Stage 6 & Mechanical scratches & Flicker freq. [Hz] & \multicolumn{3}{c}{$12$} \\
Stage 6 & Mechanical scratches & Curvature prob. & \multicolumn{3}{c}{$0.2$} \\
Stage 6 & Mechanical scratches & Lifespan range & \multicolumn{3}{c}{$[2,9]$} \\
Stage 6 & Mechanical scratches & White color prob. & \multicolumn{3}{c}{$0.6$} \\
\bottomrule
\end{tabular}
\end{table}

\paragraph{Stage 1.}
Each source frame is cropped to a padded working patch of $(H{+}2p_y){\times}(W{+}2p_x)$\,px ($286{\times}266$ by default). The asymmetric margins $p_y{=}15$, $p_x{=}5$ are essential: the vertical and horizontal gate-weave operators in Stage~4 each trim their axis by $2p$, restoring the output to exactly $H{\times}W{=}256{\times}256$. A crop origin is sampled once per clip and fixed across all $T$ frames, so the degraded sequence and ground truth are aligned to the same spatial location. The asymmetry $p_y>p_x$ reflects the physical dominance of vertical register drift over lateral sprocket play, consistent with the wider $\sigma_y$ bounds in Table~\ref{tab:severity}.

\paragraph{Stage 2.}
At each frame, \(K_t \sim \mathrm{Unif}\{1, 2, 3\}\) slots are activated; lifetimes
are drawn from
$\operatorname{Categorical}(\{1,3,5\},\,[0.95,\,0.04,\,0.01])$.
Each slot selects a blend mode uniformly from $\{\text{add},\,\text{sub},\,\text{mul}\}$
(modelling overexposure, dust occlusion, and staining respectively) and is
composited at opacity $\omega\!\sim\!\mathcal{U}(0.8,1.0)$.
A per-clip spatial offset $(u_0,v_0)$ locks texture placement across frames.
Prior to compositing, each texture undergoes random rotation with
interior-rectangle cropping, optional morphological erosion, and contrast
enhancement up to $4\times$ (probability $0.7$), simulating the broad variation
of real-world film contamination.
Additionally, 30\,\% of clips activate a \emph{moving-line} element with
Brownian horizontal drift of $5$--$15$\,px per frame.
Stage~2 concludes with a grayscale conversion; all subsequent stages operate
on single-channel images (Section~\ref{sec:appendix-targets}).

\paragraph{Stage 5.}
After photometric jitter, the clean target is sharpened with the unsharp mask
described in Section~\ref{sec:appendix-targets}, identically for both
pipelines; the degraded frame and the target are then quantised to 8 bits.

\begin{figure*}[ht]
  \centering
  \footnotesize
  \def\tiercmp{0.22\textwidth}
  \setlength{\tabcolsep}{6pt}

  \begin{tabular}{@{} c c @{}}
    \begin{tabular}{@{} c @{}}
       \includegraphics[width=\tiercmp]{figures/tier_severity_visual/grid_frame000_gt} \\
       \addlinespace[4pt]
       \textbf{GT}
    \end{tabular}
    &
    \begin{tabular}{@{} c c c c @{}}
      \toprule
      & \textsc{Light} & \textsc{Medium} & \textsc{Heavy} \\
      \cmidrule(l){2-4}
      \rotatebox[origin=c]{90}{\textbf{Ours}} &
      \raisebox{-0.5\height}{\includegraphics[width=\tiercmp]{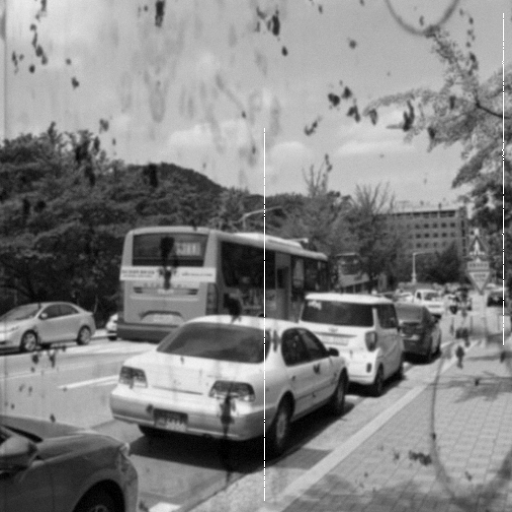}} &
      \raisebox{-0.5\height}{\includegraphics[width=\tiercmp]{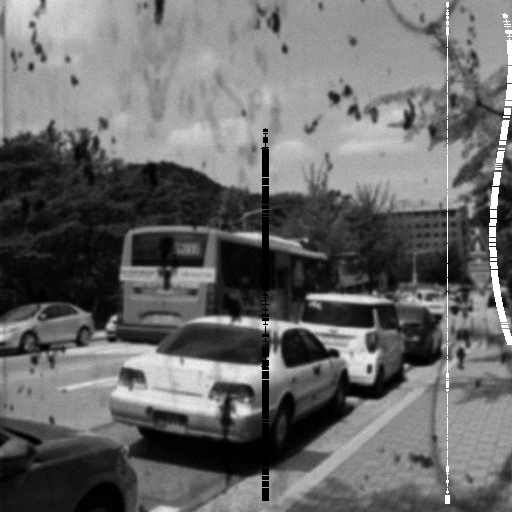}} &
      \raisebox{-0.5\height}{\includegraphics[width=\tiercmp]{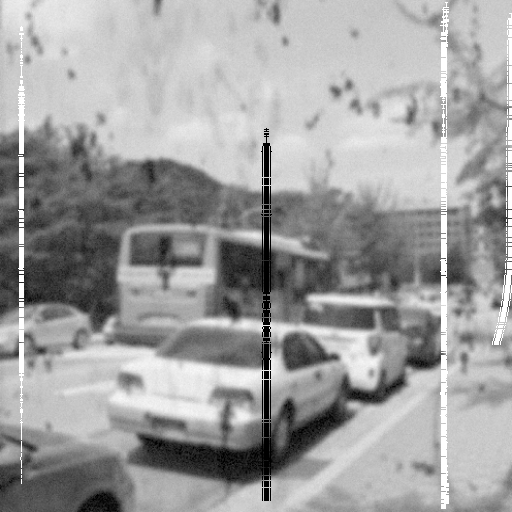}} \\
      \addlinespace[4pt]
      \rotatebox[origin=c]{90}{\textbf{Bringing Old Films~\citep{wan2022oldfilm}}} &
      \raisebox{-0.5\height}{\includegraphics[width=\tiercmp]{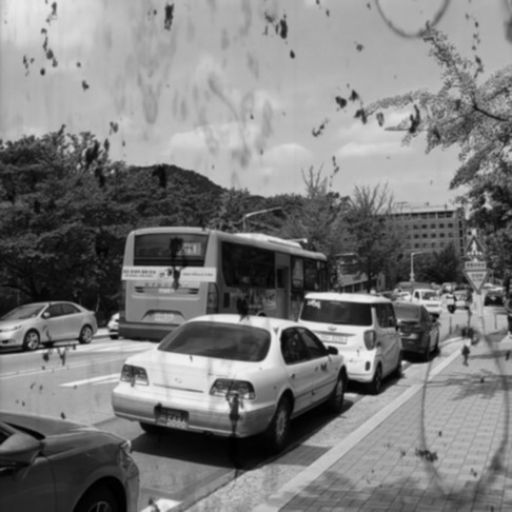}} &
      \raisebox{-0.5\height}{\includegraphics[width=\tiercmp]{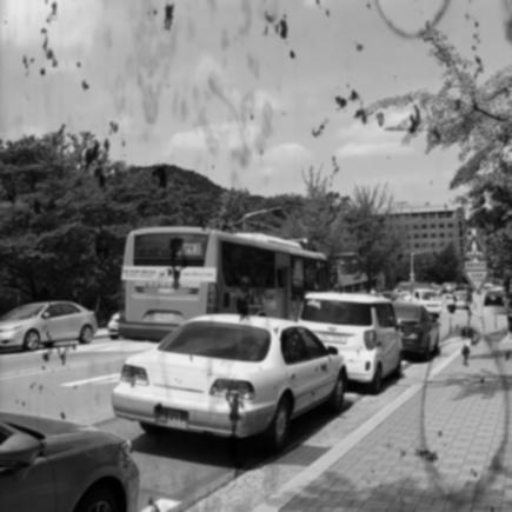}} &
      \raisebox{-0.5\height}{\includegraphics[width=\tiercmp]{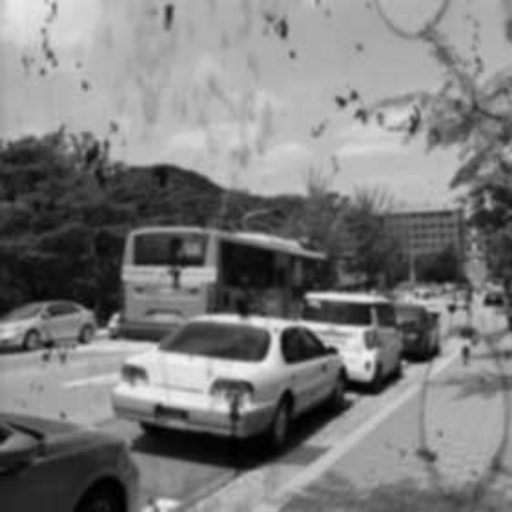}} \\
      \bottomrule
    \end{tabular}
  \end{tabular}

  \caption{Tier comparison on REDS frame 000 (single fixed crop): ground truth (left; shown in colour for reference only, as all training inputs and targets are grayscale, see Section~\ref{sec:appendix-targets}) and tier-conditioned outputs for our pipeline (top row) and Bringing Old Films Back to Life~\citep{wan2022oldfilm} (bottom row).}
  \label{fig:tier_comparison_fixed_frame}
\end{figure*}

\begin{figure*}[ht]
  \centering
  \footnotesize
  \setlength{\tabcolsep}{3pt}
  \renewcommand{\arraystretch}{1.0}
  \def\DistColW{0.235\linewidth}
  \def\DistCap#1{\scriptsize #1\par}
  \noindent
  \begin{tabular}{@{}c@{\hspace{3pt}}c@{\hspace{3pt}}c@{\hspace{3pt}}c@{}}
    \begin{minipage}[t]{\DistColW}\centering
      \includegraphics[width=\linewidth]{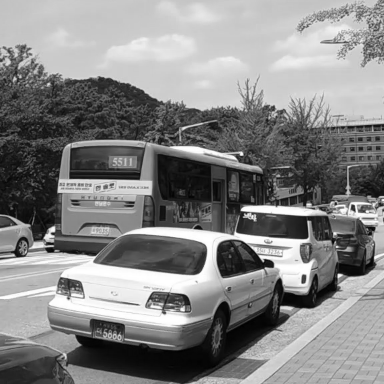}\\[2pt]
      \DistCap{\textbf{Ground truth}}
    \end{minipage} &
    \begin{minipage}[t]{\DistColW}\centering
      \includegraphics[width=\linewidth]{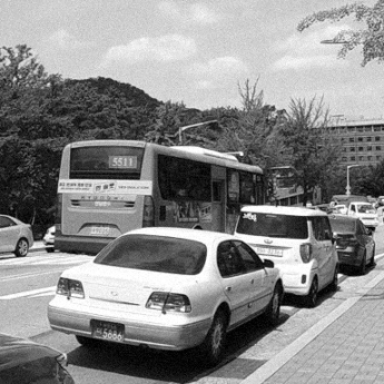}\\[2pt]
      \DistCap{\textbf{Negative-film chain}}
    \end{minipage} &
    \begin{minipage}[t]{\DistColW}\centering
      \includegraphics[width=\linewidth]{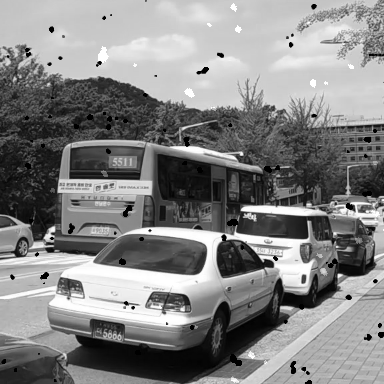}\\[2pt]
      \DistCap{\textbf{Texture overlays}}
    \end{minipage} &
    \begin{minipage}[t]{\DistColW}\centering
      \includegraphics[width=\linewidth]{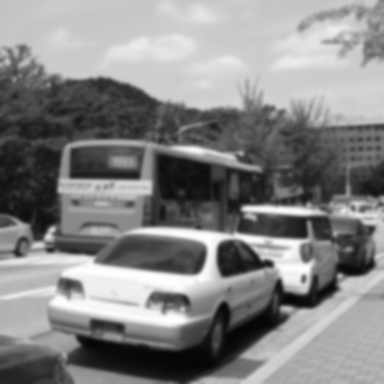}\\[2pt]
      \DistCap{\textbf{Optical blur}}
    \end{minipage} \\[15pt]
    \begin{minipage}[t]{\DistColW}\centering
      \includegraphics[width=\linewidth]{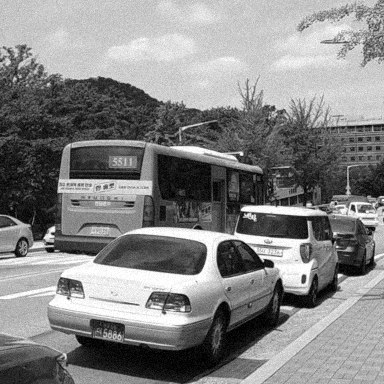}\\[2pt]
      \DistCap{\textbf{Film grain}}
    \end{minipage} &
    \begin{minipage}[t]{\DistColW}\centering
      \includegraphics[width=\linewidth]{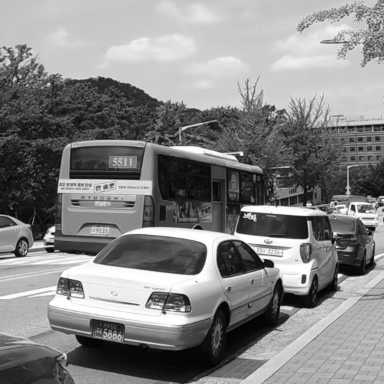}\\[2pt]
      \DistCap{\textbf{JPEG}}
    \end{minipage} &
    \begin{minipage}[t]{\DistColW}\centering
      \includegraphics[width=\linewidth]{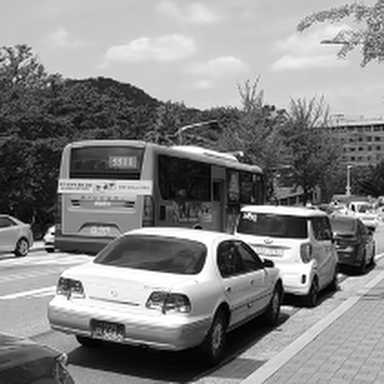}\\[2pt]
      \DistCap{\textbf{Resampling}}
    \end{minipage} &
    \begin{minipage}[t]{\DistColW}\centering
      \includegraphics[width=\linewidth]{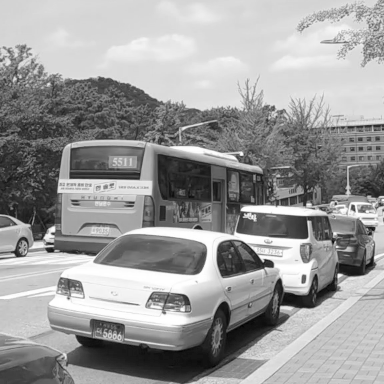}\\[2pt]
      \DistCap{\textbf{Photometric ($\gamma$)}}
    \end{minipage} \\[15pt]
    \begin{minipage}[t]{\DistColW}\centering
      \includegraphics[width=\linewidth]{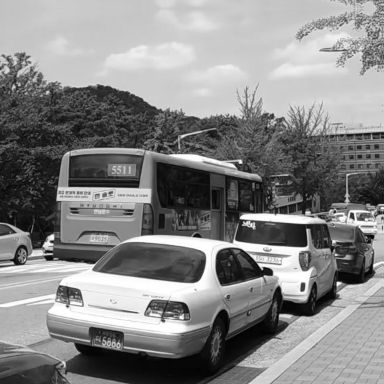}\\[2pt]
      \DistCap{\textbf{Gate weave}}
    \end{minipage} &
    \begin{minipage}[t]{\DistColW}\centering
      \includegraphics[width=\linewidth]{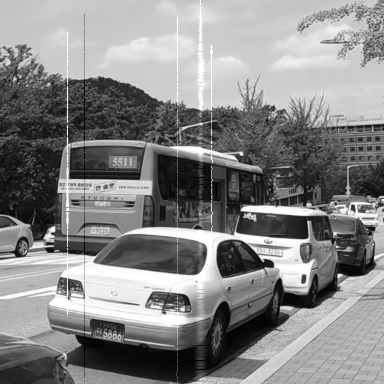}\\[2pt]
      \DistCap{\textbf{Scratches}}
    \end{minipage} &
    \begin{minipage}[t]{\DistColW}\centering
      \includegraphics[width=\linewidth]{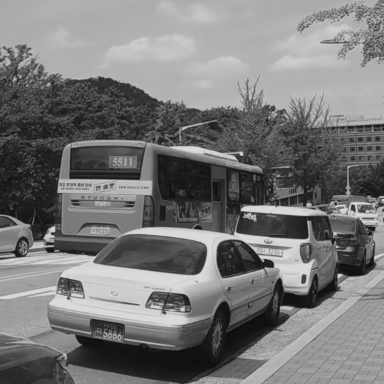}\\[2pt]
      \DistCap{\textbf{Color jitter}}
    \end{minipage} &
    \begin{minipage}[t]{\DistColW}\centering
      \includegraphics[width=\linewidth]{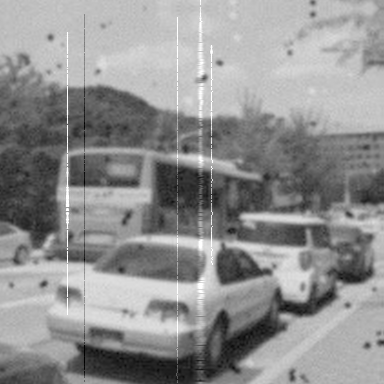}\\[2pt]
      \DistCap{\textbf{All degradations}}
    \end{minipage}
  \end{tabular}
  \caption{%
    \textbf{AbsoluteDegradation operator atlas}.
    Reference crop (top-left), ten single-operator panels, and a chained \emph{all degradations} preview (bottom-right)}
  \label{fig:distortion_operator_atlas}
\end{figure*}

\section{Old Video Dataset Details}
\label{sec:appendix-old-video-dataset}

Evaluating restoration models on real archival footage requires a benchmark
that faithfully represents the target domain.
As discussed in Section~\ref{sec:related}, the real-world partition of the
SRWOV benchmark~\citep{mambaofr} is the only publicly available resource of
this kind, yet its low resolution, lossy JPEG storage, heterogeneous
content (including cartoons), the frequent presence of overlay watermarks not
found on original archival scans, and the relative scarcity of physical analog
degradations across the footage, limit its value as a domain-representative
reference.
To address these shortcomings, we construct three benchmark
datasets from 30 public-domain archival videos (1896--1918) downloaded from the
Library of Congress.
All source footage is stored as lossless PNG at resolutions ranging from
$1440{\times}1080$ to $1920{\times}1080$, preserving the fidelity of the
original scans and ensuring that downstream compression from sharing or
further processing does not confound the analog degradations under study.

\subsection{Data Acquisition and Processing}
\label{sec:appendix-old-video-dataset:acquisition}

\paragraph{Scene splitting.}
Raw archival videos typically interleave heterogeneous content, narrative
footage, title cards, intertitles, and blank leaders, within a single file.
To isolate coherent visual segments we employ a feature-matching-based scene
detector built on SuperPoint~\citep{DeTone_2018_CVPR_Workshops} and
LightGlue~\citep{Lindenberger_2023_ICCV}.
For every pair of sampled frames (one frame per ten source frames), SuperPoint
extracts up to 128 keypoints and LightGlue matches them; a scene boundary is
declared when the match count drops below an experimentally determined
threshold of 25 matching keypoints.
Detected clips shorter than 100 frames are discarded, and 0.1\,s is trimmed
from each clip boundary to remove transition artefacts.

\paragraph{Manual curation.}
All clips produced by the automatic splitter were reviewed by a human
researcher.
Clips consisting of title cards, intertitles, or blank frames were
removed.
Subtle within-clip scene changes missed by the automatic detector were
identified and the affected clips were further subdivided.
Leading and trailing blank or non-informative frames were trimmed from each
clip respecting the lower limit of 100 frames.

\paragraph{Post-processing.}
After curation, all clips were organized into a consistent hierarchical structure
and frames were compressed as lossless PNG, reducing the on-disk size by over
50\% compared to the raw source while preserving bit-exact pixel values.

\subsection{Introduced Datasets}
\label{sec:appendix-old-video-dataset:datasets}

From the curated footage we derive three benchmark datasets, each targeting a
different evaluation or training scenario.

\subsubsection{Full scene-split dataset.}
The primary dataset contains every curated single-scene clip, totalling
\textbf{81,576 frames}.
Each clip depicts a single continuous shot with no internal cuts, making it
directly suitable for evaluating temporal restoration methods that assume
scene-level coherence.

\subsubsection{Compact scene-split dataset subset.}
A smaller evaluation dataset is constructed by selecting up to five clips per
source video (shorter videos may contain fewer scenes) and retaining up to
150 central frames of each clip (clips as short as 100 frames are included).
This yields \textbf{13,252 frames}, a lightweight benchmark that preserves
the diversity of the full dataset while being practical for large models evaluation and ablation studies.

\subsubsection{Multi-scene fixed-length dataset.}
To support extensive evaluation of models that operate on full, unedited archival
footage, including scene transitions and content discontinuities, the
source videos are divided into fixed-length clips of 150 frames
\emph{without} prior scene splitting.
After manual removal of blank, title-card, and non-informative segments, this
dataset contains \textbf{121,200 frames}.
It serves as a benchmark for methods capable of handling scene changes
without explicit prior preprocessing.

These datasets provide a large pool of unlabeled
archival frames that can be used as a solid foundation for research on unsupervised restoration approaches.

\subsection{Scope and generalisation}
\label{sec:appendix-scope}
The benchmark targets early black-and-white analog film (1896--1918), the
dominant part of the archival holdings of our partner institutions, and the
task is grayscale restoration only. The released models, synthesis
configuration, and benchmark are scoped to this domain; we make no claim about
colour film, VHS, or other legacy media. Within this scope, gains hold on the
more heterogeneous SRWOV benchmark, and the modular pipeline can be
re-parameterised for other eras or media.

\section{Human Preference Study}
\label{sec:appendix-human}

\begin{figure}[ht]
  \centering
  \includegraphics[width=0.6\linewidth]{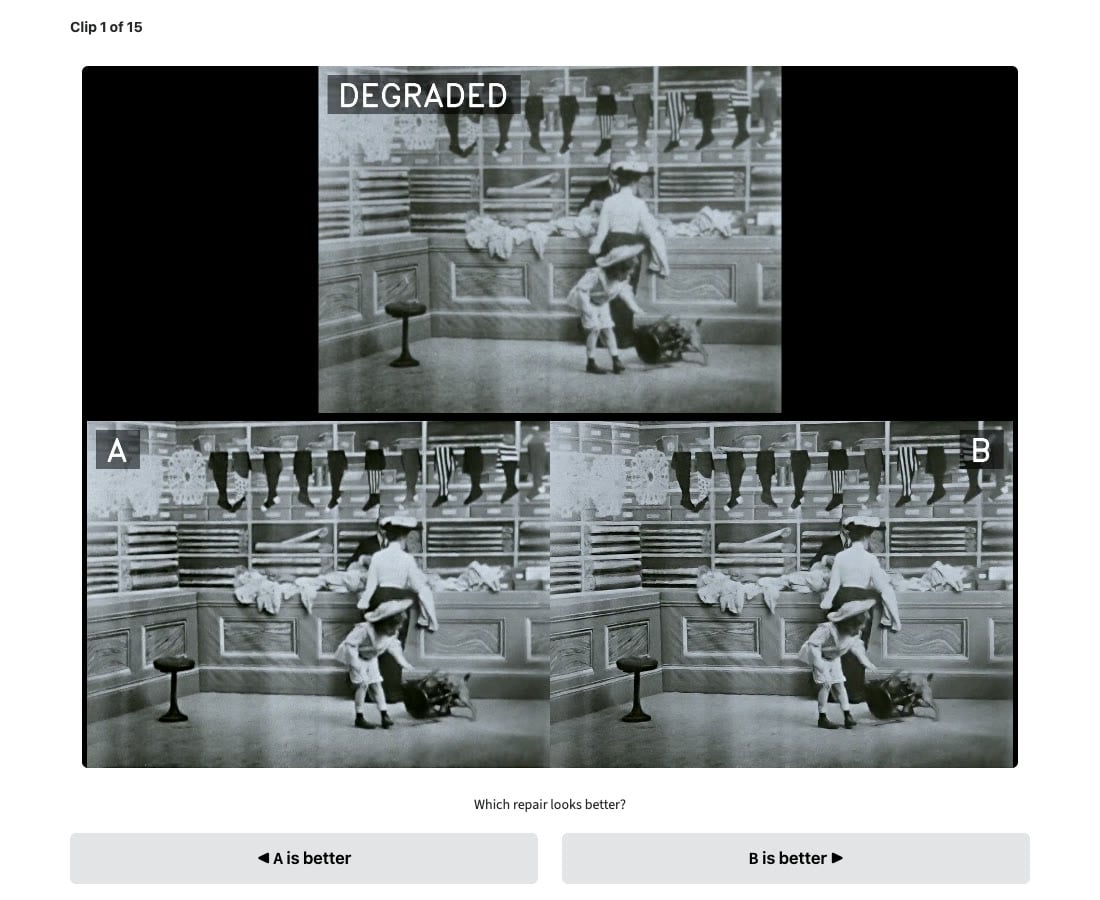}
  \caption{Interface of the human preference study: the degraded clip is shown on top as a reference, and the two restorations (Video~A, Video~B) below; all three play in sync in a looping player.}
  \label{fig:study_interface}
\end{figure}

\paragraph{Protocol.}
We use a blinded two-alternative forced-choice (2AFC) design rather than an
absolute mean opinion score: archival restoration has no clean reference to
anchor an absolute scale, so absolute ratings would largely reflect each rater's
private prior about how old film ought to look. For each of the six
architectures and 29 clips from our archival benchmark (174 pairs), raters saw
the degraded input on top as a reference and, below it, the restorations of the
same architecture trained with the Bringing Old Films pipeline and with
AbsoluteDegradation, side by side as Video~A and Video~B
(Figure~\ref{fig:study_interface}). All three videos play in sync, in a single
looping player, so flicker and inconsistent artifact removal are directly
visible. The side of each method was assigned by a seeded per-pair coin flip,
filenames were opaque, and no method was identifiable from the page. The choice
buttons unlock only after one full loop (7\,s), and there is no tie option.
Each rater completed a block of 15 clips covering all six architectures (2 or 3
clips per architecture, with the 3-clip slots rotated across raters), giving
near-equal per-architecture totals of 228--237 trials. Incomplete sessions were
excluded.

\paragraph{Participants.}
Raters were recruited through two independent channels: the Prolific platform
(42 raters) and direct invitation (51 raters), yielding 93 completed surveys
and 1,395 trials. Prolific raters were paid on average \pounds 11.60 per hour
(median completion time 6\,min 31\,s), above the platform's recommended rate.
The study involved only viewing short public-domain archival clips and
choosing a preferred restoration; no formal institutional ethics review was
obtained.

\paragraph{Statistics.}
The win rate is the number of trials in which the AbsoluteDegradation
restoration was chosen divided by the number of trials. Two-sided exact binomial
tests are taken against chance ($p=0.5$) and 95\% confidence intervals use
Wilson's formula~\citep{wilson1927}. The per-trial tests treat trials as
independent and are not corrected for clustering by rater or by clip.
Results are reported in Table~\ref{tab:human} in the main text. The two
recruitment channels agree closely (63.7\% vs.\ 65.6\% overall).
With roughly 235 trials per architecture, the design
resolves win rates above approximately 57--58\%; MambaOFR and RTN are therefore
undetermined at this sample size, consistent with anything from parity to a
modest benefit. All six architectures nevertheless favour AbsoluteDegradation
(two-sided sign test across architectures, $p=0.031$).

\paragraph{Instructions shown to participants.}
The following text is shown verbatim on the start screen of the study.

\begin{quote}\footnotesize
\textbf{Intro.}
Video restoration is the process of improving the quality of degraded, old, or
low-resolution videos by removing artifacts like noise, blur, and compression errors.
While artificial intelligence and modern algorithms can enhance these videos in many
different ways, the ultimate test of a restored video is how it looks to the human eye.
We are conducting this research to understand viewer preferences. Your feedback will
help us determine which restoration techniques produce the most natural, visually
pleasing, and high-quality results.

\textbf{Survey Instructions.}
Please review these quick guidelines to ensure your setup is ready and you know exactly
what to look for. You will rate 15 clips; it takes about 10 minutes.

\textbf{1. Optimize Your Setup.}
\emph{Screen Size:} Use a laptop, desktop, or tablet. Avoid smartphones, as the screen is
too small to evaluate fine details.
\emph{Brightness:} Turn your screen brightness up to at least 80\%.
\emph{Connection:} Give the video a moment to fully load in high resolution before you
start watching.

\textbf{2. How to Evaluate (2AFC Format).}
This survey uses a Two-Alternative Forced Choice (2AFC) format. For every question,
you will see the original, degraded video at the top of the screen as a reference,
along with a pair of restored videos (Video A on the bottom left and Video B on
the bottom right) to compare. All three play together, in sync, in a single player.
\emph{Use the Reference:} Look at the top video to understand the original quality and
flaws before evaluating the two restorations.
\emph{Watch completely:} Let the clip play all the way through at least once. It loops
automatically, so you can rewatch it as many times as you need --- the choice buttons
unlock after a few seconds.
\emph{Forced Choice:} You must select which restored video looks better overall. There is
no ``tie,'' ``neutral,'' or ``both'' option. Even if the differences are extremely subtle,
trust your gut and pick the one that looks slightly better.

\textbf{3. What to Look For.}
When deciding between Video A and Video B, pay close attention to these factors:
\emph{Sharpness \& Detail:} Are the edges crisp? Can you see fine textures (like hair,
grass, or brick) clearly?
\emph{Visual Artifacts:} Does one video have weird visual glitches, blocky patches, or
artificial shapes that shouldn't be there?
\emph{Temporal Smoothness:} Does the motion look natural across the frames, or does it
flicker and warp unnaturally?
\emph{Overall Realism:} Which version looks more like a natural, high-quality video rather
than something artificially processed?

\textbf{4. Making Your Selection.}
There are no trick questions and no ``right'' or ``wrong'' answers. We simply want your
subjective opinion as a viewer.
Choose either Video A or Video B, then proceed to the next pair.
\end{quote}

On each trial, a progress header (``Clip 1 of 15'') is shown; the prompt reads
``Watch the clip --- the buttons unlock in a moment.'' until the buttons unlock,
then changes to ``Which repair looks better?'', with the buttons
``A is better'' and ``B is better''. After the last clip, Prolific participants
are shown a link confirming completion.

\clearpage
\section{Additional Results and Ablations}
\label{sec:appendix-results}

\subsection{Clip-level and temporal metrics}
\label{sec:appendix-temporal}

Table~\ref{tab:dover} reports DOVER~\citep{wu2023dover} scores of the restored
sequences for all six architectures on three test sets. Across the 18
model--dataset comparisons, AbsoluteDegradation wins 16 on DOVER-technical (the
branch sensitive to temporal distortion and flicker), including all six
architectures on our archival benchmark, and 14 on both DOVER-overall and
DOVER-aesthetic. The gains on DOVER-technical on our benchmark are largest for
Shift-Net and BasicVSR++, which also show the strongest human preference
(Section~\ref{sec:appendix-human}).

\begin{table}[ht]
\centering
\caption{Clip-level DOVER~\citep{wu2023dover} scores ($\uparrow$), reported as AbsoluteDegradation / Bringing Old Films~\citep{wan2022oldfilm}. The better value of each pair is bolded.}
\label{tab:dover}
\scriptsize
\setlength{\tabcolsep}{3pt}
\begin{tabular}{@{}lccc|ccc|ccc@{}}
\toprule
& \multicolumn{3}{c|}{\textbf{Synthetic paired set}} & \multicolumn{3}{c|}{\textbf{Our dataset}} & \multicolumn{3}{c}{\textbf{SRWOV}~\citep{mambaofr}} \\
\textbf{Model} & aes & tech & overall & aes & tech & overall & aes & tech & overall \\
\midrule
BasicVSR++   & \textbf{.8673}/.7971 & \textbf{.0464}/.0462 & \textbf{.2679}/.2310 & \textbf{.4898}/.4296 & \textbf{.0351}/.0246 & \textbf{.1251}/.0920 & \textbf{.7727}/.6663 & \textbf{.0381}/.0322 & \textbf{.2081}/.1576 \\
Shift-Net    & \textbf{.8832}/.8474 & \textbf{.0511}/.0490 & \textbf{.2910}/.2672 & \textbf{.5713}/.4980 & \textbf{.0416}/.0304 & \textbf{.1545}/.1167 & \textbf{.7678}/.7161 & \textbf{.0388}/.0339 & \textbf{.2079}/.1760 \\
MambaOFR     & \textbf{.8230}/.7910 & \textbf{.0462}/.0458 & \textbf{.2412}/.2293 & \textbf{.6187}/.5937 & \textbf{.0428}/.0401 & \textbf{.1681}/.1553 & \textbf{.7683}/.7500 & .0358/\textbf{.0375} & \textbf{.1998}/.1950 \\
RTN          & \textbf{.8280}/.8160 & \textbf{.0460}/.0452 & \textbf{.2461}/.2392 & \textbf{.6148}/.5963 & \textbf{.0384}/.0380 & .1441/\textbf{.1502} & \textbf{.7636}/.7320 & .0343/\textbf{.0357} & \textbf{.1943}/.1856 \\
DeepRemaster & \textbf{.6526}/.6028 & \textbf{.0326}/.0277 & \textbf{.1655}/.1367 & .5354/\textbf{.5541} & \textbf{.0391}/.0318 & \textbf{.1478}/.1351 & .7697/\textbf{.8061} & \textbf{.0377}/.0351 & .2065/\textbf{.2107} \\
RVRT         & .6256/\textbf{.6682} & \textbf{.0245}/.0243 & .1243/\textbf{.1337} & .5008/\textbf{.6061} & \textbf{.0245}/.0241 & .1043/\textbf{.1205} & \textbf{.7305}/.6799 & \textbf{.0341}/.0325 & \textbf{.1806}/.1599 \\
\bottomrule
\end{tabular}
\end{table}

Table~\ref{tab:interframe} reports the inter-frame stability measures tLPIPS
and CLIP-F. On real footage they mostly favour the baseline. We report them in
full because the disagreement with DOVER is informative: both are
\emph{self-similarity} measures, maximised in the limit by a model that
temporally over-smooths, and they cannot distinguish ``stable because the
artifact was removed consistently'' from ``stable because detail was
flattened''. Over-smoothing under the baseline pipeline is the behaviour
identified in Section~\ref{sec:experiments} and isolated in the Gaussian-grain
ablations (Section~\ref{sec:appendix-ablations}). On the synthetic set, where
the degradation is matched by construction, AbsoluteDegradation wins tLPIPS for
5 of 6 architectures. The human study, in which raters judged looping clips and
could see flicker directly, adjudicates in favour of AbsoluteDegradation.

\begin{table}[ht]
\centering
\caption{Inter-frame stability, tLPIPS~\citep{chu2020tecogan} ($\downarrow$) and CLIP-F ($\uparrow$), reported as AbsoluteDegradation / Bringing Old Films~\citep{wan2022oldfilm}.}
\label{tab:interframe}
\scriptsize
\setlength{\tabcolsep}{3pt}
\begin{tabular}{@{}lccc|ccc@{}}
\toprule
& \multicolumn{3}{c|}{\textbf{tLPIPS}~$\downarrow$} & \multicolumn{3}{c}{\textbf{CLIP-F}~$\uparrow$} \\
\textbf{Model} & Synthetic & Our dataset & SRWOV & Synthetic & Our dataset & SRWOV \\
\midrule
BasicVSR++   & 0.235 / 0.215 & 0.134 / 0.118 & 0.052 / 0.044 & 0.9712 / 0.9705 & 0.9883 / 0.9879 & 0.9861 / 0.9876 \\
Shift-Net    & 0.236 / 0.238 & 0.136 / 0.127 & 0.050 / 0.047 & 0.9722 / 0.9678 & 0.9880 / 0.9875 & 0.9854 / 0.9877 \\
MambaOFR     & 0.236 / 0.238 & 0.127 / 0.117 & 0.058 / 0.049 & 0.9678 / 0.9639 & 0.9875 / 0.9880 & 0.9825 / 0.9860 \\
RTN          & 0.235 / 0.239 & 0.139 / 0.134 & 0.058 / 0.053 & 0.9703 / 0.9690 & 0.9884 / 0.9888 & 0.9841 / 0.9862 \\
DeepRemaster & 0.248 / 0.251 & 0.139 / 0.125 & 0.088 / 0.064 & 0.9480 / 0.9436 & 0.9833 / 0.9845 & 0.9712 / 0.9839 \\
RVRT         & 0.270 / 0.299 & 0.126 / 0.150 & 0.049 / 0.050 & 0.9438 / 0.9398 & 0.9867 / 0.9837 & 0.9864 / 0.9876 \\
\bottomrule
\end{tabular}
\end{table}

\subsection{Ablation studies}
\label{sec:appendix-ablations}

\paragraph{BasicVSR++.}
Table~\ref{tab:ablations_basicvsr} ablates two components on BasicVSR++, the
architecture with the largest gain from AbsoluteDegradation. The full pipeline
achieves the best CLIPIQA+, MUSIQ, and MANIQA. Gaussian-only grain obtains the
best BRISQUE while performing worse on all three learned perceptual metrics,
further supporting the observation that BRISQUE can favour excessive smoothing
over faithful restoration.

\begin{table}[ht]
\centering
\caption{Ablations on BasicVSR++~\citep{chan2022basicvsrplusplus}, evaluated on \textbf{our dataset}. The full-pipeline row is the mean of seeds 42, 123, and 9999 (Table~\ref{tab:seeds}). Best results per column are \textbf{bolded}; second-best are \textit{italicized}.}
\label{tab:ablations_basicvsr}
\scriptsize
\begin{tabular}{@{}lcccc@{}}
\toprule
\textbf{Variant} & \textbf{BRISQUE}~$\downarrow$ & \textbf{CLIPIQA+}~$\uparrow$ & \textbf{MUSIQ}~$\uparrow$ & \textbf{MANIQA}~$\uparrow$ \\
\midrule
Gaussian grain only & \textbf{18.38} & \textit{0.358} & \textit{42.05} & \textit{0.242} \\
No moving lines     & \textit{20.84} & 0.359 & 41.43 & 0.227 \\
\midrule
\textbf{AbsoluteDegradation (full)} & 21.97 & \textbf{0.362} & \textbf{42.27} & \textbf{0.250} \\
\bottomrule
\end{tabular}
\end{table}

\paragraph{RTN.}
The broader ablation suite of Table~\ref{tab:ablations} was run on
RTN~\citep{wan2022oldfilm}, the architecture purpose-built for old-film
restoration and our strongest qualitative performer during development. We
prioritised it to probe the behaviour of a strong restoration-specific model
under synthetic degradations; because an ablation on one architecture cannot
attribute the gains observed on others, we complement it with the BasicVSR++
ablations above. These RTN ablations were conducted with a preliminary
parameterisation of the pipeline during development; the \emph{full} row of
Table~\ref{tab:ablations} is the reference for this study only and therefore
differs from the RTN results in Table~\ref{tab:results}.

The variants replace our signal-dependent blue noise with \textit{Gaussian
grain only}; remove \textit{gate-weave} camera shake; omit the negative-film
distortions; drop \textit{moving lines}; disable \textit{temporal coherence} by
limiting texture lifespan to a single frame; and restrict training to
\textit{medium severity only}. Although gate-weave is difficult to visualise
statically, introducing camera shake makes temporal alignment and optical-flow
estimation harder during training, which forces the network to extract and
aggregate cross-frame information more robustly.
Quantitative metrics again partly contradict visual quality: Gaussian-only
grain yields higher metric scores on our dataset, yet Figure~\ref{fig:ablation}
shows that it causes the network to over-smooth and hallucinate features when
uncertain; for example, the man's eyes are reduced to simple lines, while finer
details, such as subtle fabric folds in the hat and the faint star emblem, become
less distinct. Simulating moving lines explicitly teaches the network to
eliminate kinetic vertical scratches, and without sustaining synthesised
textures across multiple frames the network misinterprets temporally persistent
artifacts in real footage as scene geometry. Restricting the pipeline to medium
severity or omitting the negative-film distortions reduces training diversity
and generalisation.

\begin{table}[ht]
\centering
\caption{Ablation study on \textbf{our real dataset} and \textbf{SRWOV}~\citep{mambaofr} with RTN~\citep{wan2022oldfilm} as the backbone, conducted with a preliminary parameterisation of the pipeline; the full row is the reference for this study only.
  Best results per column are \textbf{bolded}; second-best results are \textit{italicized}.}
\label{tab:ablations}
\scriptsize
\setlength{\tabcolsep}{5pt}
\renewcommand{\arraystretch}{1.1}
\begin{tabular}{@{}l|ccc|ccc@{}}
\Xhline{2.5\arrayrulewidth}
& \multicolumn{3}{c|}{\textbf{Our dataset}} & \multicolumn{3}{c}{\textbf{SRWOV dataset~\citep{mambaofr}}}\\
\Xhline{1\arrayrulewidth}
\textbf{Variant} & \textbf{CLIPIQA+}$\uparrow$ & \textbf{MUSIQ}$\uparrow$ & \textbf{MANIQA}$\uparrow$ & \textbf{CLIPIQA+}$\uparrow$ & \textbf{MUSIQ}$\uparrow$ & \textbf{MANIQA}$\uparrow$ \\
\Xhline{1\arrayrulewidth}
Gaussian grain only         & \textbf{0.377} & \textbf{42.73} & \textbf{0.229}  & 0.415          & \textit{56.64} & \textit{0.300} \\
No gate-weave               & 0.353          & 40.89          & 0.213           & 0.418          & 56.23          & 0.289          \\
No neg./pos.\ distortions   & 0.335          & 36.71          & 0.198           & \textit{0.423} & 55.02          & 0.287          \\
No moving lines             & 0.350          & 39.93          & 0.208           & 0.406          & 55.74          & 0.282          \\
No temporal coherence       & 0.357          & 40.21          & 0.209           & 0.405          & 56.34          & 0.297          \\
Medium severity only        & 0.344          & 38.50          & 0.202           & \textbf{0.425} & 56.05          & 0.292          \\
\Xhline{1\arrayrulewidth}
\textbf{AbsoluteDegradation (full)} & \textit{0.360} & \textit{42.62} & \textit{0.214} & 0.420 & \textbf{56.96} & \textbf{0.301} \\
\Xhline{2.5\arrayrulewidth}
\end{tabular}
\end{table}

\begin{figure}[ht]
  \centering
  \setlength{\tabcolsep}{1pt}
  \renewcommand{\arraystretch}{0.75}
  \scriptsize
  \begin{tabular}{@{}c c c c c c@{}}
    \textbf{Baseline} & \textbf{Gaussian grain} &
    \textbf{Baseline} & \textbf{No moving lines} &
    \textbf{Baseline} & \textbf{No temporal coh.} \\
    \includegraphics[width=0.16\linewidth]{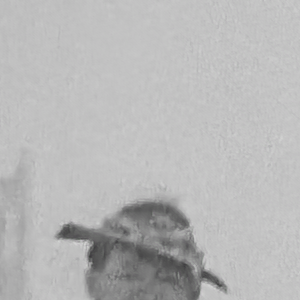} &
    \includegraphics[width=0.16\linewidth]{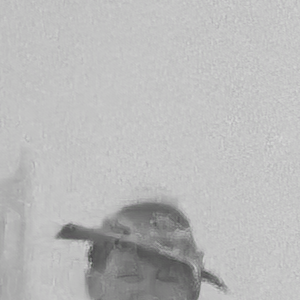} &
    \includegraphics[width=0.16\linewidth]{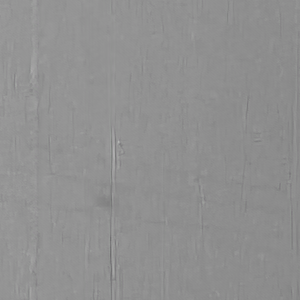} &
    \includegraphics[width=0.16\linewidth]{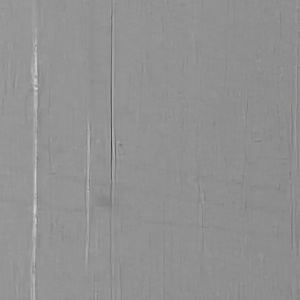} &
    \includegraphics[width=0.16\linewidth]{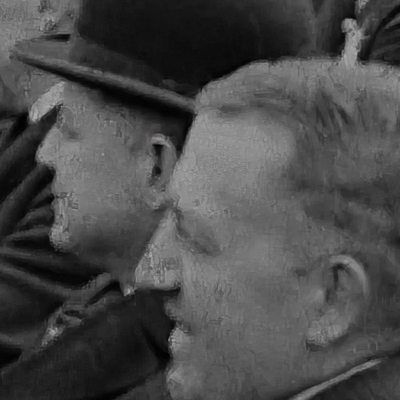} &
    \includegraphics[width=0.16\linewidth]{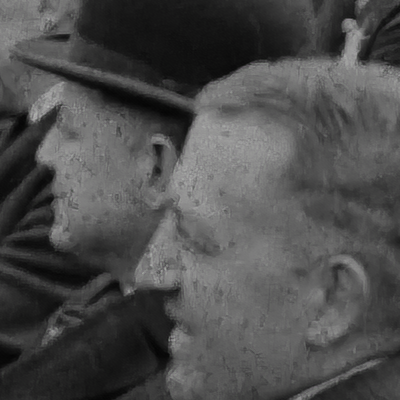} \\
  \end{tabular}
  \caption{Degradation pipeline ablations on our real dataset with RTN~\citep{wan2022oldfilm} (preliminary parameterisation, Table~\ref{tab:ablations}): full pipeline (baseline) vs.\ ablated variant.}
  \label{fig:ablation}
\end{figure}

\clearpage
\subsection{Qualitative comparisons}

\begin{figure}[ht]
  \centering
  \small
  \setlength{\tabcolsep}{2pt}
  \renewcommand{\arraystretch}{0.5}

  \newcommand{\makeimgrow}[2]{%
    \textbf{#1} &
    \raisebox{-0.5\height}{\includegraphics[width=0.28\textwidth]{figures/restoration_comparison/old_real_dataset/data_last_real_BQIGiE5vSSM_20-25/#2.jpg}} &
    \raisebox{-0.5\height}{\includegraphics[width=0.28\textwidth]{figures/restoration_comparison/old_real_dataset/rtn_new_last_real_BQIGiE5vSSM_20-25/#2.jpg}} &
    \raisebox{-0.5\height}{\includegraphics[width=0.28\textwidth]{figures/restoration_comparison/old_real_dataset/rtn_old_last_real_BQIGiE5vSSM_20-25/#2.jpg}} \\
  }

  \begin{tabular}{m{1.5cm} c c c}
    & \parbox[c]{0.3\textwidth}{\centering Original frames}
    & \parbox[c]{0.3\textwidth}{\centering RTN\\ AbsoluteDegradation (Ours)}
    & \parbox[c]{0.3\textwidth}{\centering RTN \\ Bringing Old Films Pipeline \citep{wan2022oldfilm}} \\
    \addlinespace[4pt]

    \makeimgrow{Frame 124}{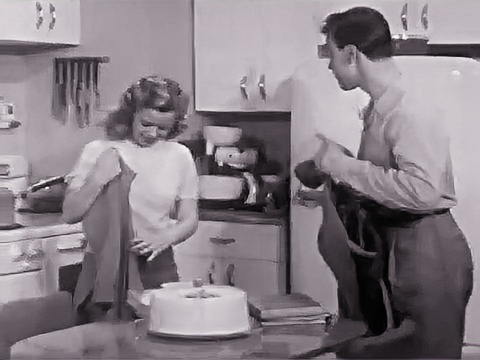}
    \addlinespace[2pt]

    \makeimgrow{Frame 125}{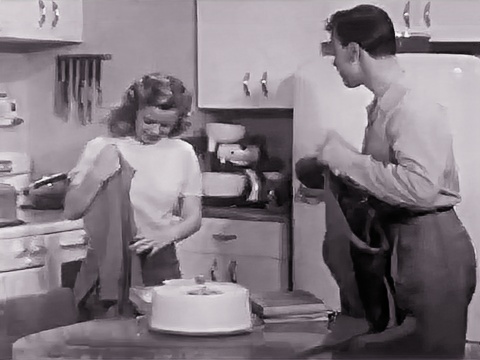}
    \addlinespace[2pt]

    \makeimgrow{Frame 126}{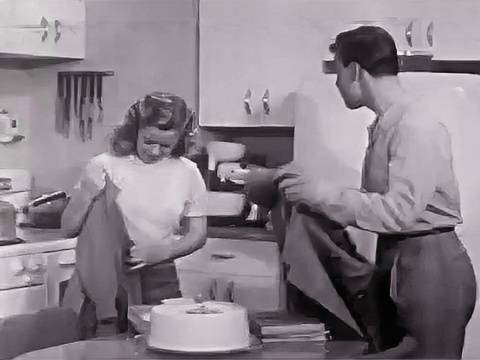}
    \addlinespace[2pt]

    \makeimgrow{Frame 127}{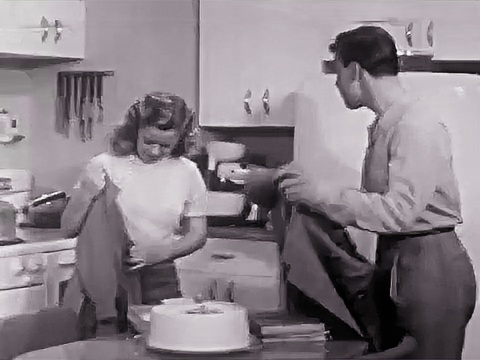}
    \addlinespace[2pt]

  \end{tabular}

  \caption{Visual comparison of our pipeline with Bringing Old Films Pipeline \citep{wan2022oldfilm} on video from SRWOV benchmark. Output of RTN~\citep{wan2022oldfilm} model.
  }
  \label{fig:restoration_comparison_srwov}
\end{figure}

\begin{figure}[ht]
  \centering
  \small
  \setlength{\tabcolsep}{2pt}
  \renewcommand{\arraystretch}{0.5}

  \newcommand{\makerow}[2]{%
    \parbox[c]{3.0cm}{\centering #1} &
    \raisebox{-0.5\height}{\includegraphics[width=0.18\textwidth]{figures/restoration_comparison/duel/#2_frame1.jpg}} &
    \raisebox{-0.5\height}{\includegraphics[width=0.18\textwidth]{figures/restoration_comparison/duel/#2_frame2.jpg}} &
    \raisebox{-0.5\height}{\includegraphics[width=0.18\textwidth]{figures/restoration_comparison/duel/#2_frame3.jpg}} &
    \raisebox{-0.5\height}{\includegraphics[width=0.18\textwidth]{figures/restoration_comparison/duel/#2_frame3_roi.jpg}} \\

  }
  \begin{tabular}{m{3cm} c c c c}
    & \textbf{Frame 1} & \textbf{Frame 2} & \textbf{Frame 3} & \textbf{ROI} \\

    \makerow{\textbf{Original frames}}{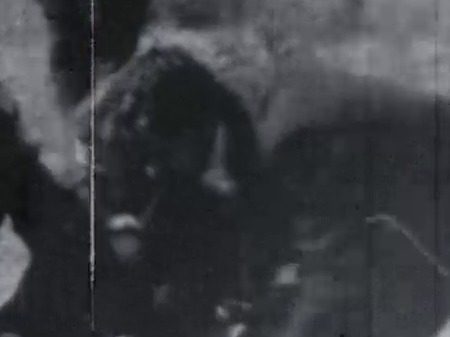}
    \addlinespace[2pt]

    \makerow{\textbf{MambaOFR}\\\scriptsize AbsoluteDegradation (Ours)}{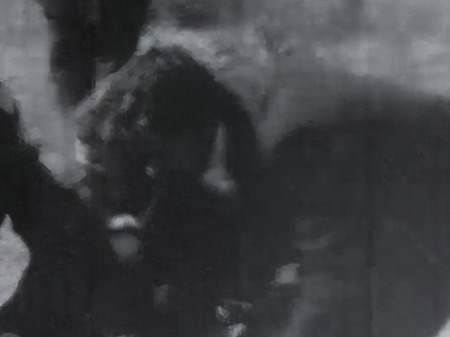}
    \addlinespace[2pt]

    \makerow{\textbf{MambaOFR}\\\scriptsize Bringing Old Films Pipeline \citep{wan2022oldfilm}}{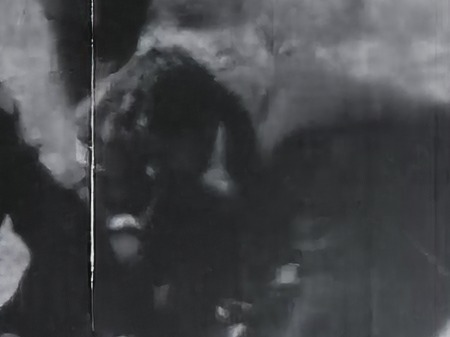}
    \addlinespace[2pt]

    \makerow{\textbf{RTN}\\\scriptsize AbsoluteDegradation (Ours)}{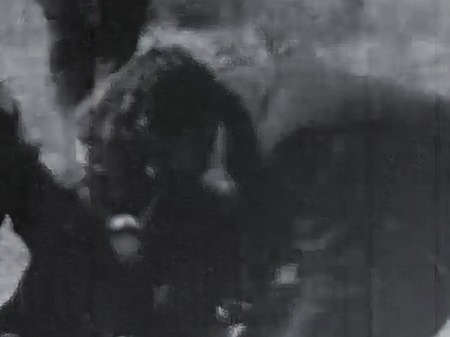}
    \addlinespace[2pt]

    \makerow{\textbf{RTN}\\\scriptsize Bringing Old Films Pipeline \citep{wan2022oldfilm}}{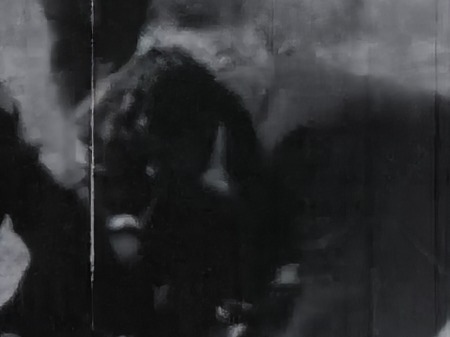}
    \addlinespace[2pt]

    \makerow{\textbf{BasicVSR++}\\\scriptsize AbsoluteDegradation (Ours)}{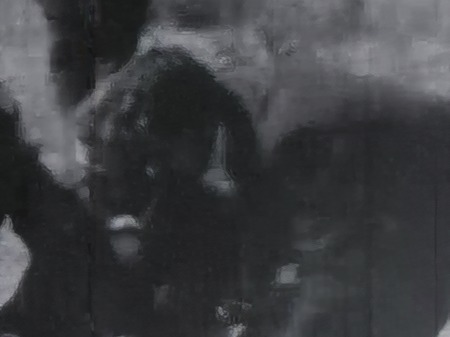}
    \addlinespace[2pt]

    \makerow{\textbf{BasicVSR++}\\\scriptsize  Bringing Old Films Pipeline \citep{wan2022oldfilm}}{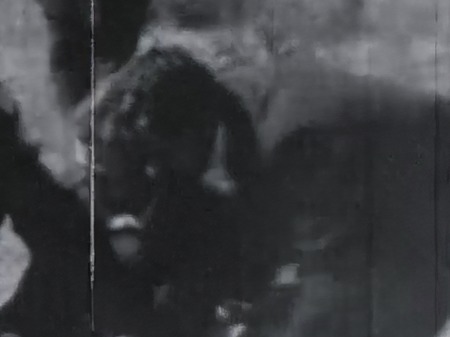}
    \addlinespace[2pt]

    \makerow{\textbf{RVRT}\\\scriptsize AbsoluteDegradation (Ours)}{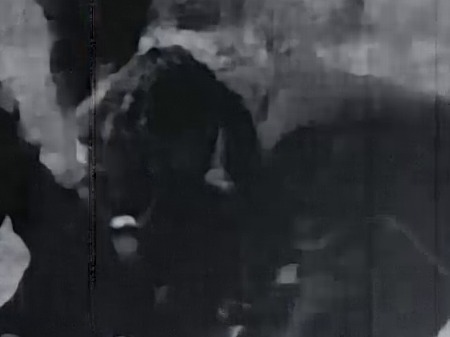}
    \addlinespace[2pt]

    \makerow{\textbf{RVRT}\\\scriptsize  Bringing Old Films Pipeline \citep{wan2022oldfilm}}{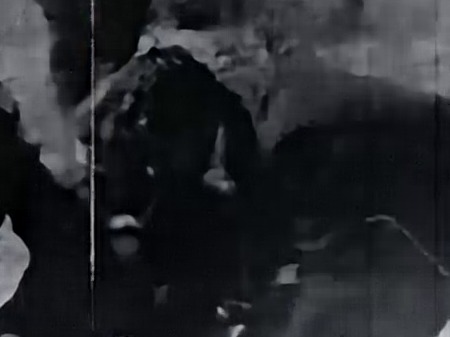}
    \addlinespace[2pt]

  \end{tabular}

  \caption{Visual comparison against Bringing Old Films Pipeline \citep{wan2022oldfilm}. Our approach outperforms the baseline in scratch removal in every tested restoration model
  }
  \label{fig:restoration_comparison_duel}
\end{figure}

\begin{figure}[ht]
  \centering
  \small
  \setlength{\tabcolsep}{2pt}
  \renewcommand{\arraystretch}{0.5}

  \newcommand{\makerow}[5]{%
    \parbox[c]{3.0cm}{\centering #1} &
    \raisebox{-0.5\height}{\includegraphics[width=0.18\textwidth]{figures/restoration_comparison/rosevelt/full_with_crop/#2}} &
    \raisebox{-0.5\height}{\includegraphics[width=0.18\textwidth]{figures/restoration_comparison/rosevelt/crops/#3}} &
    \raisebox{-0.5\height}{\includegraphics[width=0.18\textwidth]{figures/restoration_comparison/cowboys/full_with_crop/#4}} &
    \raisebox{-0.5\height}{\includegraphics[width=0.18\textwidth]{figures/restoration_comparison/cowboys/crops/#5}} \\
  }

  \begin{tabular}{m{3cm} c c c c}
    & \parbox[c]{0.18\textwidth}{\centering \textbf{Clip 1}} & \parbox[c]{0.18\textwidth}{\centering \textbf{ROI}} &
      \parbox[c]{0.18\textwidth}{\centering \textbf{Clip 2}} & \parbox[c]{0.18\textwidth}{\centering \textbf{ROI}} \\
    \addlinespace[4pt]

    \makerow{\textbf{Original frame}}{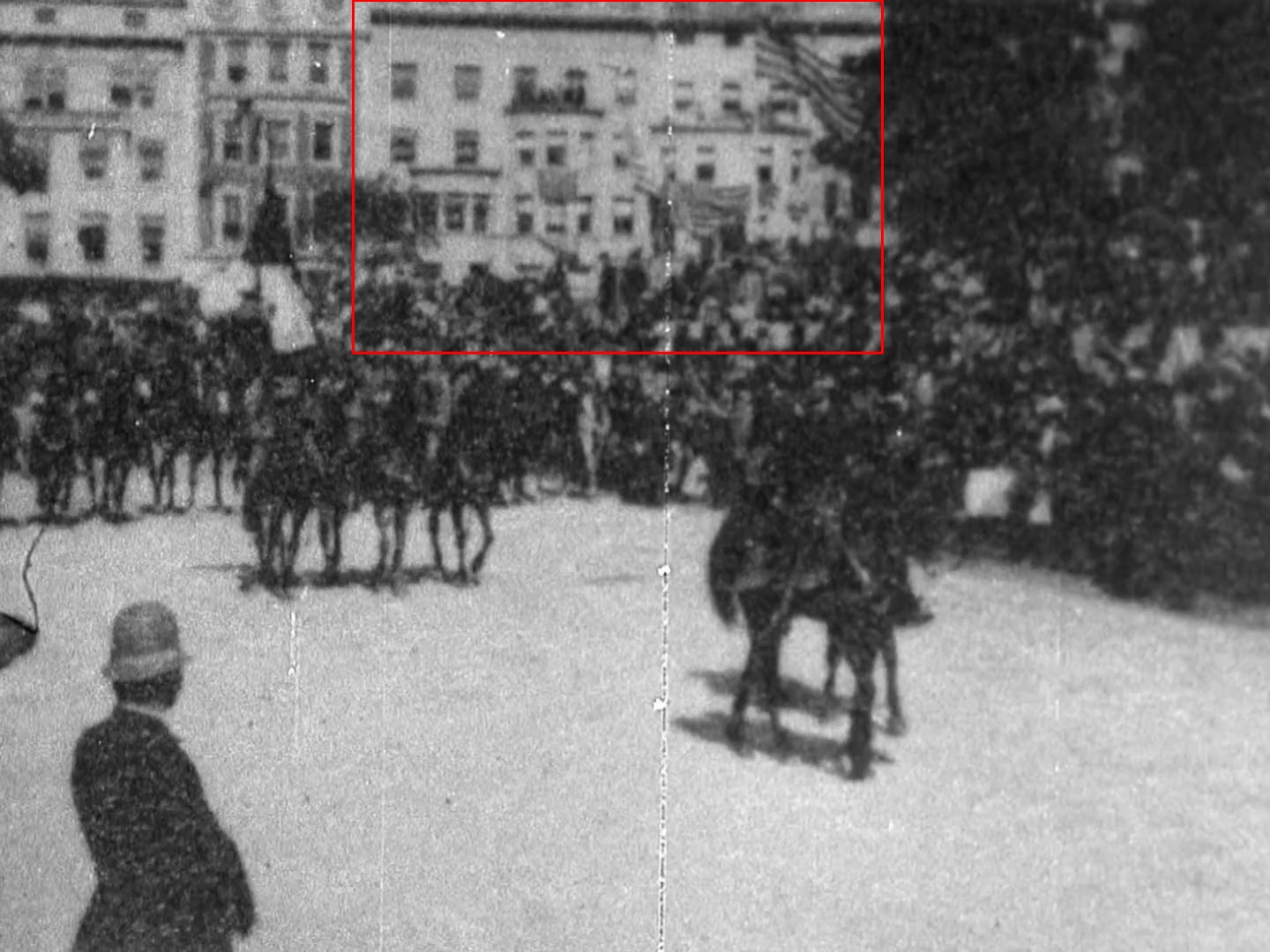}{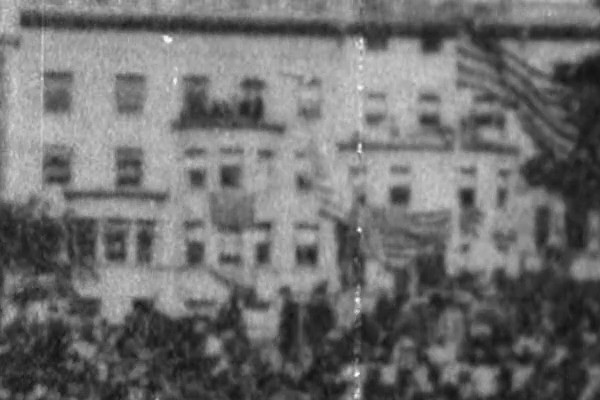}{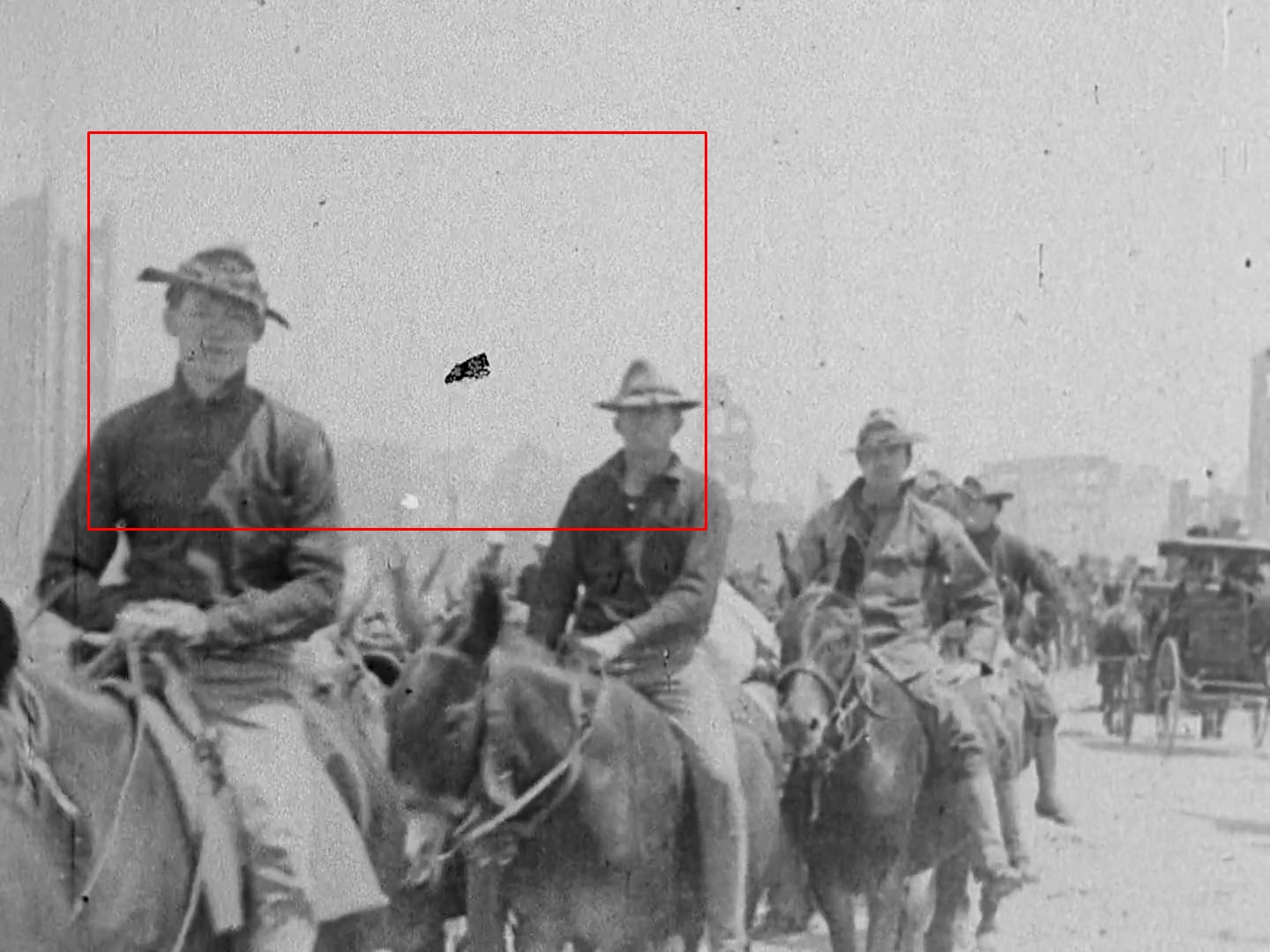}{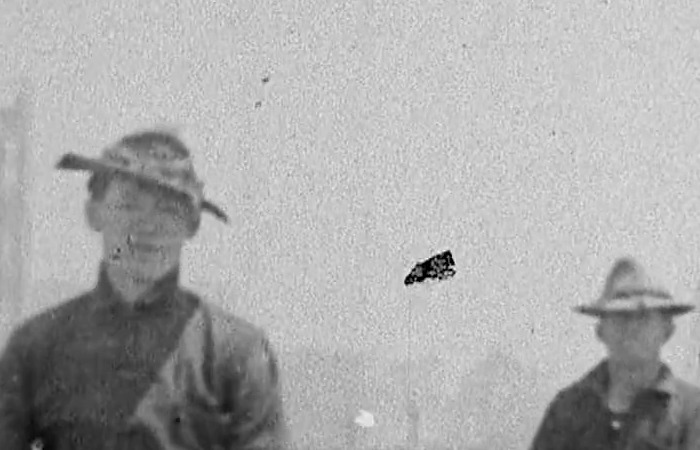}
    \addlinespace[2pt]

    \makerow{\textbf{MambaOFR}\\\scriptsize AbsoluteDegradation (Ours)}{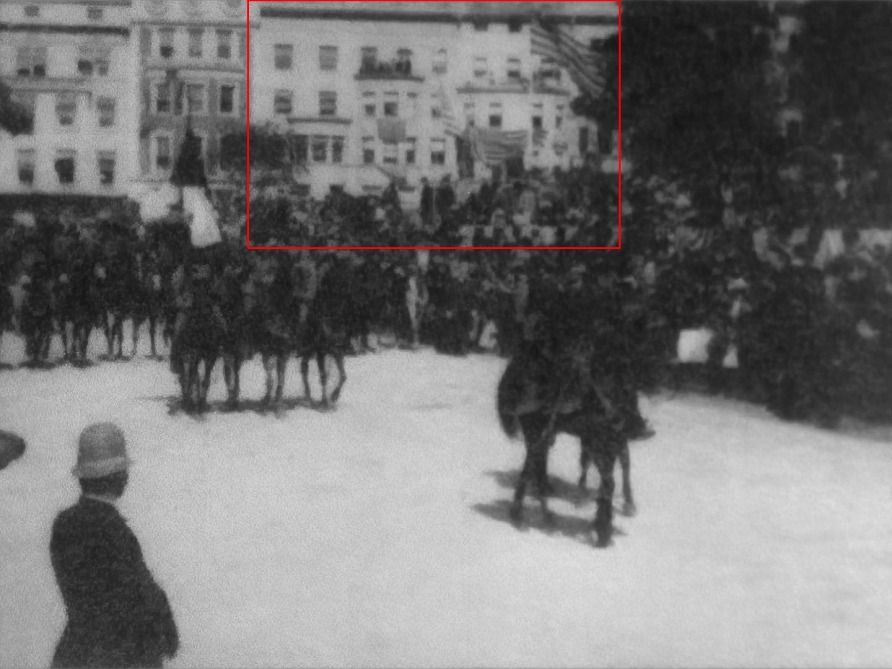}{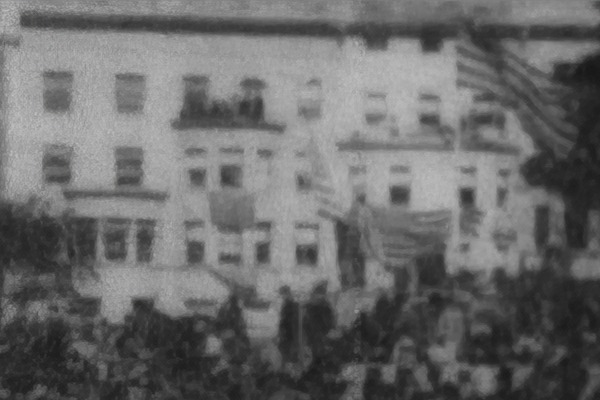}{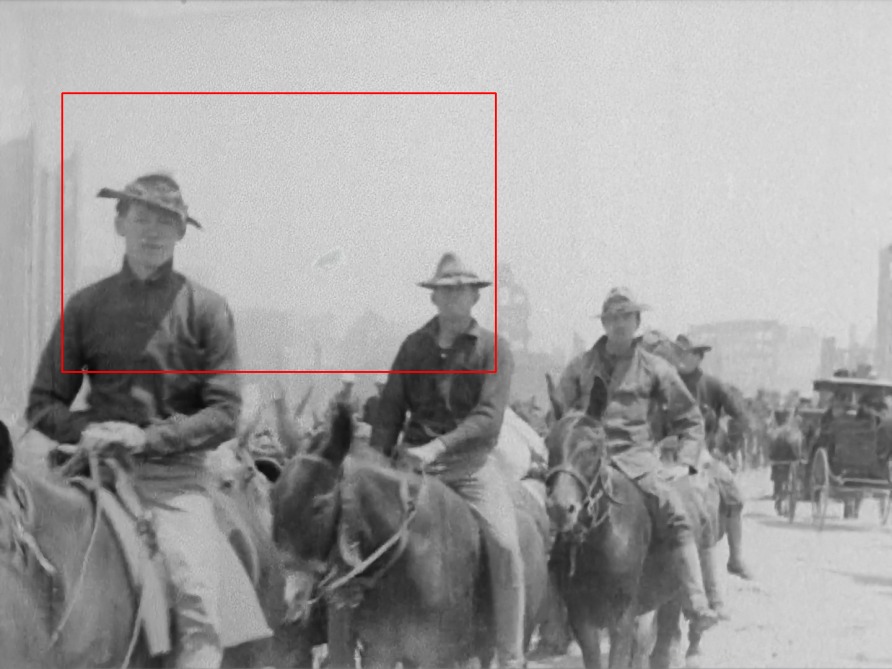}{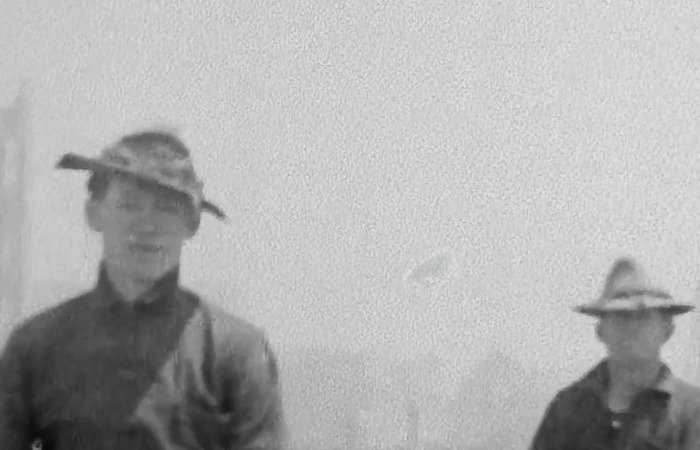}
    \addlinespace[2pt]

    \makerow{\textbf{MambaOFR}\\\scriptsize Bringing Old Films Pipeline \citep{wan2022oldfilm}}{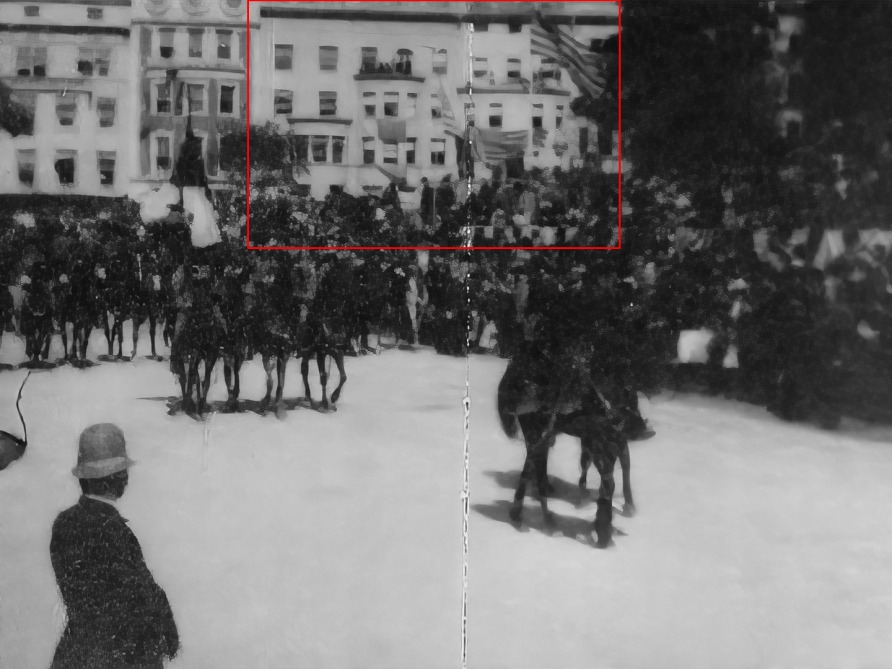}{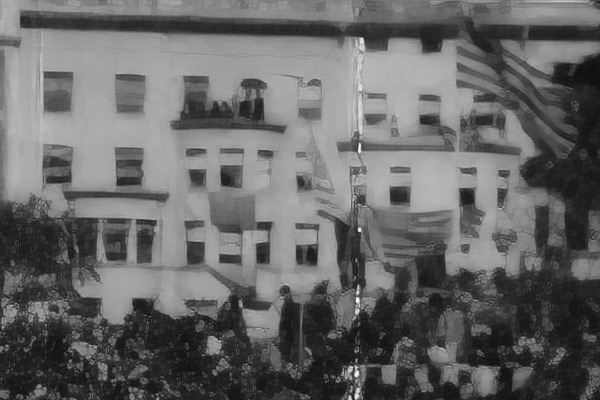}{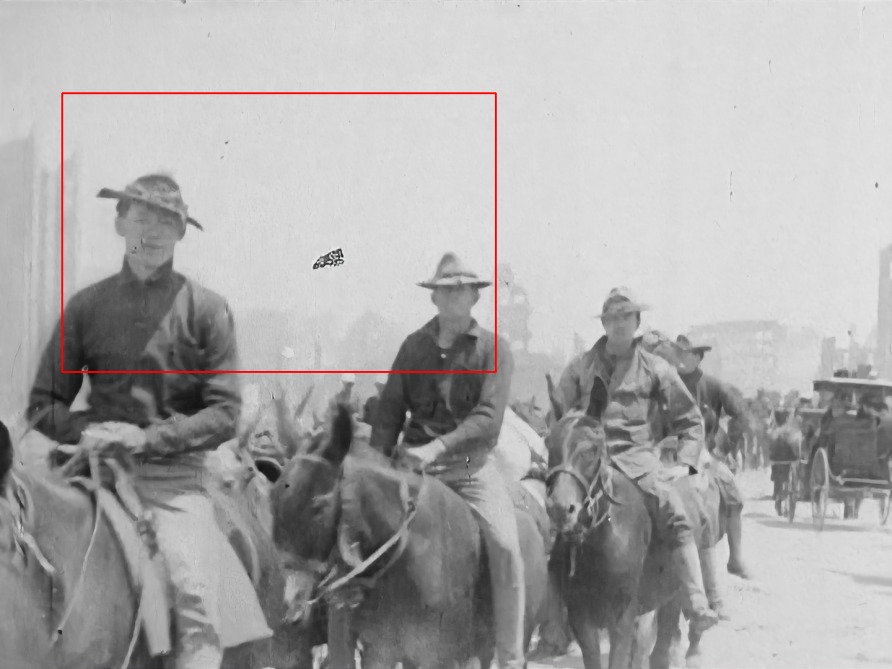}{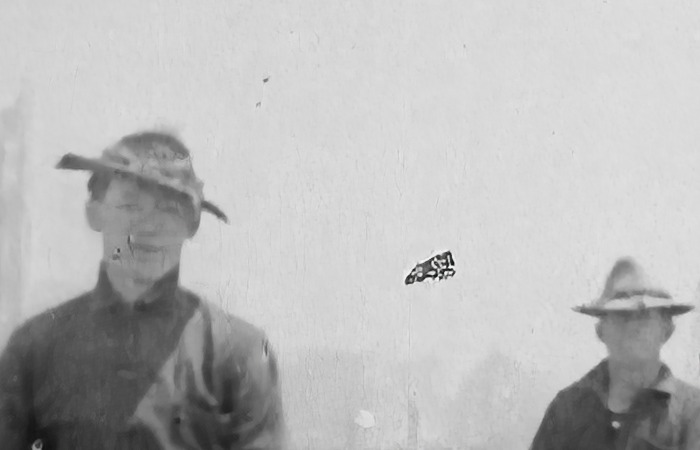}
    \addlinespace[2pt]

    \makerow{\textbf{RTN}\\\scriptsize AbsoluteDegradation (Ours)}{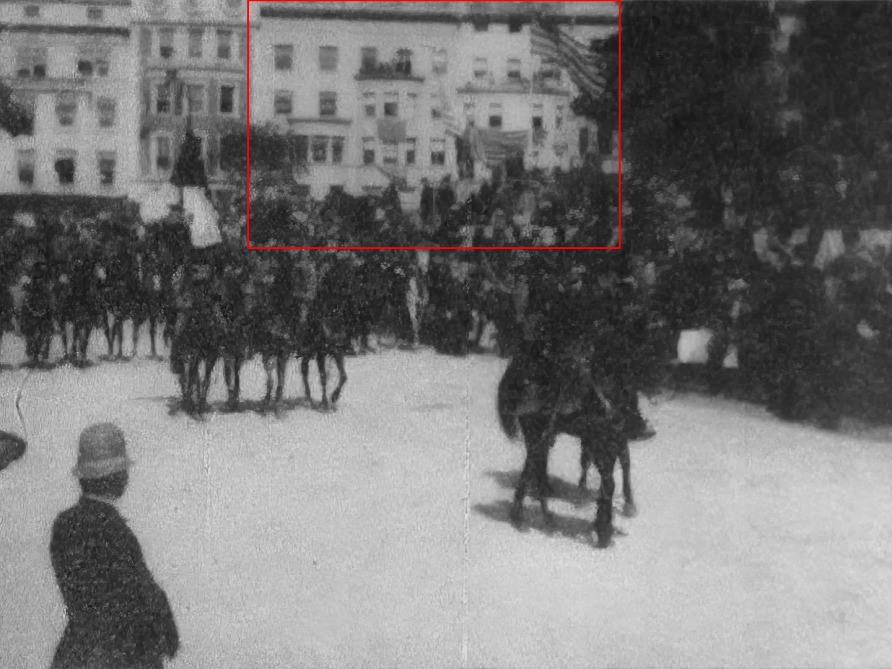}{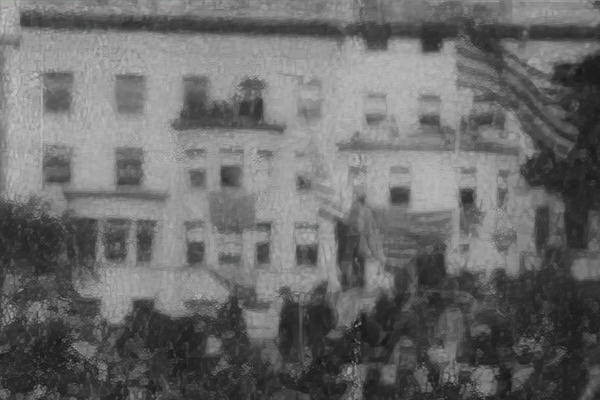}{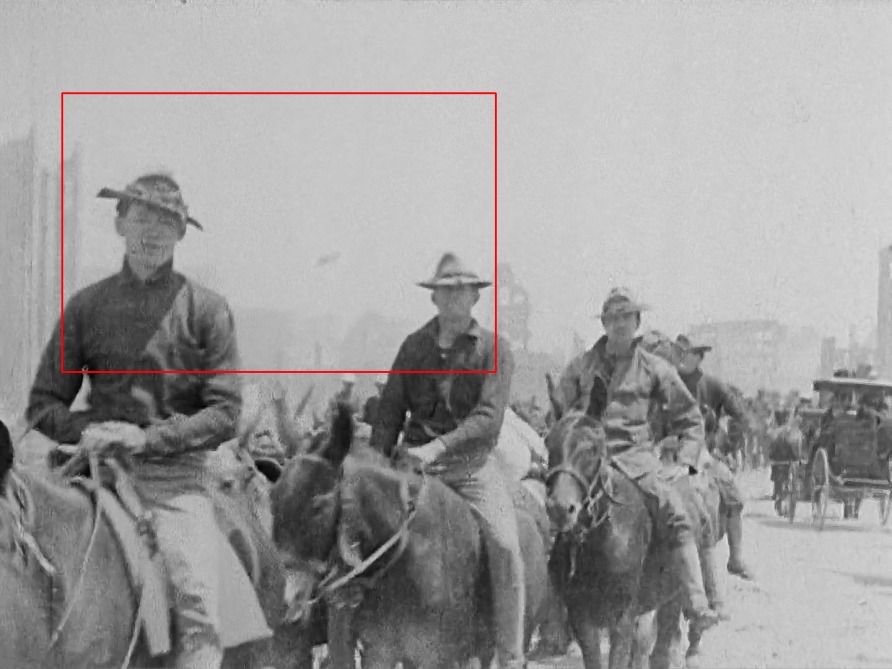}{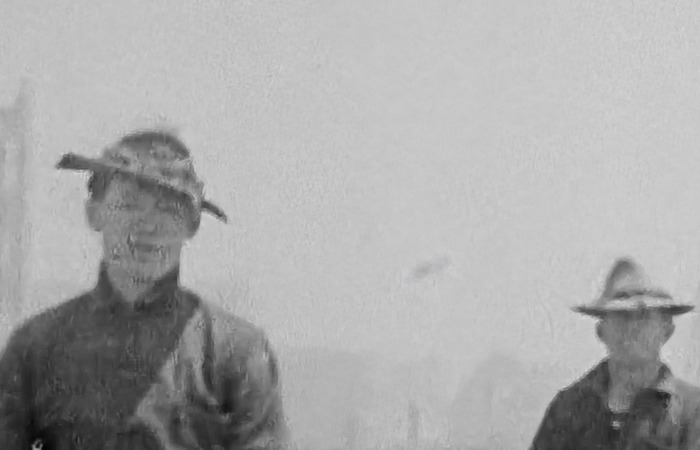}
    \addlinespace[2pt]

    \makerow{\textbf{RTN}\\\scriptsize Bringing Old Films Pipeline \citep{wan2022oldfilm}}{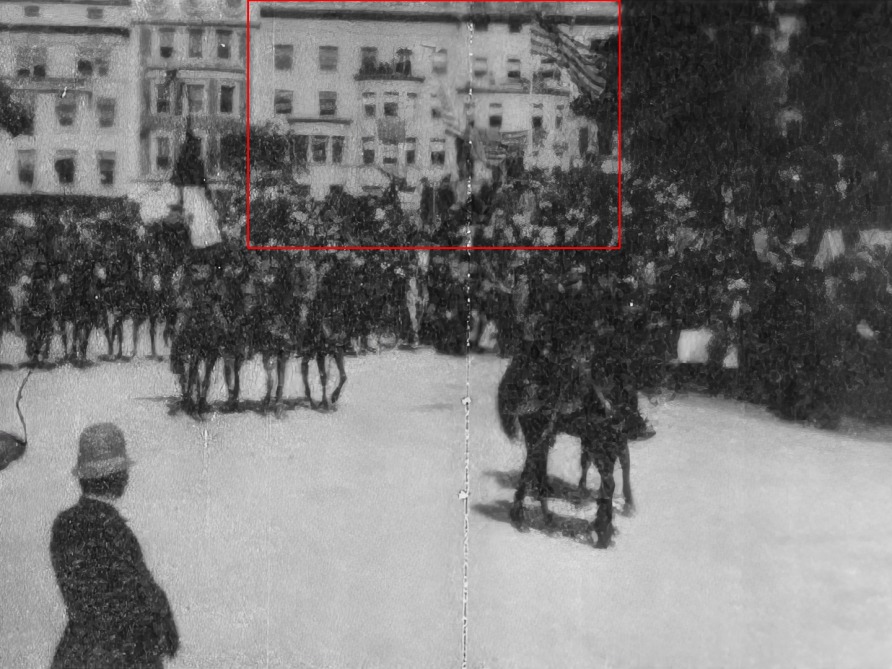}{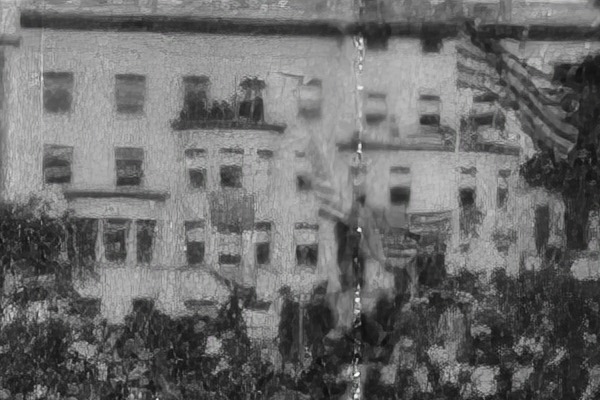}{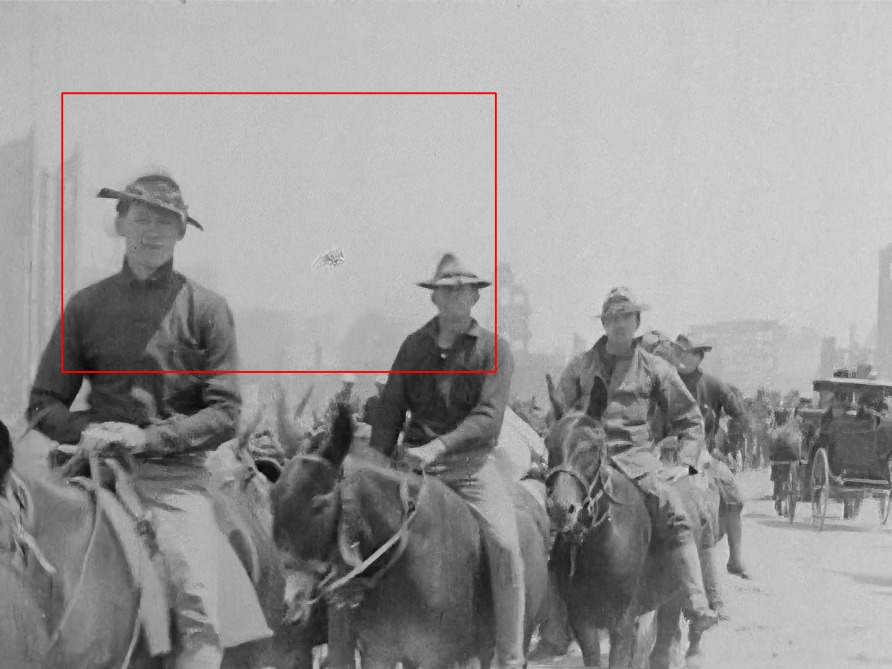}{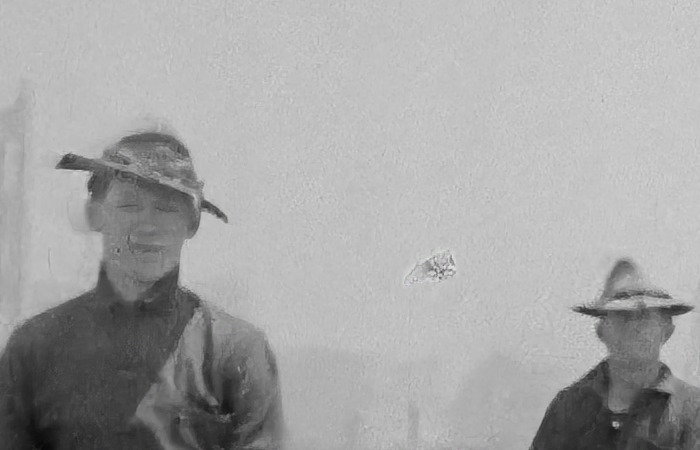}
    \addlinespace[2pt]

    \makerow{\textbf{BasicVSR++}\\\scriptsize AbsoluteDegradation (Ours)}{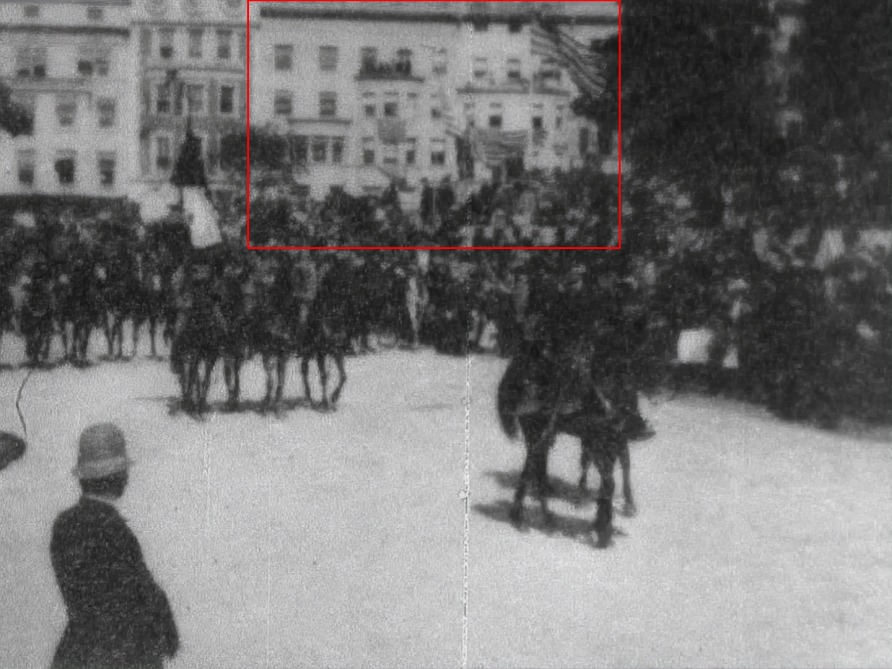}{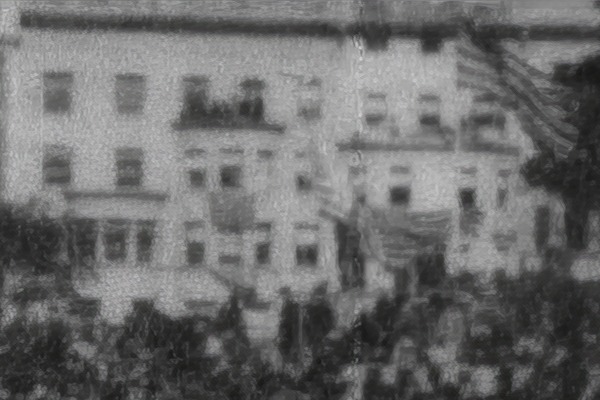}{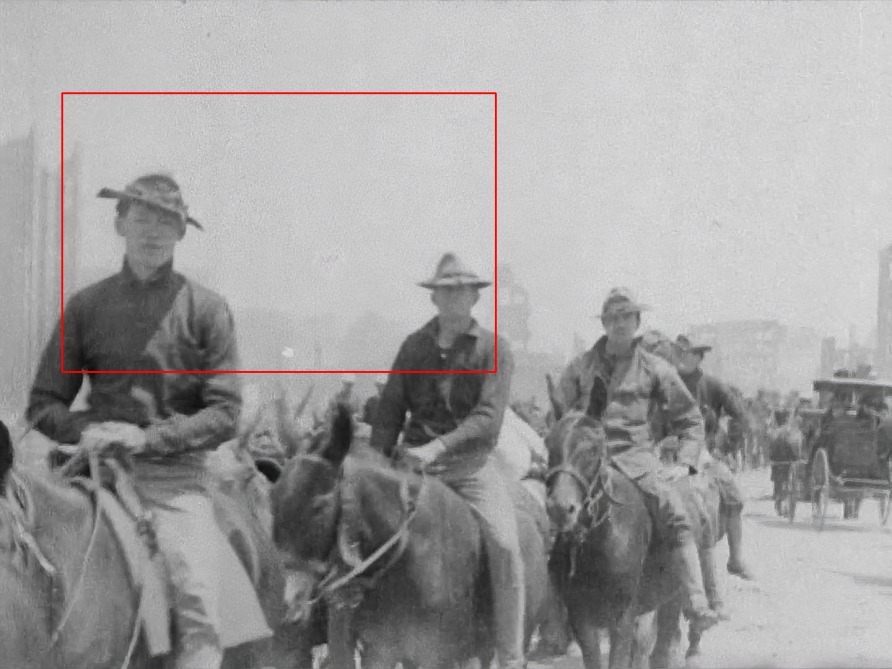}{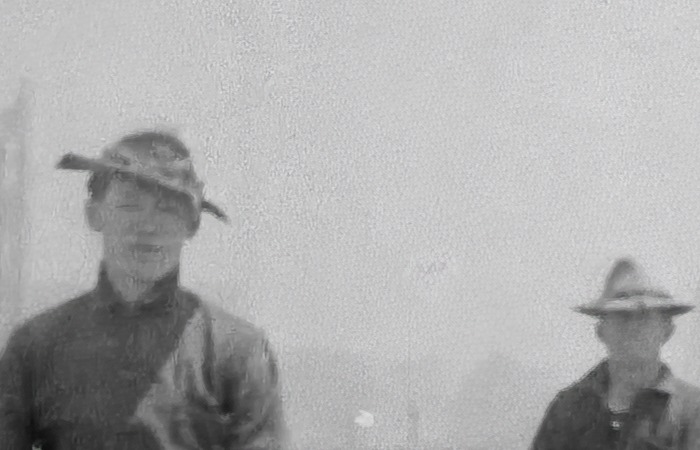}
    \addlinespace[2pt]

    \makerow{\textbf{BasicVSR++}\\\scriptsize Bringing Old Films Pipeline \citep{wan2022oldfilm}}{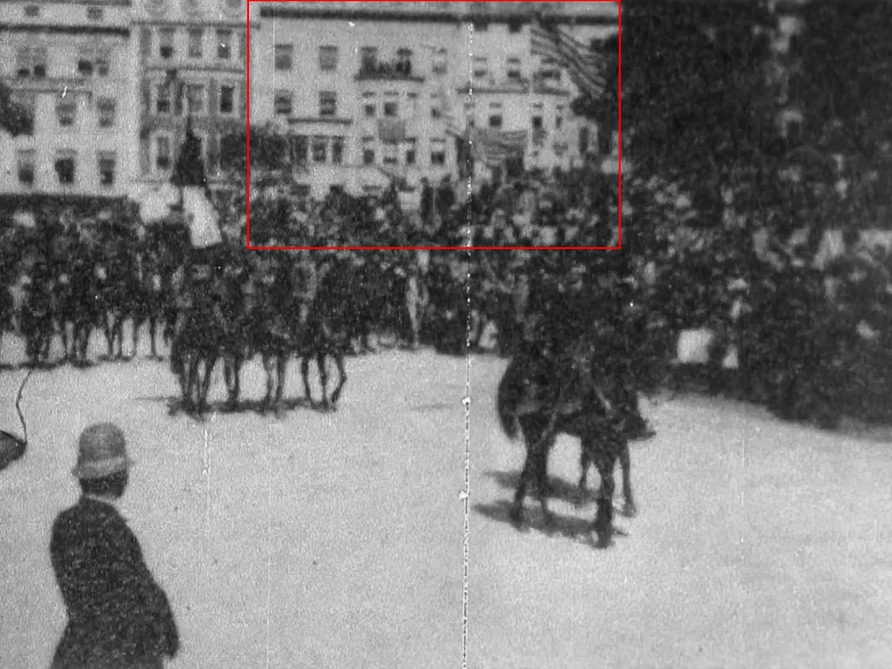}{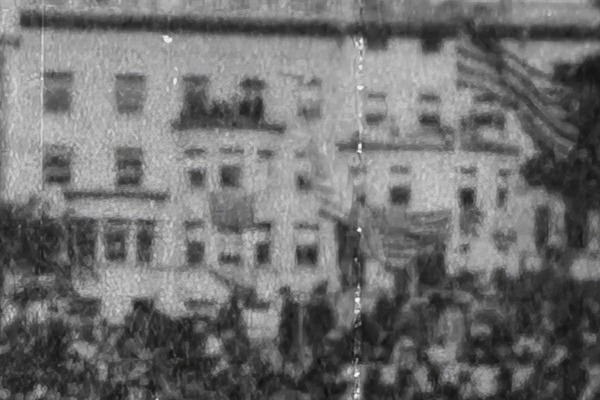}{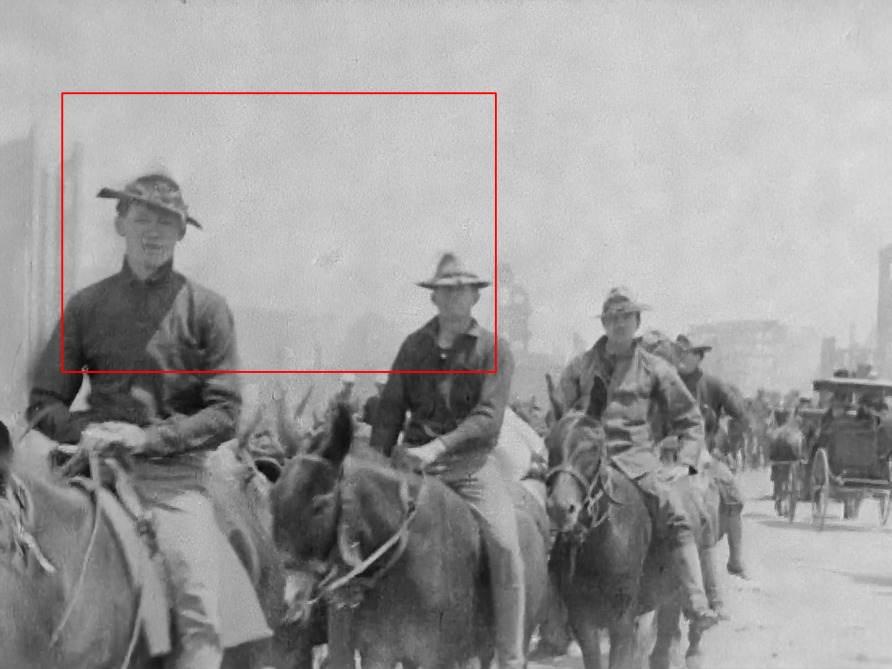}{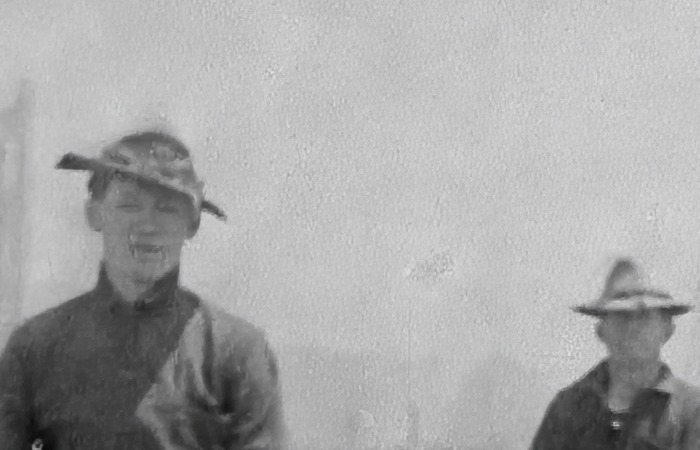}

  \end{tabular}

  \caption{Visual comparison of our pipeline with Bringing Old Films Pipeline \citep{wan2022oldfilm}.
  Our pipeline handles removal of scratches, while the other tends to over-smooth or sharpen degradations they did not remove.
  }
  \label{fig:restoration_comparison_roosevelt}
\end{figure}

\clearpage

\section{Open Resources, Licenses, and Safeguards}
\label{sec:appendix-resources}

\paragraph{Reproducibility and open resources.}
To support reproducibility, we release a code repository and data
resources with executable commands and configuration files:
\url{https://github.com/TytanMikJas/AbsoluteDegradation}
(code), \url{https://www.kaggle.com/absolutedegradation/datasets}
(datasets), and source links documented in the README, including REDS
\url{https://seungjunnah.github.io/Datasets/reds.html} and the
Bringing Old Films repository
\url{https://github.com/raywzy/Bringing-Old-Films-Back-to-Life}.
The released pipeline includes deterministic execution via fixed seed
(\texttt{seed=42}) and the complete degradation configuration of
Table~\ref{tab:severity}.

\paragraph{Licenses, terms, and safeguards.}
External assets used in our pipeline are limited to REDS
(CC-BY-4.0), Bringing Old Films texture resources (as distributed
by their repository), and Library of Congress public-domain archival films
(\url{https://www.loc.gov}). We respectfully acknowledge the Library of
Congress and REDS in the Kaggle release materials and in the Croissant metadata
file provided with the dataset package. As a release safeguard, all archival
clips are manually reviewed to remove harmful, unethical, or
non-restoration-relevant segments, and the release is scoped to restoration
benchmarking with source provenance metadata for each curated subset.
The accompanying Croissant metadata file is fully compliant with the
Croissant~1.0 specification and includes all required Responsible AI (RAI)
fields.

\end{document}